# A clinical-grade universal foundation model for intraoperative pathology

Zihan Zhao[1†], Fengtao Zhou[2†], Ronggang Li[3†], Bing Chu[4], Xinke Zhang[1], Xueyi Zheng[1], Ke Zheng[1], Xiaobo Wen[1], Jiabo Ma[2], Yihui Wang[2], Jiewei Chen[1], Chengyou Zheng[1], Jiangyu Zhang[5], Yongqin Wen[6], Jiajia Meng[7], Ziqi Zeng[8], Xiaoqing Li[3], Jing Li[9], Dan Xie[1], Yaping Ye[10], Yu Wang[8], Hao Chen[2,11,12,13,14*], Muyan Cai[1*]

1 Department of Pathology, State Key Laboratory of Oncology in South China, Guangdong Provincial Clinical Research Center for Cancer, Sun Yat-sen University Cancer Center, Guangzhou, China.
2 Department of Computer Science and Engineering, Hong Kong University of Science and Technology, Hong Kong, China.
3 Department of Pathology, Jiangmen Central Hospital, Jiangmen, China.
4 Department of Pathology, Zhongshan People's Hospital, Zhongshan, China.
5 Department of Pathology, Affiliated Cancer Hospital & Institute of Guangzhou Medical University, Guangzhou, China.
6 Department of Pathology, The Tenth Affiliated Hospital, Southern Medical University, Dongguan People's Hospital, Dongguan, China.
7 Department of Pathology, Huizhou Central People's Hospital, Huizhou, China.
8 Department of Pathology, Zhujiang Hospital, Southern Medical University, Guangzhou, China.
9 Zhongshan School of Medicine, Sun Yat-sen University, Guangzhou, China.
10 Department of Pathology, Nanfang Hospital, Southern Medical University, Guangzhou, China.
11 Department of Chemical and Biological Engineering, Hong Kong University of Science and Technology, Hong Kong, China.
12 Division of Life Science, Hong Kong University of Science and Technology, Hong Kong, China.
13 HKUST Shenzhen-Hong Kong Collaborative Innovation Research Institute, Futian, Shenzhen, China.

14 State Key Laboratory of Nervous System Disorders, The Hong Kong University of Science and Technology, Hong Kong, China.

†Contributed equally.

*Drs. Muyan Cai and Hao Chen are the co-corresponding authors.

Correspondence: Dr. Muyan Cai, Department of Pathology, Collaborative Innovation Center for Cancer Medicine, State Key Laboratory of Oncology in South China, Sun Yat-sen University Cancer Center, Guangzhou, 510060, China. Email: caimy@sysucc.org.cn;

Hao Chen, Department of Computer Science and Engineering, Hong Kong University of Science and Technology, Hong Kong, China. Email: jhc@cse.ust.hk.

**Abstract**

Intraoperative pathology is pivotal to precision surgery, yet its clinical impact is constrained by diagnostic complexity and the limited availability of high-quality frozen section data. While computational pathology has made significant strides, the lack of large-scale, prospective validation has impeded its routine adoption in surgical workflows. Here, we introduce CRISP, a clinical-grade foundation model developed on over 100,000 frozen sections from eight medical centers, specifically designed to provide Clinical-grade Robust Intraoperative Support for Pathology (CRISP). CRISP was comprehensively evaluated on more than 15,000 intraoperative slides across nearly 100 retrospective diagnostic tasks, including benign-malignant discrimination, key intraoperative decision-making, and pan-cancer detection, etc. The model demonstrated robust generalization across diverse institutions, tumor types, and anatomical sites-including previously unseen sites and rare cancers. In a prospective cohort of over 2,000 patients, CRISP sustained high diagnostic accuracy under real-world conditions, directly informing surgical decisions in 92.6% of cases. Human-AI collaboration further reduced diagnostic workload by 35%, avoided 105 ancillary tests and enhanced detection of micrometastases with 87.5% accuracy. Together, these findings position CRISP as a clinical-grade paradigm for AI-driven intraoperative pathology, bridging computational advances with surgical precision and accelerating the translation of artificial intelligence into routine clinical practice.

# Introduction

Rapid and accurate intraoperative diagnosis is a cornerstone of modern surgical practice, directly influencing patient outcomes and guiding critical decisions in real time.[1,2] For over a century, frozen section examination has served as an indispensable modality for intraoperative pathological assessment, enabling surgeons to make informed choices regarding tumor resection, margin status, and differential diagnosis across a wide spectrum of surgical specialties.[3-6] The widespread adoption of frozen sections is driven by their expedited preparation and diagnostic accuracy, which closely parallels that of formalin-fixed paraffin-embedded (FFPE) specimens.[7,8] As a result, frozen section analysis has become integral to precision surgical management, supporting timely interventions that can profoundly improve patient outcomes.

Despite its clinical utility, frozen section diagnosis presents formidable challenges. The inherent morphological complexity of frozen tissue, compounded by the urgency of intraoperative decision-making, demands rapid and highly accurate interpretation. These constraints introduce significant risks, including the potential for misdiagnosis and suboptimal surgical outcomes.[9,10] The need for robust, scalable support systems to assist pathologists and surgeons in real-time intraoperative workflows is therefore both pressing and universal.

Recent advances in computational pathology and artificial intelligence (AI) have begun to reshape the landscape of diagnostic medicine. Deep learning models have demonstrated remarkable promise in automating and augmenting pathological assessment, offering the potential to enhance diagnostic consistency and efficiency. However, the application of AI to intraoperative frozen section analysis remains in its infancy.[11] Prior studies have explored deep learning approaches in specific disease contexts,[12-14] yet these efforts are frequently constrained by small, homogeneous datasets that fail to capture the full spectrum of morphological and technical variability encountered in clinical practice.[15] As a result, the development of generalizable and clinically practical AI models for intraoperative diagnosis has remained elusive.

The emergence of large-scale AI models, or foundation models, has further fueled optimism for the clinical translation of computational pathology.[15-21] Pretrained on vast and diverse datasets, these models have shown strong generalizability across a range of diagnostic tasks.[22] However, most existing foundation models are built upon FFPE specimens,[23] which differ fundamentally from frozen sections in both preparation and tissue morphology. Direct application of FFPE-trained models to intraoperative frozen section risks compromising diagnostic accuracy,[10,24,25] while attempts to bridge this gap using virtual staining techniques raise concerns regarding reliability and clinical trustworthiness of synthetic images.[26-28] Additionally, the lack of prospective validation in most studies limits their readiness for real-world deployment.[29] Addressing these challenges requires the development of a robust and generalizable foundation model specifically tailored to frozen section pathology, with demonstrated clinical validity in diverse and realistic settings. In this study, we introduce CRISP, an AI-enabled comprehensive deep learning framework designed to support a wide spectrum of intraoperative diagnostic and predictive tasks. Leveraging the largest frozen section dataset assembled to date-comprising over 100,000 slides from eight medical centers across different regions-CRISP was pretrained on 50 million image patches from 25 anatomical sites. The model's performance was systematically evaluated across multiple downstream tasks commonly encountered in intraoperative pathology, with external validation cohorts drawn from independent institutions. To further assess its real-world clinical utility, we conducted a registered prospective observational study involving more than 2,000 patients in real-world surgical workflows. Through both rigorous retrospective and prospective analyses, our findings demonstrate the practical value and broad applicability of a frozen-specific foundation model, making a significant advance toward the integration of AI-driven decision support in precision surgery.

## Results

### Patient cohorts and study design

The development of a clinically robust foundation model for intraoperative diagnosis necessitates a comprehensive understanding of the biological and technical heterogeneity inherent in real-world surgical pathology.[30] To address this, we constructed the largest multi-center frozen section dataset to date, comprising 101,796 frozen section slides obtained primarily from eight medical centers and spanning a broad spectrum of anatomical sites (Fig. 1a-c). This dataset was specifically curated to reflect the diversity of clinical scenarios and laboratory conditions encountered during intraoperative consultations, as detailed in the

Methods.

For model pre-training, we utilized 81,667 frozen section slides from three major centers and the TCGA frozen section dataset,[31] generating over 50 million pathological image patches (Fig. 1d). To rigorously assess the model's robustness and generalizability, we established 98 downstream diagnostic tasks designed to emulate critical intraoperative decision-making scenarios. Validation was conducted across multiple centers, including pan-cancer analyses, rare disease subtypes, and challenging conditions (Fig. 1e).

To further bridge the gap between proof-of-concept and clinical translation, we conducted a large-scale prospective study (Fig. 1f). This study enrolled 2,071 consecutive patients undergoing thyroid, breast, or lung surgery with intraoperative consultation at Sun Yat-sen University Cancer Center between November 11, 2024, and July 1, 2025. Strict inclusion and exclusion criteria were applied to minimize selection bias and ensure cohort representativeness, with all eligible patients systematically recorded. Detailed criteria for both retrospective and prospective cohorts are presented in the Methods section. Baseline clinical characteristics of each cohort are provided in Supplementary Tables S1.1-1.11.

**Classification of common benign and malignant lesions**

A primary clinical application of intraoperative frozen section analysis is the rapid discrimination between benign and malignant lesions, a determination that directly informs surgical strategy and patient management.[32] To evaluate CRISP's performance in this critical context, we validated the model using datasets from five independent centers (Fig. 2a,b), spanning six common anatomical sites (thyroid, lung, breast, brain, ovary, and head and neck), as well as complex abdominal lesion differentiation tasks (Supplementary Table S2.1).

CRISP was benchmarked against several state-of-the-art pathology foundation models[15-18,33] across 21 downstream diagnostic tasks (Supplementary Table S2.2). In malignancy discrimination, CRISP consistently achieved superior performance, with an average AUROC of 0.970, an average accuracy of 0.945, and an average F1 score of 0.924-significantly surpassing the next-best model (Fig. 2c; $P < 0.001$). Notably, CRISP was the only model to achieve a median F1 score above 0.9 (0.902; Fig. 2d).

Across all 21 downstream tasks, CRISP maintained consistent outperformance relative to other approaches (Fig. 2e, Supplementary Fig. S1a). In the subset of 14 tasks derived from previously unseen institutions, CRISP was the only model to achieve an average AUC above 0.95 (0.967±0.018), maintaining AUC values greater than 0.9 across all tasks. Even in particularly challenging datasets—such as NFH-Abdomen and ZSH-Abdomen—or in heterogeneous multi-institutional data (ZHDG-Breast, ZHG-H&N, ZSGZ-Brain), CRISP sustained an average AUC of 0.957, outperforming Virchow2 by 3.5% ($P < 0.001$). These findings underscore CRISP's robust predictive capability across diverse tissue types and laboratory environments. Detailed performance metrics for each downstream task are provided in Supplementary Tables S2.3-2.9. To further evaluate diagnostic sensitivity, we analyzed CRISP's performance in detecting malignant lesions across 1,729 cancer cases spanning 21 downstream tasks. The model correctly identified 1,634 cases, yielding an overall sensitivity of 94.5% (Supplementary Fig. S1b). At sensitivity thresholds of 90% and 95%, CRISP achieved median specificities of 0.917 and 0.790, respectively, again significantly outperforming the second-best model (+8.0% and +7.7% over Virchow2, $P = 0.004$ and $P = 0.013$; Fig. 2f).

Visualization analyses further demonstrated CRISP's ability to accurately localize malignant regions within frozen tissue sections, supporting its capacity for sensitive and precise cancer detection across a wide array of anatomical sites, institutions, and clinical contexts (Fig. 2g, Supplementary Fig. S2). In summary, these results establish CRISP as a robust and generalizable foundation model for intraoperative pathology, capable of delivering expert-level diagnostic support across diverse clinical scenarios and institutional settings.

**Model evaluation in key intraoperative decision-making tasks**

A central challenge for clinical-grade foundation models lies in their ability to generalize across the complex and dynamic decision-making scenarios encountered in real-world surgical practice. Intraoperative consultations inform every stage of the surgical workflow-from initial surgical strategy, through intraoperative guidance, to final assessment of resection completeness-making robust AI support essential for precision medicine.[3,34] To rigorously evaluate CRISP's generalizability and clinical utility, we established 20 multi-center downstream tasks spanning the entire surgical continuum: preoperative planning, intraoperative guidance, and postoperative assessment (Fig. 3a, Supplementary Table S3.1). These tasks were designed to reflect the diversity and complexity of real-world clinical decisions, with detailed clinical significance described in the Methods. Across both binary and multiclass tasks, CRISP consistently achieved the highest performance among all evaluated models (Supplementary Tables S3.2-3.3).

**Early intraoperative decision:** Accurate identification of primary central nervous system lymphoma (PCNSL) is critical for avoiding unnecessary resections and minimizing neurological morbidity.[35] In both the SYCC and NFH cohorts (Fig. 3b, Supplementary Fig. S3a), CRISP achieved outstanding discrimination, with AUCs of 0.998 and 0.987, respectively-representing statistically significant improvements over the second-performing comparator (both $P < 0.001$; Fig. 3b-3c, Supplementary Fig. S3b, Supplementary Table S3.4). Notably, CRISP correctly identified 64 of 66 patients with PCNSL, whereas alternative models missed up to 2.5 times as many cases, underscoring the clinical impact of enhanced diagnostic precision (Fig. 3d). In fine-grained tumor classification, CRISP demonstrated a pronounced 20.4% gain in top-1 accuracy in the external cohort (Fig. 3e, Supplementary Fig. S3c, Supplementary Table S3.5), highlighting its potential to support nuanced early intraoperative decision-making.

**Mid-procedure decision:** Defining the extent of resection is a pivotal intraoperative decision, particularly in complex cases such as ovarian, thyroid, and parathyroid surgery.[36-39] CRISP was evaluated across eight cohorts spanning ovarian (Ovarian Cancer Coarse/Fine-Grained Subtyping, OCCGS/OCFGS), thyroid (Thyroid Nodule Differential Diagnosis, TNDD), and parathyroid tissue (Parathyroid Gland Detection, PGD), consistently achieving the highest performance in all tasks, with a mean macro-AUC of 0.954—the only model to exceed 0.95, representing a 2.0% improvement over the second-best baseline ($P < 0.001$; Fig. 3f-3g, Supplementary Fig. S3d, Supplementary Tables S3.6-3.9). Confusion matrix analyses underscored CRISP's robust discrimination between distinct tissue types (Supplementary Fig. S4). For ovarian and thyroid tumors, where accurate subtype classification directly informs surgical strategy, CRISP showed the highest top-k (k = 1, 2, 3) accuracies in external cohorts, with top-1 accuracy improving by 13.6% over CONCH (Fig. 3h-j). In PGD, a particularly challenging task, CRISP was the only model to maintain AUCs consistently above 0.9. In external validation, it reached a specificity of 0.559 at 95% sensitivity, outperforming all other models and demonstrating potential to mitigate inadvertent parathyroid resection (Fig. 3k).

Similar results were observed in internal cohorts (Supplementary Fig. S3e-h), further supporting CRISP's robustness in fine-grained intraoperative decision-making.

**Late intraoperative assessment:** Accurate detection of residual tumor is critical for preventing postoperative recurrence and optimizing patient outcomes.[40] CRISP's performance was evaluated across eight cohorts focused on sentinel and cervical lymph node metastasis detection (SLMD, CLMD) and breast cancer surgical margin assessment (BSMA) (Fig. 4a).[41-43] The model achieved mean AUCs of 0.984, 0.992, and 0.992 across these tasks, outperforming all other evaluated models (Fig. 4b, Supplementary Tables S3.10-3.12). Under high-sensitivity conditions, CRISP demonstrated remarkable specificity: at 95% sensitivity, it reached 0.851 and 0.907 for SLMD and BSMA, respectively—improvements of 30.7% and 30.0% over the second-best model. For CLMD, specificity at 90% sensitivity increased to 0.962, an 18.9% gain. Among 263 positive patients, CRISP correctly identified 254 (96.6%), representing a 40% reduction in missed diagnoses compared to the second-best model (Fig. 4c, Supplementary Fig. S5). Visualization analyses further confirmed its ability to detect small clusters of residual tumor cells (Fig. 4d), supporting its utility in postoperative assessment.

Across the complete set of 20 downstream tasks, CRISP consistently outperformed all comparative models, which exhibited variable and dispersed performance (Fig. 4e). These findings collectively validate CRISP's generalizability and robust predictive power throughout the entire surgical decision-making process, establishing it as a highly effective tool for supporting precision intraoperative pathology in diverse clinical settings.

**Robustness under clinical heterogeneity**

A defining measure of clinical-grade foundation model is their capacity to generalize under data-scarce conditions, particularly when confronted with unobserved samples and rare cancer types-scenarios that frequently challenge conventional diagnostic approaches (Supplementary Table S4.1).[15,44] To rigorously evaluate this aspect, we conducted a large-scale pan-cancer detection study encompassing 13,851 intraoperative frozen slides from six independent centers (Fig. 5a). The CRISP model was deployed to extract patch embeddings from these slides, with the downstream aggregator trained exclusively on 1,573 slides from the SYCC cohort, representing six common anatomical sites: lung, breast, thyroid, head and neck, brain, and ovary (Supplementary Table S4.2).

Generalizability was systematically assessed across three external cohorts, which included both six observed and more than eleven previously unseen cancer types (Fig. 5b; Supplementary Tables S4.3-4.5). Across these diverse datasets, CRISP consistently achieved high diagnostic accuracy, with AUCs of 0.953, 0.973 and 0.960-each exceeding the 0.95 threshold and outperforming all comparators (Fig. 5c, Supplementary Table S4.6). Notably, CRISP showed pronounced advantages in the ZSH and HDGZ cohorts, with AUC improvements of 7.7% and 6.0% over the next-best model ($P < 0.001$), respectively, underscoring its robust cross-institutional generalizability.

To further dissect model performance, we analyzed results across 17 site-specific subgroups, encompassing both observed and unseen anatomical subtypes. CRISP consistently delivered the highest AUCs, ranging from 0.914 to 1.0 (Fig. 5d). While all models exhibited some decline in performance on unseen subgroups, CRISP maintained a distinct relative advantage, outperforming the next-best model by 10.4% in lymph nodes, 9.1% in cervix, 9.7% in bile duct, and 9.2% in pancreas. This suggests enhanced recognition and adaptability to unfamiliar cancer morphologies. Across 51 subgroup analyses from three unseen cohorts, CRISP achieved the top AUC in 47 and ranked within the top two in all, whereas Virchow2 led in only 16 unseen subgroups (Fig. 5e; Supplementary Tables S4.7-4.23), highlighting CRISP's superior adaptability to novel clinical scenarios.

CRISP's performance in rare tumors was equally compelling (Fig. 5f). According to the National Cancer Institute, rare cancers are defined as those with an annual incidence below 40,000 cases in the United States.[45,46] Across seven rare tumor types, CRISP achieved an overall AUC of 0.983, with individual cohort AUCs of 0.993, 0.974, 0.983, and 0.963. By contrast, the second-best model, Virchow2, yielded lower AUCs of 0.992, 0.916, 0.975, and 0.925, underscoring CRISP's superior generalizability, especially in unseen institutional settings. Furthermore, CRISP exhibited remarkable stability, maintaining AUCs above 0.9 in nearly all categories. In the ZSH-Else cohort, where all models performed less optimally (< 0.9), the results likely reflect the complexity and heterogeneity of tissue sources.

We further investigated the impact of unseen and rare samples on model performance (Fig. 5g, h). CRISP's diagnostic accuracy was minimally affected by either factor, and even when analysis was restricted to rare tumor subgroups from previously unseen anatomical sites, CRISP maintained an AUC exceeding 0.95. Under high-sensitivity conditions, CRISP also achieved markedly higher median specificity than the second-best model across both unseen and rare subgroups (Fig. 5i, j). Comprehensive performance metrics are detailed in Supplementary Tables S4.24-4.28. Collectively, these findings demonstrate CRISP's robust adaptability to complex and heterogeneous clinical data sources, including data-scarce scenarios and rare cancer types. This level of generalization is essential for real-world deployment, ensuring reliable diagnostic support across the full spectrum of intraoperative pathology.

**Prospective validation of clinical utility**

Prospective validation in real-world clinical environments is the definitive benchmark for assessing the practical utility of diagnostic AI models.[47] Building on retrospective findings, we conducted a large-scale prospective observational study at SYCC to rigorously evaluate CRISP's ability to support intraoperative decision-making under authentic surgical conditions. A total of 2,071 consecutive patients were enrolled, encompassing six clinically relevant downstream tasks: benign-malignant discrimination of pulmonary, thyroid, and breast nodules (PNBM, TNBM, BNBM), cervical and sentinel lymph node metastasis detection (CLMD, SLMD), and breast surgical margin assessment (BSMA). Importantly, the model's aggregators-trained exclusively on retrospective data-were deployed directly, without task-specific fine-tuning, providing an unbiased assessment of CRISP's robustness and translational potential (Fig. 6a).

Across all six real-world tasks, CRISP demonstrated exceptional diagnostic performance, achieving an overall AUC of 0.975 and outperforming all comparator models in every task (Fig. 6b, Supplementary Table S5.1). Compared to Virchow2, the second-best baseline, CRISP delivered AUC improvements ranging from 3.2% to 5.9% and elevated the specificity at 95% sensitivity from 0.495 to 0.809 (+63.4%), underscoring its clinical relevance in high-sensitivity scenarios (Supplementary Table S5.2). To further test resilience to laboratory variability, the cohort was stratified into 27 subgroups based on histotechnologists (Fig. 6c, Supplementary Table S5.3). CRISP maintained robust generalization, with 21 subgroups exceeding the overall AUC, 7 reaching perfect accuracy (AUC = 1.0), and all others—except the heterogeneous multi-doctor "Else" subgroup—remaining above 0.95. These findings highlight CRISP's ability to sustain high diagnostic precision across diverse, prospective, and heterogeneous clinical conditions.

We further evaluated CRISP's predictive accuracy in clinically challenging cases, defined by high diagnostic uncertainty in intraoperative frozen reports (n = 371). CRISP achieved an overall AUC of 0.936 in this subset (Fig. 6d, Supplementary Table S5.4), with performance largely maintained in PNBM, BNBM, and CLMD tasks. Modest decreases were observed in the TNBM ($-0.031$), SLMD ($-0.041$), and BSMA ($-0.047$) tasks, yet score distributions remained well separated between benign and malignant cases (Fig. 6e). In diagnostically elusive cases, such as sentinel lymph node micrometastases, CRISP correctly identified 7 of 8 micrometastatic and 40 of 41 macrometastatic cases (Fig. 6f), with predicted values showing clear separation from negative controls (Fig. 6g). Visualization analyses revealed that CRISP focused attention on peripheral zones-regions prone to metastatic involvement-and precisely highlighted minute malignant areas (Fig. 6h). These findings demonstrate CRISP's robust performance in difficult and easily overlooked cases, supporting its role as a diagnostic aid for pathologists.

To explore CRISP's potential to alleviate clinical workload, we examined its utility in triaging cases requiring immunohistochemistry (IHC), a common but resource-intensive diagnostic adjunct. Among 510 patients requiring additional IHC evaluation, CRISP achieved an overall AUC of 0.963, with task-specific AUCs ranging from 0.943 to 1.000 (Fig. 6i, Supplementary Table S5.5). At 95% sensitivity, CRISP correctly identified 74.5% of IHC-negative patients, underscoring its value as a screening tool to reduce unnecessary testing (Fig. 6j). We further assessed whether CRISP could safely triage typical cases without compromising pathologists' diagnostic accuracy.[48] When patients with predicted scores > 0.9956 or < 0.0044 were managed directly by CRISP, the combined CRISP-pathologist workflow reduced diagnostic workload by 34.9%, with 79.0% of positive cases included (Fig. 6k). Under this strategy, sensitivity and specificity remained comparable to conventional frozen section reports (sensitivity, $P = 0.60$; specificity, $P = 1.00$; Fig. 6l), highlighting CRISP's potential as a clinical prescreening tool for streamlining the diagnosis of typical positive cases. Together, these prospective findings demonstrate that CRISP not only generalizes effectively to real-world clinical settings but also delivers tangible improvements in diagnostic accuracy, efficiency, and workflow optimization. This establishes CRISP as a clinically valuable tool for intraoperative pathology, with broad implications for enhancing surgical decision-making and patient care.

**Discussion**

Foundation models have redefined the landscape of artificial intelligence, but their translation into high-stakes clinical environments has remained elusive.[49,50] Nowhere is this gap more consequential than in surgery, where more than 310 million major surgical procedures are performed worldwide each year, and intraoperative decisions must balance speed with diagnostic precision.[51,52] The consequences of diagnostic uncertainty are profound, leading to unnecessary reoperations, overtreatment, or undertreatment, and imposing substantial economic burdens that exceed one billion dollars annually in the United States alone.[53] Against this backdrop, the development of a clinically reliable AI platform capable of guiding intraoperative decisions represents a paradigm shift-not only for surgical pathology, but also for the broader clinical adoption of foundation models in medicine.

In this study, we present CRISP, the first clinical-grade foundation model specifically dedicated to intraoperative pathology. Distinct from existing pathological foundation models trained on FFPE specimens, CRISP leverages large-scale frozen section data for pre-training. This strategic focus directly addresses a critical clinical gap: FFPE-trained models often falter in intraoperative settings due to frozen section-specific artifacts such as ice crystal formation and tissue compression.[25] By constructing a large and diverse dataset sourced from eight independent medical centers and reserving data from five centers exclusively for external validation, we ensured rigorous assessment of CRISP's diagnostic performance under authentic, heterogeneous clinical conditions. Furthermore, prospective real-time deployment provided a robust test of the model's reliability in actual surgical workflows.

CRISP's robust performance across a spectrum of intraoperative diagnostic tasks highlights its potential as a generalizable paradigm for frozen section-based decision support. Beyond achieving superior accuracy in benchmark comparisons, CRISP demonstrated the ability to address clinically critical scenarios-such as the detection of subtle malignant foci in surgical margins or lymph nodes-where diagnostic sensitivity is paramount to avoid undertreatment.[54,55]

Importantly, CRISP maintained consistent advantages across the entire surgical workflow, from early intraoperative differential diagnosis to mid-procedure tumor classification and late assessment of resection completeness. Given that misinterpretation of intraoperative frozen sections can lead to inappropriate surgical decisions and increased patient risk,[56,57] CRISP's breadth of applicability suggests it could substantially reduce variability and time constraints, offering a scalable and clinically meaningful complement to routine pathology practice.

A key strength of CRISP lies in its capacity to learn generalized morphological features, enabling robust generalization even under previously unseen laboratory conditions.[58] The model maintained high predictive accuracy for tissue types with scarce or absent training samples-a crucial benchmark for foundation model generalizability. When applied to previously unseen institutional data and novel samples, CRISP-based pan-cancer detection models demonstrated stable and reliable performance, outperforming alternative model by up to 11% on challenging datasets. This provides compelling evidence that CRISP has effectively internalized universal tissue morphology from large-scale frozen sections, equipping it to navigate the complexity of diverse clinical scenarios. Notably, CRISP also excelled in the diagnosis of rare diseases, addressing a common challenge in clinical practice where labeled data are limited and expertise may be lacking.[59-61] Such capabilities position CRISP as a valuable tool for resource-limited settings, potentially mitigating disparities in healthcare access and supporting expert-level pathology where it is most needed.[62]

Prospective evaluation further reinforced CRISP's feasibility as a clinical-grade tool. When deployed in real-time intraoperative settings, CRISP demonstrated robust performance across diverse patient cohorts and laboratory conditions, bridging the gap between experimental development and practical implementation. Remarkably, CRISP alone enabled precise surgical intervention in 92.6% of patients. In contexts where clinical resources are limited, the ability to optimize resource allocation becomes essential.[63,64] In routine cases, CRISP-based triage reduced pathologists' workload by 35% and prevented 75% of unnecessary ancillary testing, while in complex cases, it provided a standardized reference to support diagnostic reconsideration and reduce missed diagnoses. These results highlight CRISP's potential for seamless integration into intraoperative diagnostic workflows, with tangible benefits for both clinical efficiency and patient safety.

Despite these promising findings, several limitations warrant consideration. First, while extensive retrospective and prospective validation was performed on nearly 20,000 multi-center intraoperative frozen sections and over 2,000 prospective multi-site cases, a fully prospective interventional study is still necessary to assess its real-world deployment and impact on clinical outcomes.[65,66] Second, although CRISP demonstrated strong generalization across multiple centers, further validation in more diverse ethnic and geographic populations is essential to ensure global applicability.[67,68] Finally, large-scale deployment and productization will require addressing technical and operational challenges common to AI clinical translation, including integration with existing workflows, regulatory compliance, and user training.[69]

In summary, CRISP establishes a versatile and clinically relevant AI paradigm for intraoperative pathology, illustrating the transformative potential of large-scale, data-driven approaches to advance precision surgery. By bridging the gap between technological innovation and clinical practice, CRISP paves the way for the next generation of AI-enabled surgical decision support, with broad implications for improving patient outcomes and healthcare delivery worldwide.

## Methods

### Ethical approval

This study was approved by the Ethics Committee of Sun Yat-sen University Cancer Center (Approval No. SL-B2024-708-03) and carried out in compliance with the Declaration of Helsinki. Informed consent was obtained from all subjects enrolled in the prospective cohort, while the requirement for informed consent was waived for retrospective cohorts. The trial was prospectively registered in the Chinese Clinical Trial Registry (ChiCTR2500106350).

### Large-scale intraoperative frozen section dataset

To develop a clinical-grade foundation model for intraoperative pathology, we constructed a large-scale dataset of frozen section whole-slide images (WSIs), comprising 101,796 slides from 50,093 patients. Retrospective data were collected from eight independent medical centers, including Sun Yat-sen University Cancer Center (SYCC), Zhujiang Hospital, Southern Medical University (ZJH), Jiangmen Central Hospital (JMH), Zhongshan City People's Hospital (ZSH),

Nanfang Hospital, Southern Medical University (NFH), Affiliated Cancer Hospital & Institute of Guangzhou Medical University (GZH), Huizhou Central People's Hospital (HZH), and The Tenth Affiliated Hospital, Southern Medical University, Dongguan People's Hospital (DGH). In total, these cohorts encompassed 38,253 patients and 81,678 intraoperative frozen WSIs collected between May 2017 and July 2024. The following inclusion criteria were applied to all cases: (1) at least one available intraoperative frozen section; (2) presence of identifiable tissue on the slide; (3) absence of scanning artifacts such as defocus or blurring; and (4) availability of complete diagnostic and clinical information. The detailed inclusion period, number of patients, number of slides, and baseline characteristics of each cohort are summarized in Supplementary Tables S1.1-1.8. To further increase histological diversity, we additionally curated 15,975 frozen WSIs from The Cancer Genome Atlas (TCGA). The detailed selection process was as follows: diagnostic slides prepared from formalin-fixed paraffin-embedded (FFPE) tissues were excluded, resulting in 21,094 slides. These were subsequently reviewed by two experienced pathologists, and immunohistochemistry as well as other non-frozen section slides were removed. To ensure consistency, all patients were categorized into 24 major anatomical sites according to the definitions provided by the Genomic Data Commons (GDC) system.[70] For model development, WSIs from ZJH, JMH, TCGA, and a subset of SYCC (43,013 slides) were used for CRISP pretraining. The remaining SYCC data (5,178 slides) were reserved for downstream task development and internal evaluation, ensuring no data leakage between pretraining and testing. External validation was performed using ZSH, NFH, GZH, HZH, and DGH cohorts.

For the prospective real-world deployment study of CRISP, patients undergoing thyroid, breast, or lung surgery with intraoperative consultation at Sun Yat-sen University Cancer Center were recruited between November 2024 and July 2025, representing the most common sites for intraoperative consultation. Inclusion criteria were: (1) at least one available intraoperative frozen section; (2) presence of identifiable tissue on the slide; (3) absence of scanning artifacts such as defocus or blurring; (4) complete diagnostic and clinical information; (5) age ≥18 years at the time of surgery; and (6) documented informed consent. Following screening, a total of 2,071 patients with 4,143 intraoperative frozen sections were included. Based on the tissue site, patients were assigned to three prospective cohorts: PCS-THYR (n=1,330), PCS-BREA

(n=391), and PCS-LUNG (n=350). Recruitment periods and detailed clinical characteristics for each cohort are provided in Supplementary Tables S1.9-1.11.

**Data digitization and preprocessing**

All clinical intraoperative frozen sections included in this study were digitized at 40× magnification (0.25 μm/pixel). Slides from the SYCC cohort were primarily scanned using an Aperio AT2 scanner (Leica Biosystems, Wetzlar, Germany) and stored in SVS format. Data from the ZJH, GZH, HZH, and DGH cohorts were scanned with a PHILIPS Ultra-Fast Scanner (Philips Electronics N.V., Amsterdam, Netherlands) and saved in iSyntax format. NFH and ZSH cohort slides were mainly scanned using a KF-PRO-120-HI scanner (Jiangfeng Biotech, Ningbo, China) and stored in KFB format. JMH cohort slides were scanned using an SQS-600P scanner (Shengqiang Technology Co., Ltd., Shenzhen, China) and saved in SDPC format. All slides from the PCS cohort were scanned with a PHILIPS Ultra-Fast Scanner (Philips Electronics N.V., Amsterdam, Netherlands) and saved in iSyntax format.

WSIs were processed using vendor-provided API tools to extract full-slide thumbnail images. Given the extensive presence of non-informative background areas, tissue regions were automatically segmented using the CLAM toolkit to identify and isolate relevant histological content for subsequent analysis.[61] Segmented tissue regions were partitioned into a series of non-overlapping patches of 512×512 pixels at a resolution of 0.25 microns per pixel (MPP). To mitigate center-specific biases introduced by variations in optical resolution, slide preparation, and scanning settings, redundant patches from higher-resolution images were downsampled. This approach ensured a balanced representation of patches from each center, reducing the risk of model overfitting to specific institutional characteristics. In total, approximately 50 million patches were extracted from 81,667 slides for model pretraining.

**Network architecture**

Building on previous studies of FFPE-based pathology foundation models, we focus here on intraoperative frozen sections and introduce CRISP, a specialized pathology foundation model for their analysis. CRISP is built upon Virchow2, a state-of-the-art pathology foundation model pretrained on 1 million hematoxylin and eosin (H&E)-stained WSIs.[15,33] To adapt this general-purpose model to the frozen section domain while preserving its learned representations and generalizability, we employed low-rank adaptation (LoRA).[71] LoRA is a parameter-efficient fine-tuning (PEFT) method that injects trainable rank decomposition matrices into the self-attention layers of the transformer architecture, keeping the remaining model weights frozen during training. This approach drastically reduces the number of trainable parameters, computational overhead, and risk of catastrophic forgetting. In this study, LoRA modules were integrated into the query and value projection matrices of each attention layer, initialized with a rank of 8 and a scaling factor of 16. This configuration resulted in only 2 million trainable parameters, constituting merely 0.3% of the model's total 633 million parameters.

**Model training**

For pretraining, we utilized DINO, a self-supervised learning strategy effective for learning visual representations without labeled data.[72] Specifically, DINO operates on a teacher-student paradigm in which two different augmented views of the same image are processed by the student and teacher networks, respectively. The student network learns to predict the output distribution of the teacher network, thereby encouraging the learning of semantically meaningful features without manual annotation. The teacher network's weights are updated via an exponential moving average (EMA) of the student weights, with a momentum coefficient set to 0.9995. The model was trained using the AdamW optimizer ($\beta_1$= 0.9, $\beta_2$= 0.999) with mixed-precision (FP16) arithmetic to accelerate training and reduce memory consumption. Pretraining was conducted using 8 NVIDIA H800 GPUs with a batch size of 384 for 640,000 iterations.

For downstream intraoperative tasks, a slide-level prediction pipeline was implemented. In computational pathology, WSIs are typically too large to be processed directly by deep learning models. Therefore, the analysis pipeline was divided into three steps: patch-level feature extraction, feature aggregation, and slide-level prediction. In this study, each 512×512 patch (at 0.25 MPP) was resized to 224×224 pixels to serve as input to the frozen CRISP encoder. From the final layer of the transformer, the [CLS] token embedding and the mean-pooled embeddings of all patch tokens were concatenated, forming a comprehensive 2,560-dimensional feature vector for each patch. To identify diagnostically critical regions, we employed an attention-based multiple instance learning (ABMIL) model as the feature aggregator.[73] The ABMIL module learns to assign an attention weight to each patch feature vector, effectively highlighting salient patches and generating a weighted slide-level feature representation. This aggregated feature vector was then passed through a single fully connected classification layer to produce the final slide-level prediction. For supervised downstream tasks, the cross-entropy loss function was used. Only the ABMIL aggregator and the classification layer were trained from scratch, while the CRISP encoder remained frozen to leverage the pretrained features. The model was optimized using the Adam optimizer with a learning rate of $2\times10^{-4}$ and weight decay of $1\times10^{-5}$. Early stopping based on validation loss was implemented to halt training upon performance saturation and prevent overfitting. In all downstream task comparisons, identical strategies were employed across comparator models to guarantee fairness in performance assessment.

**Downstream task**

To comprehensively evaluate CRISP, 98 retrospective downstream tasks were established. The reference labels for all tasks were based on postoperative FFPE morphology integrated with immunohistochemical and molecular analyses, and were independently verified by a pathologist. Aggregator development for all downstream tasks was conducted using the SYCC cohort, with patients in the internal test sets kept independent from the training and validation sets to prevent data leakage.

**Classification of common benign and malignant lesions:** Each benign-malignant classification task was formulated as an independent slide-level binary problem, with slides containing malignant regions designated as positive. Six prevalent tissue sites were included: lung, breast, thyroid, head and neck, brain, and ovary. The task encompassed commonly encountered benign and malignant lesions, such as pulmonary adenocarcinoma and inflammatory nodules, to replicate large-scale clinical pre-screening scenarios. The abdominal metastasis classification task was incorporated to evaluate CRISP's capability in distinguishing complex metastatic lesions. Abdominal metastases represent a frequent dissemination route for multiple cancers and often signal disease progression; accurate identification can therefore inform patient management and improve clinical outcomes.[74,75] Malignant abdominal nodules comprised diverse metastatic cancers, including ovarian, intestinal, and gastric origins. Model performance was assessed on 2,771 slides across five external cohorts, with dataset partitions detailed in Supplementary Table S2.1.

**Early intraoperative decision:** The primary central nervous system lymphoma differentiation (PCNSLD) task was defined as a slide-level binary classification, aimed at accurately distinguishing primary central nervous system lymphoma (PCNSL) from other brain lesions. PCNSL is recognized as one of the most common primary malignant brain tumors, exhibiting poor prognosis, presenting diagnostic challenges, and being primarily managed with radiotherapy and chemotherapy, in contrast to other brain lesions.[76] Accurate intraoperative identification of PCNSL can prevent unnecessary surgical interventions and facilitate precision treatment. Slides containing PCNSL were labeled positive, whereas slides of other common brain lesions—including gliomas, medulloblastomas, central neurocytomas, metastases, and inflammatory lesions—were labeled negative. External validation was performed using 461 slides from the NFH cohort. The central nervous system lesion classification (CNSLC) task comprised the same dataset but was formulated as a fine-grained multi-class classification to distinguish among the aforementioned lesion types. Detailed dataset partitions for intraoperative decision tasks are provided in Supplementary Table S3.1.

**Mid-procedure decision:** The ovarian cancer coarse-grained subtyping (OCCGS) task involved slide-level three-class classification to differentiate primary ovarian carcinomas, metastatic ovarian tumors, and benign ovarian lesions. As the ovary is a common site of metastasis from extragenital malignancies, distinguishing primary from metastatic tumors remains a critical diagnostic challenge during surgery.[37,77] Accurate classification informs the surgical extent and guides additional treatment directed at the primary site. The ovarian cancer fine-grained subtyping (OCCFS) task shared the same dataset but focused on fine-grained classification of primary ovarian carcinoma subtypes, including serous, clear cell, and endometrioid adenocarcinomas.

The thyroid nodule differential diagnosis (TNDD) task involved a slide-level three-class classification of thyroid nodules into papillary thyroid carcinoma, follicular-patterned tumors, and benign lesions such as nodular goiter. Frozen section diagnosis of thyroid nodules is critical for intraoperative surgical decision-making, as papillary carcinoma and nodular goiter often dictate different surgical extents, while follicular-patterned tumors typically require diagnostic surgery pending definitive FFPE evaluation.[78]

The parathyroid gland detection (PGD) task was a slide-level binary classification for detecting parathyroid tissue in thyroidectomy frozen sections. Inadvertent parathyroid resection, occurring in up to 16% of thyroidectomies, is the primary cause of postoperative hypocalcemia.[79,80] Accurate frozen section recognition of parathyroid tissue is therefore essential for preventing misresection and optimizing surgical outcomes. Labels were assigned solely based on the presence of parathyroid tissue, with all other tissues—including thyroid, lymph nodes, and thyroid carcinoma—classified as negative, reflecting real-world intraoperative settings.

**Late intraoperative assessment:** Sentinel lymph node metastasis detection (SLMD), cervical lymph node metastasis detection (CLMD), and breast cancer surgical margin assessment (BSMA) are three slide-level binary classification tasks designed to detect residual tumor cells in lymph nodes and surgical margins. Sentinel lymph node assessment and margin evaluation are critical indicators of surgical completeness and directly influence patient prognosis and risk of recurrence.[81,82] In thyroid cancer, accurate intraoperative identification of central lymph node metastasis can help avoid unnecessary dissections, whereas in breast cancer, margin and sentinel lymph node assessment form the basis of individualized treatment.[13,83,84]

**Pan-cancer detection study:** The pan-cancer detection task was designed to develop a universal model capable of accurately identifying malignant lesions across intraoperative frozen samples of any source or type, thereby evaluating CRISP performance under data-limited conditions, particularly in unseen samples and rare cancers. Model development was based on 3,071 benign and malignant WSIs from six common frozen tissue sites in the SYCC cohort. Given that clinical decision-making is case-based, external validation was conducted at the case level. For each case, positivity was defined as the presence of at least one malignant slide. All slide-level features from a case were concatenated and directly input into the aggregator as case-level features. The model was validated across five external cohorts comprising 5,867 cases: ZSH and NFH served as two independent cohorts, whereas GZH, HZH, and DGH were combined into a single cohort. Detailed dataset partitions are provided in Supplementary Tables S4.2-4.5. For subgroup analyses, eight tissue sites (stomach, bone, esophagus, kidney, bladder, adrenal gland, testis, and thymus) were combined as "Else", owing to limited sample sizes or unidirectional labels in external cohorts, which precluded reliable calculation of metrics such as AUROC.

**Prospective cohort study**

In the prospective cohort, the diagnostic reference standard for each patient was established based on the postoperative FFPE pathology report in combination with immunohistochemistry and molecular test results, and independently verified by a board-certified pathologist blinded to the study. Patients were assigned to one of six intraoperative decision tasks (PNBM, TNBM, BNBM, CLMD, SLMD, BSMA) according to the surgical indication and sample type. Detailed dataset partitions are provided in Figure 2b. Aggregators used for validation in these six tasks were directly transferred from retrospective downstream task construction without any task-specific fine-tuning. For clinical applicability, validation was performed at the patient level.

Information on frozen section slide preparation personnel was obtained from intraoperative reports, encompassing a total of 30 histotechnologists; four with small case numbers and homogeneous labels were combined into an "Else" category for subgroup analyses. Diagnostic outcomes and confidence levels were derived from the content of the intraoperative frozen section reports. Confidence assessment was conducted according to a prespecified rule set: reports indicating "consistent with", "consider", or "favor" were categorized as relatively definitive, whereas reports stating "not excluded", "suspicious for", or "requires differential diagnosis" were categorized as indeterminate. All classifications were independently reviewed and confirmed by a pathologist blinded to the study. Immunohistochemistry requirements were similarly inferred from the report content.

Micro-metastases and macro-metastases were defined according to AJCC guidelines and determined based on both intraoperative frozen section and postoperative FFPE pathology reports, and were independently morphologically reviewed by a pathologist blinded to the study. Micro-metastases were defined as malignant foci within sentinel lymph nodes measuring greater than 0.2 mm but not exceeding 2 mm, whereas macro-metastases were defined as malignant foci larger than 2 mm.[85]

**Statistical analysis**

For binary classification tasks, model performance was evaluated using the area under the receiver operating characteristic curve (AUROC), F1 score, accuracy, and specificity at 90% and 95% sensitivity thresholds. For multi-class classification tasks, performance was assessed using macro-averaged AUROC and top-k accuracy. To compare overall model performance across multiple tasks, paired Wilcoxon signed-rank tests were performed, and p-values were adjusted for multiple comparisons using the Holm method.[86] Two-sided 95% confidence intervals (CIs) were computed via bootstrap with 1,000 iterations. Differences in predicted value distributions for multi-class tasks were assessed using one-way analysis of variance (ANOVA) followed by Tukey's honestly significant difference (HSD) test for pairwise comparisons. For binary tasks, group differences in predicted score distributions were evaluated using two-sided Student's t-tests; non-parametric alternatives were considered when normality assumptions were not met. Statistical significance was defined as not significant ($P > 0.05$), * ($P \leq 0.05$), ** ($P \leq 0.01$) and *** ($P \leq 0.001$). Boxplots depict the median (50th percentile), interquartile range (25th-75th percentile), and whiskers representing 1.5 × interquartile range. Data analysis was performed using R version 4.2.1 (2022-06-23) and Python 3.13.5, and deep learning models were implemented in PyTorch 2.8.0.

## Acknowledgements

This work was supported by grants from the National Natural Science Foundation of China (grant No. 82172646, 82202905, 82373190, 62202403 and 82403737), Natural Science Foundation of Guangdong Province (No. 2024A1515012448), Science and Technology Program of Guangzhou (No. 2025A04J3599), China Postdoctoral Science Foundation (No. 2024M753786), Chih Kuang Scholarship for Outstanding Young Physician- Scientists of Sun Yat-sen University Cancer Center (No. CKS-SYSUCC-2023005), ), Hong Kong Innovation and Technology Commission (Project No. MHP/002/22 and ITCPD/17-9), Research Grants Council of the Hong Kong Special Administrative Region, China (Project No. R6003-22 and C4024-22GF) and Young Talents Program of Sun Yat-sen University Cancer Center (No. YTP-SYSUCC-5260025).

## Author contributions

Z.H.Z., F.T.Z., Y.P.Y., B.C., R.G.L., Y.W., H.C. and M.Y.C. conceived the study and designed the experiments. Z.H.Z., Y.P.Y., B.C., X.K.Z., X.B.W., J.W.C, C.Y.Z., J.Y.Z., Y.Q.W., J.J.M., Z.Q.Z., X.Q.L., J.L., R.G.L. and Y.W. collected the data for self-supervised learning. F.T.Z. and J.B.M. performed model development for self-supervised learning. Z.H.Z., F.T.Z., X.K.Z. and M.Y.C. organized the datasets and codebases for all downstream tasks. K.Z., X.Y.Z., D.X., and M.Y.C. performed quality control of the codebase and the results. Z.H.Z., F.T.Z., H.C. and M.Y.C. prepared the paper with input from all co-authors. H.C. and M.Y.C. supervised the research.

## Conflict-of-interest disclosure

The authors declare no competing interests.

## Code and data availability

The source code will be available upon publication. The TCGA whole-slide images and associated clinical data can be accessed through the NIH Genomic Data Commons (https://portal.gdc.cancer.gov). Whole-slide images and clinical data from the internal and external cohorts are not publicly available owing to institutional policies and patient privacy regulations, but were collected under Institutional Review Board approvals. Access to these datasets for non-commercial academic use may be requested from the corresponding author. All other data supporting the findings of this study are available within the article and its Supplementary Information files.

# Figures

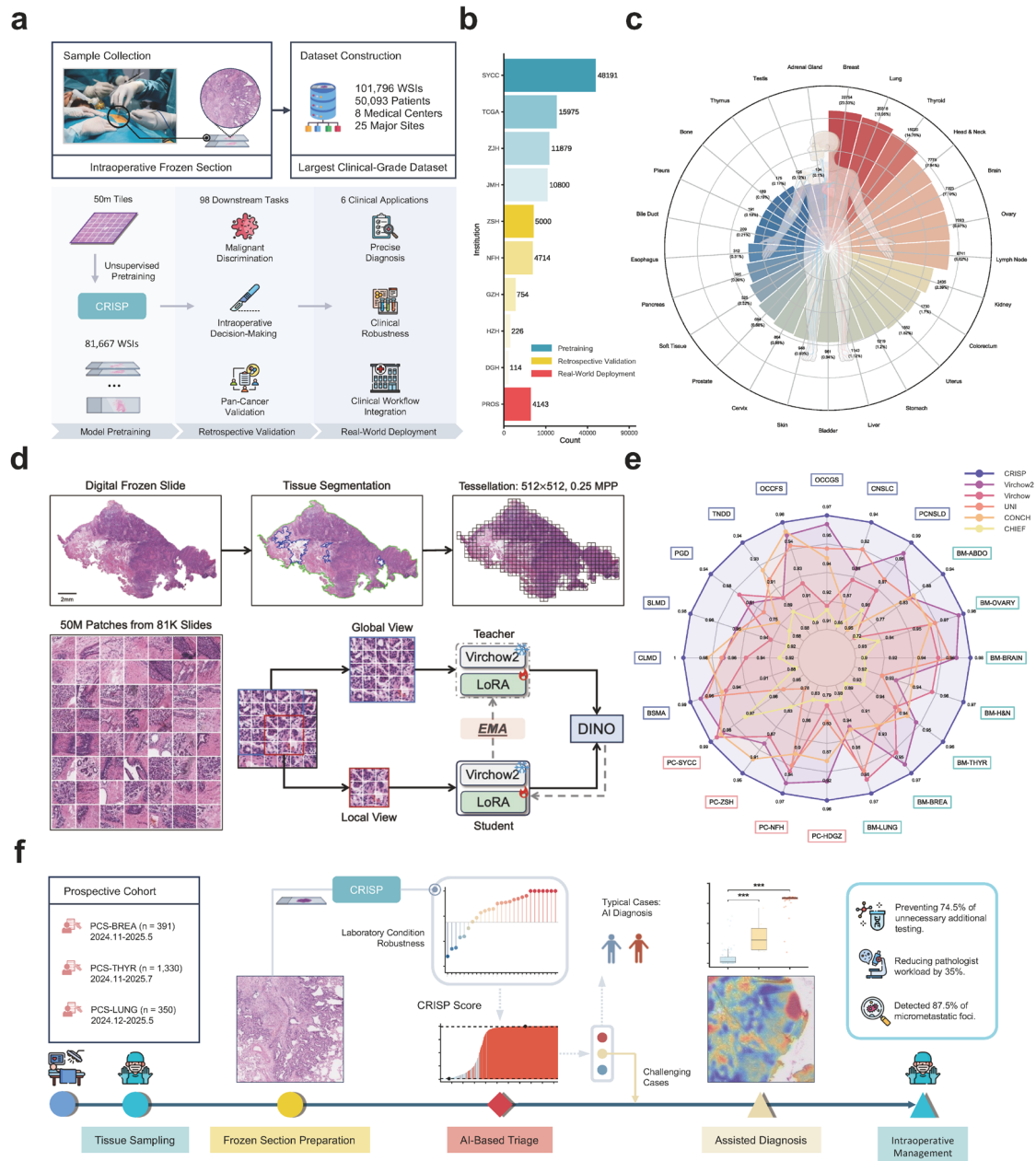

**Fig. 1. Study Cohort Characteristics and Workflow.**
(A) Dataset composition and study design. A clinical-grade frozen section dataset comprising 101,796 slides was constructed to train and validate CRISP, followed by real-world deployment for prospective evaluation. (B) Data distribution across participating medical centers. (C) Distribution of slides from 25 major anatomical sites. (D) Overview of the CRISP pretraining strategy. (E) Performance comparison of CRISP with state-of-the-art pathology foundation models on retrospective downstream tasks. (F) Workflow of CRISP in real intraoperative settings. In a prospective cohort of over 2,000 patients, CRISP delivered robust predictions on intraoperative frozen sections, reduced pathologists' workload, and provided standardized references for challenging cases.

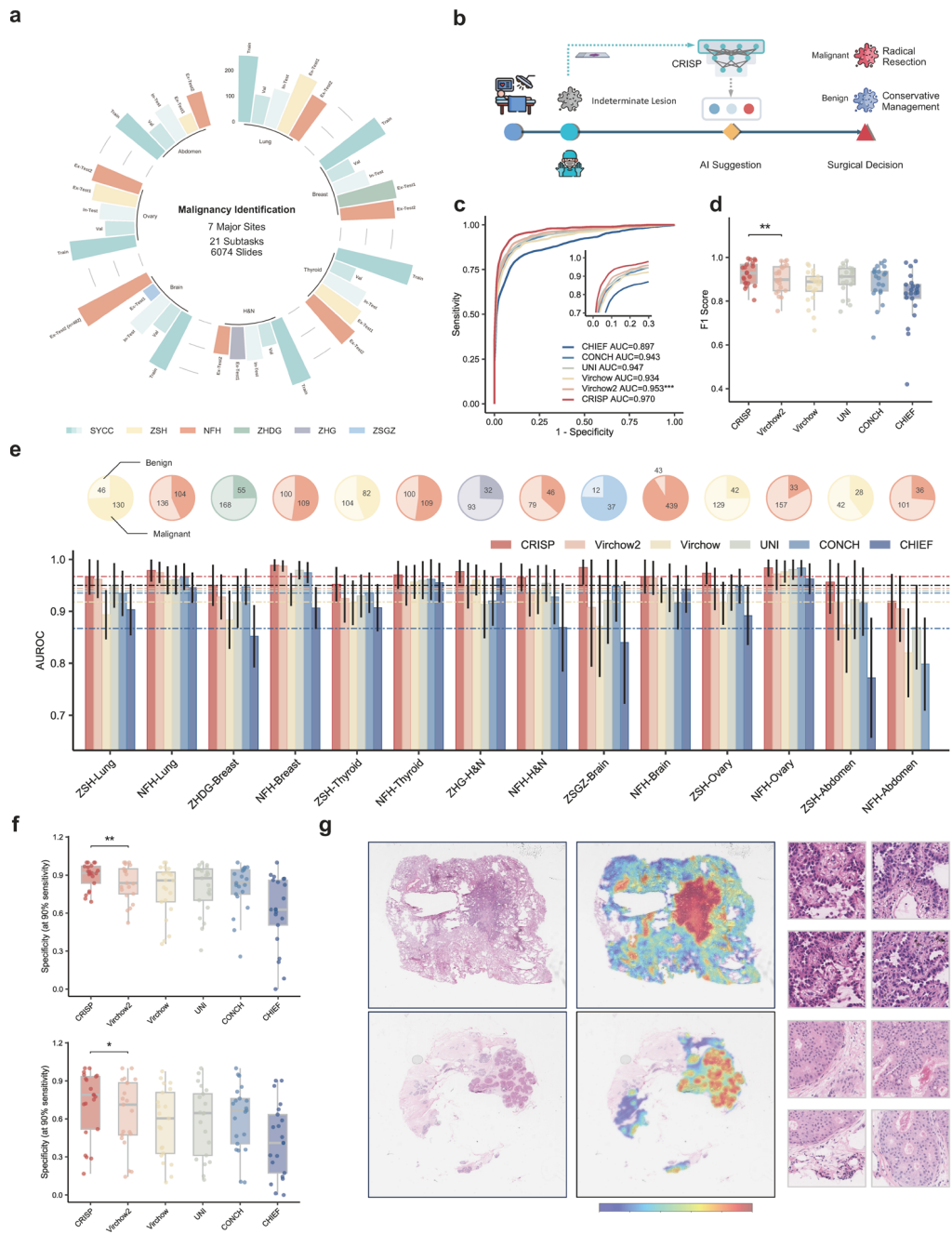

**Fig. 2. Performance of CRISP in common benign–malignant differentiation tasks.**
(A) Task design and dataset utilization for common benign–malignant differentiation tasks. (B) Clinical intraoperative application scenario, where CRISP's diagnosis of lesion nature directly determines the surgical approach. (C) Comparison of average AUROC across tasks between CRISP and other foundation models. (D) Comparison of average F1 scores across tasks between CRISP and other foundation models. (E) Composition of external validation cohorts across anatomical sites (top) and AUROC comparison between CRISP and other models (bottom); error bars denote 95% confidence intervals. (F) Specificity comparison of CRISP and other models at 90% sensitivity (top) and 95% sensitivity (bottom) across tasks. (G) Attention visualization of CRISP predictions on malignant lung (top) and breast (bottom) WSIs, showing from left to right the original WSI, CRISP heatmap, and the most attended regions. ZHDG, ZSH&HZH&DGH&GZH; ZHG, ZSH&HZH&GZH; ZSGZ, ZSH&GZH.

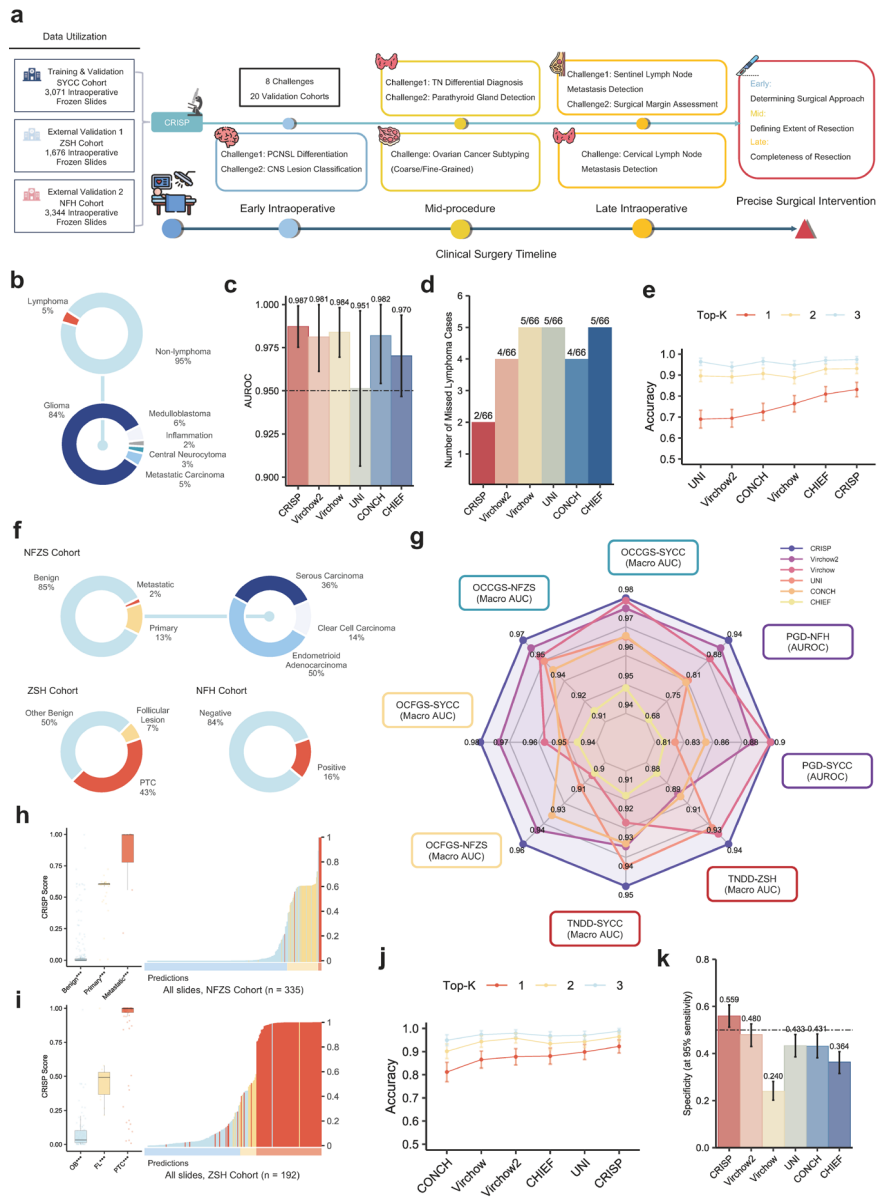

**Fig. 3. Evaluation of CRISP for intraoperative decision-making.**
(A) Task design and dataset utilization for full-process intraoperative decision-making. (B) Data composition of the external validation cohort for the PCNSLD task (top) and CNSLC task (bottom). (C) Comparison of AUROC between CRISP and other models on external validation cohorts for the PCNSLD task. (D) Number of lymphoma cases missed by each model in the PCNSLD task. (E) Top-k accuracy comparison of CRISP and other models in external validation cohorts for the CNSLC task. (F) Data composition of external validation cohorts for the OCCGS, OCFGS, TNDD, and PGD tasks. (G) Performance comparison of CRISP and other models across eight PID-procedure decision tasks; Macro-AUC was used for multiclass tasks, and AUROC for binary tasks. (H) Predicted probability distributions of CRISP across classes in the OCCGS task in the external validation cohort. (I) Predicted probability distributions of CRISP across classes in the TNDD task in the external validation cohort. (J) Top-k accuracy comparison of CRISP and other models in the external validation cohort for the OCFGS task. (K) Specificity comparison of CRISP and other models at 95% sensitivity for the PGD task in the external validation cohort. NFZS, NFH&ZSH; OB, Other Benign; FL, Follicular Lesion; PTC, Papillary Thyroid Carcinoma.

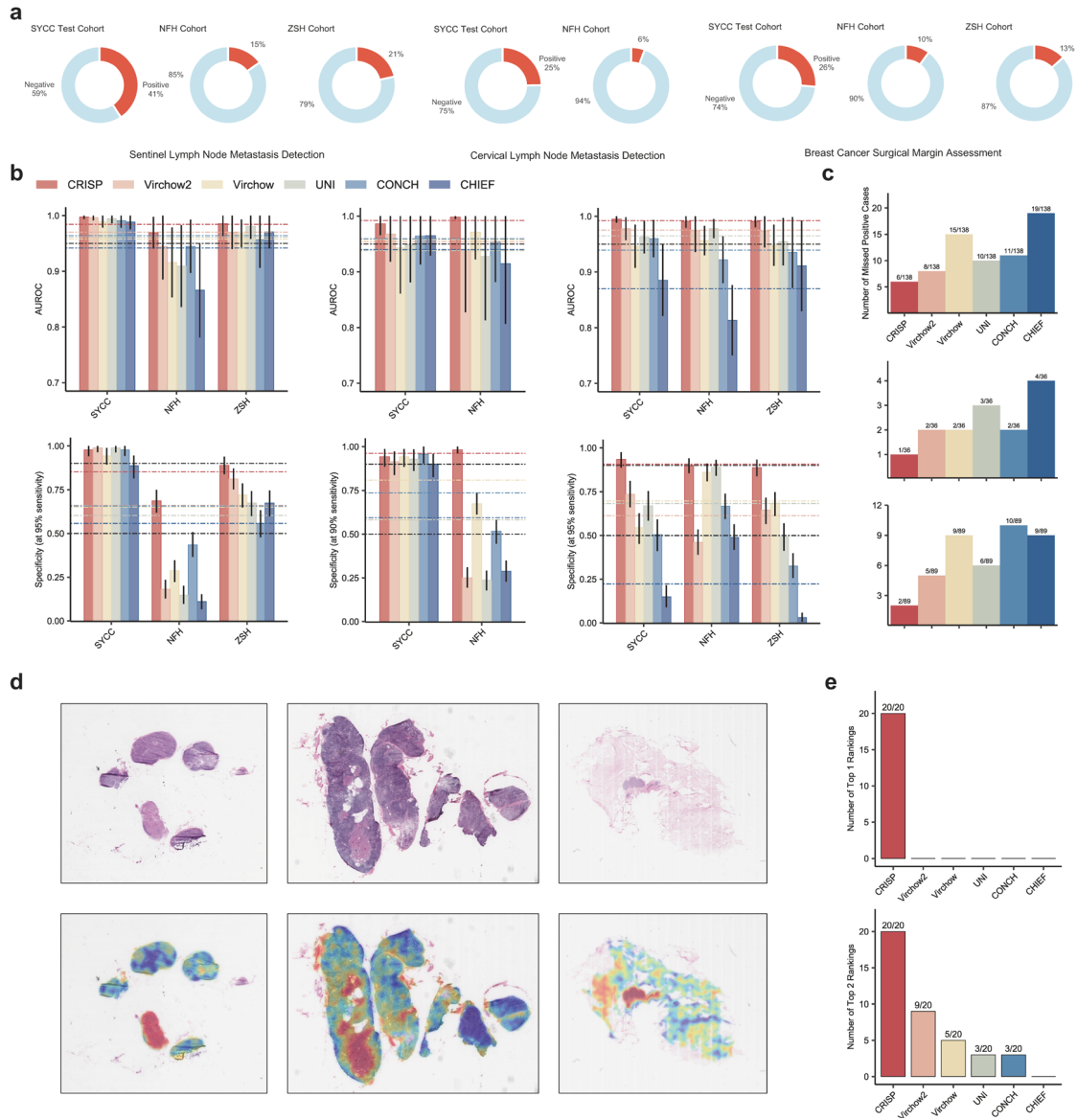

**Fig. 4. Performance of CRISP in late intraoperative assessment.**
(A) Data composition of downstream tasks in late intraoperative assessment. (B) Performance comparison of CRISP and other models across cohorts for the SLMD (left), CLMD (middle), and BSMA (right) tasks, evaluated by AUROC (top) and specificity at 95% sensitivity (bottom). (C) Comparison of the total number of positive cases missed by each model in the SLMD (top), CLMD (middle), and BSMA (bottom) tasks. (D) Representative positive frozen sections (top) and CRISP attention visualizations (bottom) in the SLMD (left), CLMD (middle), and BSMA (right) tasks. (E) Number of tasks among 20 full-process intraoperative decision-making tasks where each model achieved top-1 (top) or top-2 (bottom) AUROC.

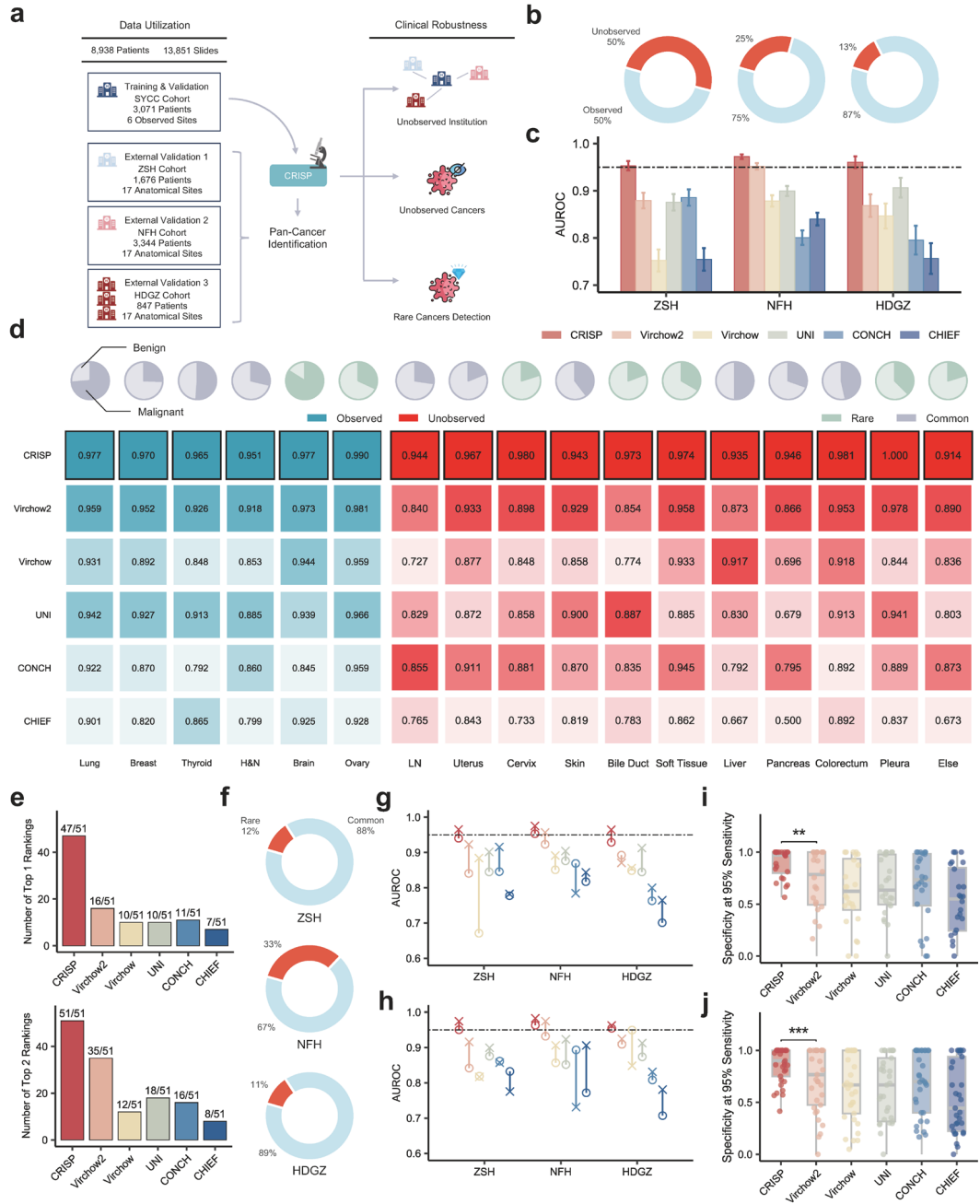

**Fig. 5. Pan-cancer detection performance of CRISP.**

(A) Data utilization and experimental design of the pan-cancer detection task. (B) Data composition of the external validation cohorts. (C) Overall AUROC comparison across external validation cohorts. (D) Data composition (top) and AUROC comparison (bottom) across different tissue sites; in the heatmap, deeper color indicates better-performing models. (E) Number of tasks (out of 51) where each model achieved top-1 (top) or top-2 (bottom) AUROC in external validation. (F) Proportion of rare cancers in the external validation cohorts. (G) AUROC comparison between observed and unobserved tissue sites across models; crosses denote observed sites, circles denote unobserved origins. (H) AUROC comparison in rare cancers across models; crosses denote all rare cancers, circles denote the subgroup of rare cancers from unobserved sites. (I) Comparison of specificity at 95% sensitivity in unobserved tissue sites. (J) Comparison of specificity at 95% sensitivity in rare cancers. HDGZ, HZH&DGH&GZH.

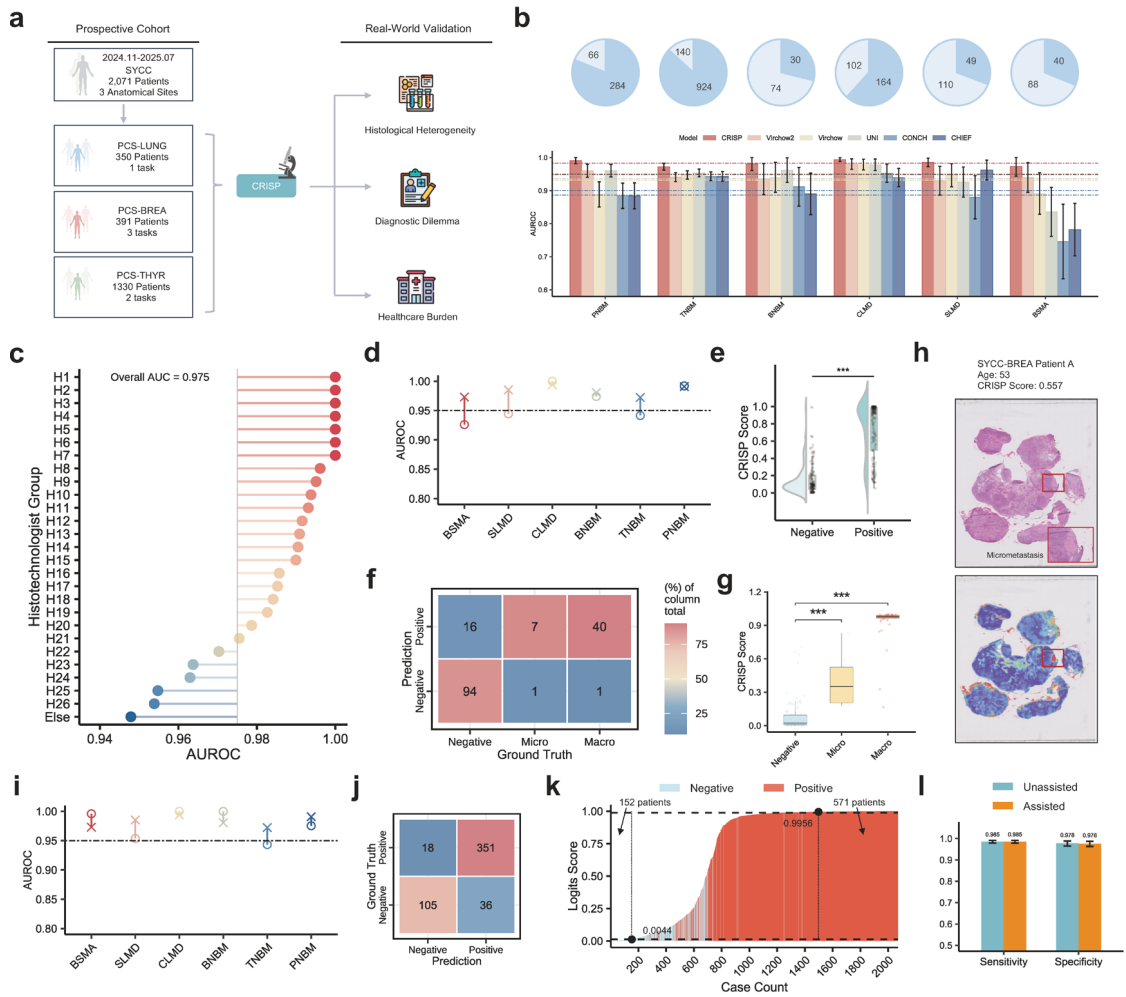

**Fig. 6. Prospective validation of the clinical utility of CRISP.**
(A) Data composition and experimental design of the real-world deployment study. (B) Data composition of each downstream task and AUROC comparison across different models. (C) AUROC comparison of CRISP predictions across histotechnologist subgroups. (D) AUROC of CRISP across all cases versus difficult cases in each downstream task; crosses denote all cases, circles denote the difficult-case subgroup. (E) Distribution of CRISP prediction scores for all difficult malignant and benign cases. (F) Confusion matrix of CRISP predictions for negative lymph nodes, micrometastases, and macrometastases in the SLMD task. (G) Distribution of CRISP prediction scores across negative, micrometastatic, and macrometastatic cases. (H) Visualization heatmap of CRISP predictions in a representative micrometastatic case. (I) AUROC of CRISP across all cases versus cases requiring immunohistochemistry (IHC) in each downstream task; crosses denote all cases, circles denote the IHC subgroup. (J) Confusion matrix of CRISP predictions in the IHC-required subgroup. (K) Waterfall plot of CRISP predictions across all 2,071 patients. Using a predefined threshold, 152 typical negative and 571 typical positive patients were identified. (L) Comparison of pathologists' diagnostic sensitivity and specificity before and after applying the CRISP-assisted prescreening strategy. Micro, Micrometastases; Macro, Macrometastases.

**Supplementary Figures**

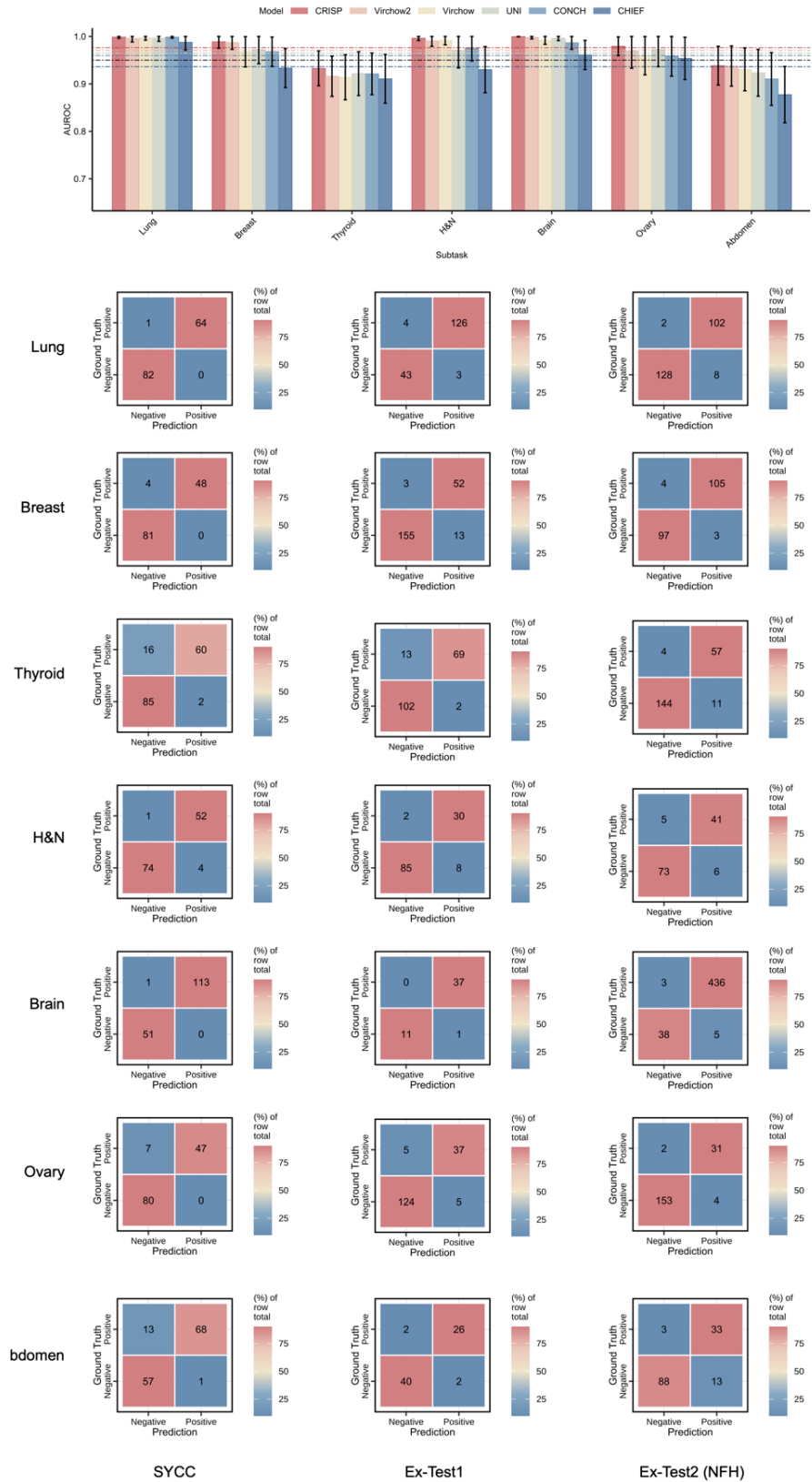

Figure S1. Extended Performance of CRISP in common benign–malignant differentiation tasks. (A) Predictive performance in SYCC cohort. (B) Confusion matrices showing CRISP's predictions for common benign–malignant differentiation tasks.

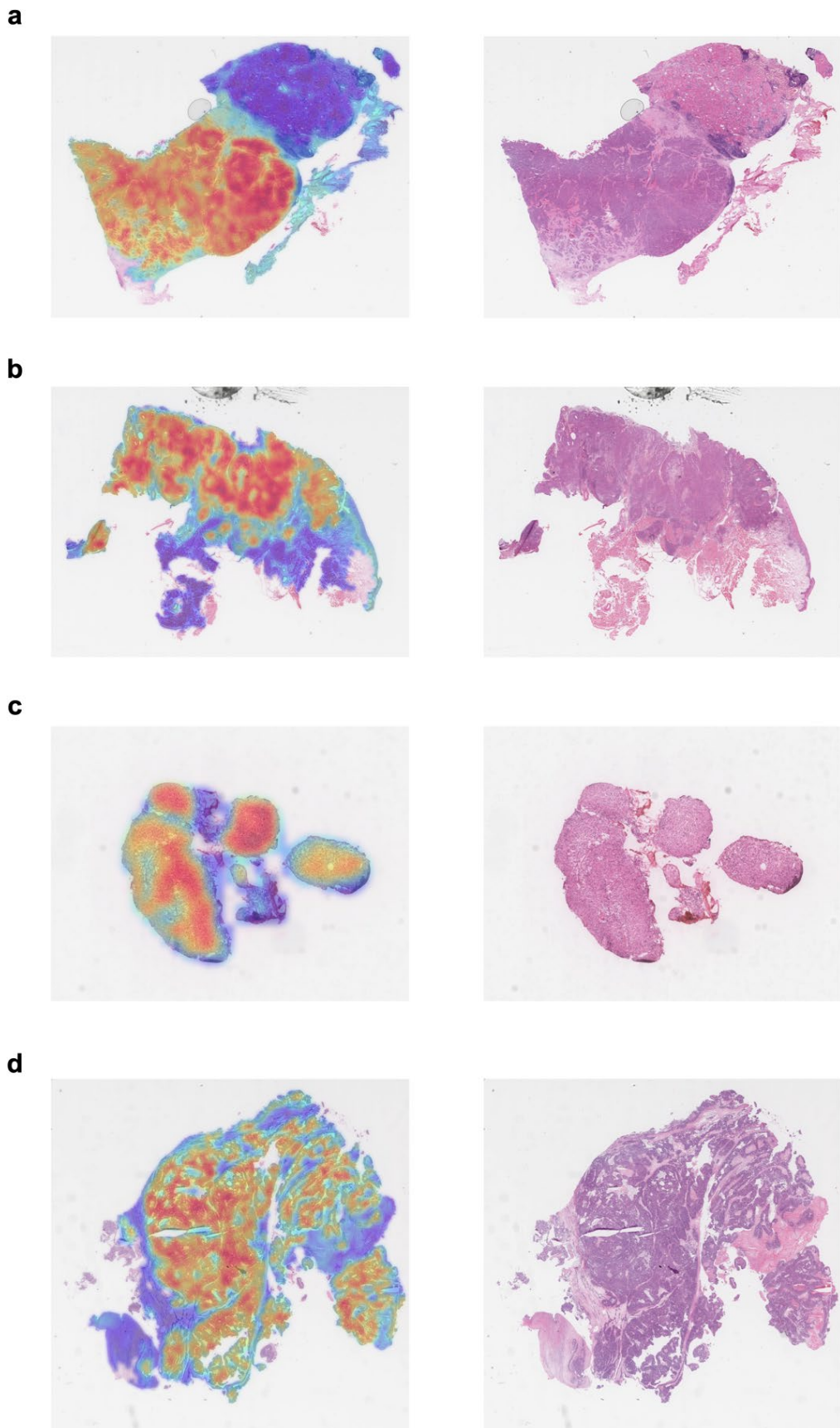

Figure S2. Extended attention visualization of CRISP predictions on malignant thyroid (A), head and neck (B), brain (C) and ovary (D) WSIs, showing the original WSIs and CRISP heatmaps.

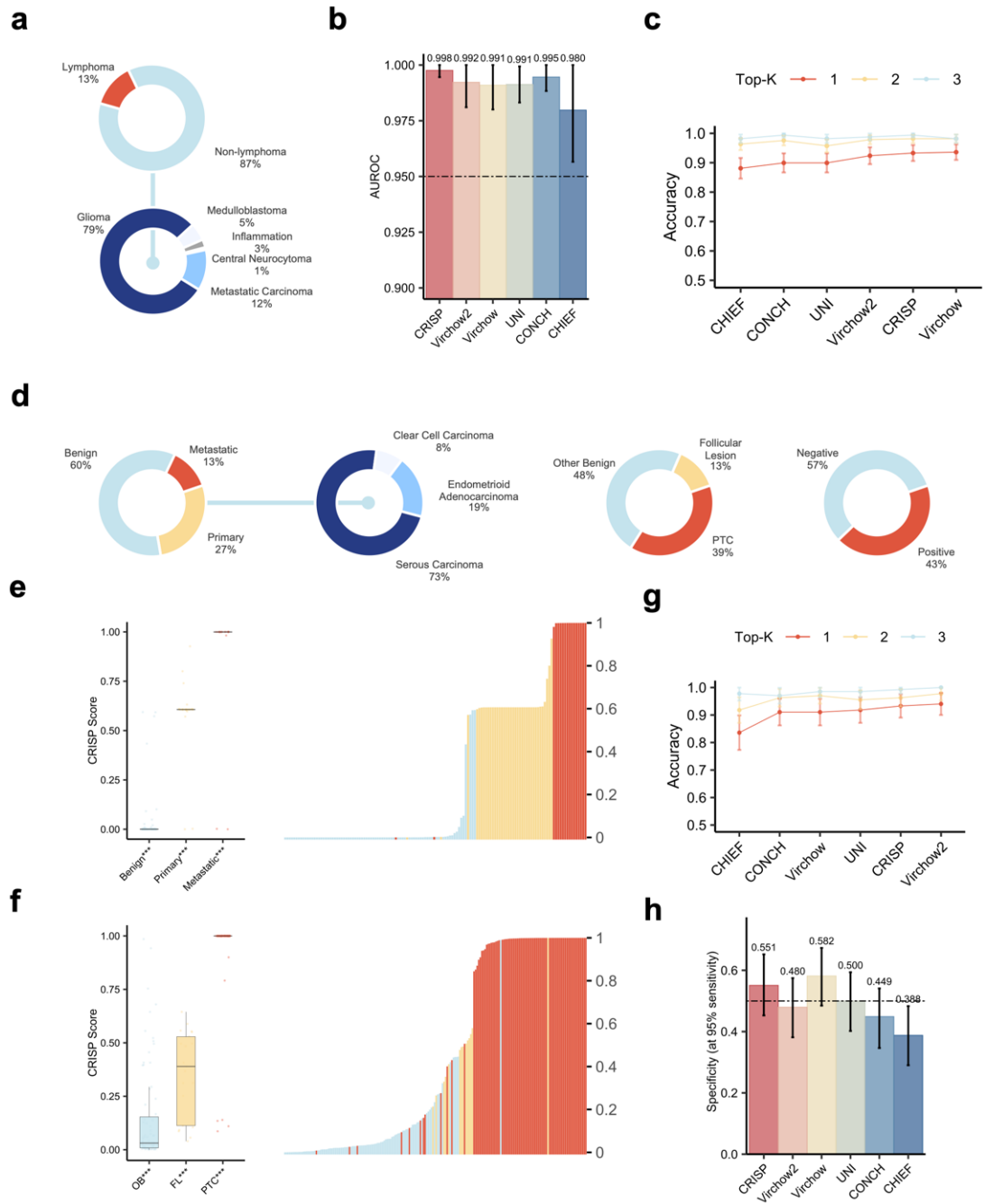

Figure S3. Extended Evaluation of CRISP for intraoperative decision-making.
(A) Data composition of the SYCC validation cohort for the PCNSLD task (top) and CNSLC task (bottom). (B) Comparison of AUROC between CRISP and other models on SYCC validation cohorts for the PCNSLD task. (C) Top-k accuracy comparison of CRISP and other models in SYCC validation cohorts for the CNSLC task. (D) Data composition of SYCC validation cohorts for the OCCGS, OCFGS, TNDD, and PGD tasks. (E) Predicted probability distributions of CRISP across classes in the OCCGS task in the SYCC validation cohort. (F) Predicted probability distributions of CRISP across classes in the TNDD task in the SYCC validation cohort. (G) Top-k accuracy comparison of CRISP and other models in the SYCC validation cohort for the OCFGS task. (H) Specificity comparison of CRISP and other models at 95% sensitivity for the PGD task in the SYCC validation cohort.

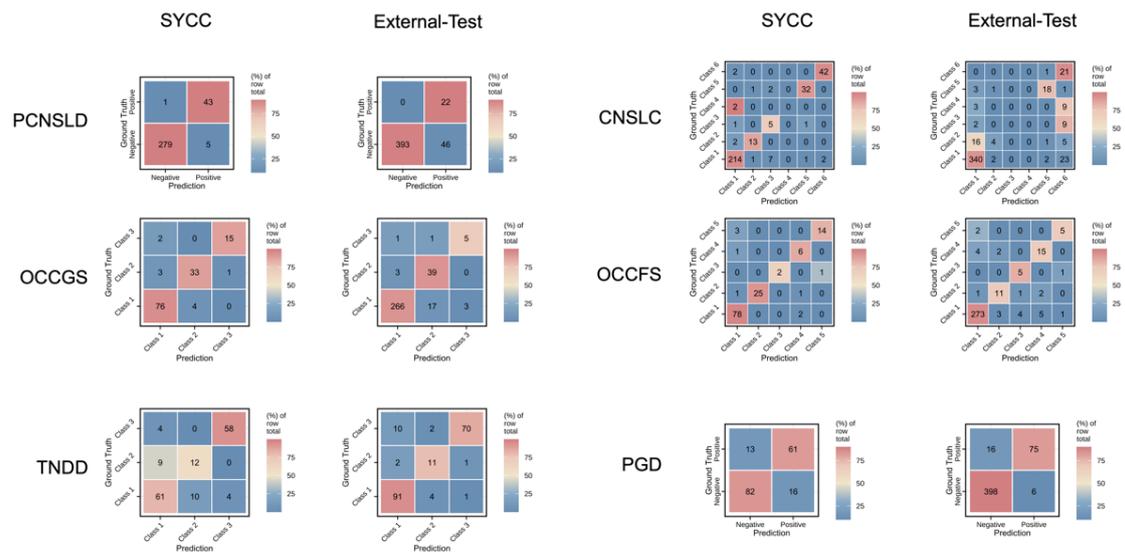

Figure S4. Confusion matrices showing CRISP's predictions for preoperative and intraoperative decision-making tasks.

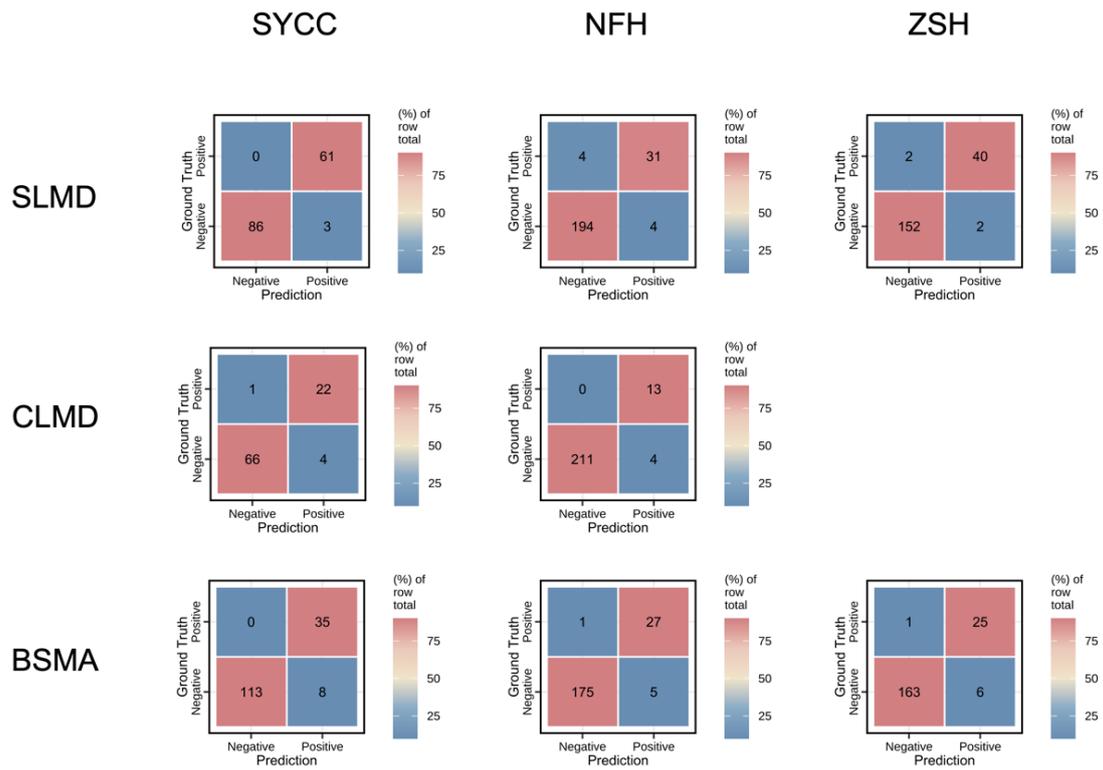

Figure S5. Confusion matrices showing CRISP's predictions in late intraoperative assessment.

| Characteristics | Number of Patients |
|---|---:|
| **Patient Counts** | 20457 |
| **Slide Counts** | 48191 |
| **Data Collection Period** | 2017.5-2023.2 |
| **Age, median (range)** | 49.39 (1, 94) |
| **Sex** | |
| Male | 6856 |
| Female | 13601 |
| **Specimen Site** | |
| Lung | 5158 |
| Breast | 4014 |
| Thyroid | 4135 |
| Head & Neck | 1917 |
| Brain | 1564 |
| Ovary | 1553 |
| Lymph Node | 414 |
| Uterus | 286 |
| Cervix | 37 |
| Skin | 46 |
| Bile Duct | 47 |
| Soft Tissue | 89 |
| Liver | 292 |
| Pancreas | 21 |
| Colorectum | 107 |
| Pleura | 25 |
| Stomach | 204 |
| Bone | 20 |
| Esophagus | 24 |
| Kidney | 142 |
| Bladder | 306 |
| Adrenal Gland | 2 |
| Testis | 32 |
| Thymus | 22 |
| **Diagnosis** | |
| Malignant | 9014 |
| Benign | 11443 |

Table S1.1. Baseline characteristics of patients in the retrospective SYCC cohort.

| Characteristics | Number of Patients |
|---|---|
| **Patient Counts** | 4719 |
| **Slide Counts** | 11879 |
| **Data Collection Period** | 2020.7-2024.7 |
| **Age, median (range)** | 49.16 (3, 87) |
| **Sex** | |
| Male | 1145 |
| Female | 3574 |
| **Specimen Site** | |
| Lung | 1224 |
| Breast | 621 |
| Thyroid | 1798 |
| Ovary | 756 |
| Lymph Node | 320 |
| **Diagnosis** | |
| Malignant | 2737 |
| Benign | 1982 |

Table S1.2. Baseline characteristics of patients in the retrospective ZJH cohort.

| Characteristics | Number of Patients |
|---|---:|
| **Patient Counts** | 7210 |
| **Slide Counts** | 10800 |
| **Data Collection Period** | 2023.3-2024.7 |
| **Age, median (range)** | 47.70 (5, 94) |
| **Sex** | |
| Male | 1345 |
| Female | 5865 |
| **Specimen Site** | |
| Lung | 1558 |
| Breast | 2822 |
| Thyroid | 713 |
| Head & Neck | 647 |
| Brain | 153 |
| Ovary | 371 |
| Lymph Node | 671 |
| Uterus | 99 |
| Cervix | 5 |
| Skin | 41 |
| Bile Duct | 3 |
| Soft Tissue | 10 |
| Liver | 29 |
| Pancreas | 9 |
| Colorectum | 20 |
| Pleura | 13 |
| Stomach | 6 |
| Bone | 7 |
| Esophagus | 6 |
| Kidney | 7 |
| Bladder | 10 |
| Testis | 1 |
| Thymus | 9 |
| **Diagnosis** | |
| Malignant | 3297 |
| Benign | 3913 |

Table S1.3. Baseline characteristics of patients in the retrospective JMH cohort.

| Characteristics | Number of Patients |
| --- | --- |
| **Patient Counts** | 1676 |
| **Slide Counts** | 5000 |
| **Data Collection Period** | 2022.1-2023.11 |
| **Age, median (range)** | 49.45 (8, 90) |
| **Sex** | |
| Male | 604 |
| Female | 1072 |
| **Specimen Site** | |
| Lung | 201 |
| Breast | 186 |
| Thyroid | 182 |
| Head & Neck | 207 |
| Brain | 30 |
| Ovary | 38 |
| Lymph Node | 486 |
| Uterus | 145 |
| Cervix | 43 |
| Skin | 42 |
| Bile Duct | 35 |
| Soft Tissue | 29 |
| Liver | 13 |
| Pancreas | 13 |
| Colorectum | 7 |
| Pleura | 7 |
| Stomach | 3 |
| Esophagus | 2 |
| Kidney | 3 |
| Bladder | 1 |
| Adrenal Gland | 1 |
| Testis | 1 |
| Thymus | 1 |
| **Diagnosis** | |
| Malignant | 614 |
| Benign | 1062 |

Table S1.4. Baseline characteristics of patients in the retrospective ZSH cohort.

| Characteristics | Number of Patients |
| --- | ---: |
| **Patient Counts** | 3344 |
| **Slide Counts** | 4714 |
| **Data Collection Period** | 2023.3-2024.1 |
| **Age, median (range)** | 50.85 (2, 90) |
| **Sex** | |
| Male | 1121 |
| Female | 2223 |
| **Specimen Site** | |
| Lung | 561 |
| Breast | 682 |
| Thyroid | 206 |
| Head & Neck | 293 |
| Brain | 669 |
| Ovary | 119 |
| Lymph Node | 241 |
| Uterus | 35 |
| Cervix | 77 |
| Skin | 56 |
| Bile Duct | 35 |
| Soft Tissue | 50 |
| Liver | 25 |
| Pancreas | 8 |
| Colorectum | 146 |
| Pleura | 11 |
| Stomach | 37 |
| Bone | 44 |
| Esophagus | 31 |
| Kidney | 3 |
| Bladder | 1 |
| Testis | 2 |
| Thymus | 12 |
| **Diagnosis** | |
| Malignant | 1779 |
| Benign | 1565 |

Table S1.5. Baseline characteristics of patients in the retrospective NFH cohort.

| Characteristics | Number of Patients |
|---|---:|
| **Patient Counts** | 508 |
| **Slide Counts** | 754 |
| **Data Collection Period** | 2023.7-2024.4 |
| **Age, median (range)** | 51.96 (15, 83) |
| **Sex** | |
| Male | 129 |
| Female | 379 |
| **Specimen Site** | |
| Lung | 142 |
| Breast | 220 |
| Thyroid | 59 |
| Head & Neck | 19 |
| Brain | 16 |
| Ovary | 16 |
| Lymph Node | 15 |
| Uterus | 4 |
| Cervix | 1 |
| Skin | 2 |
| Bile Duct | 3 |
| Soft Tissue | 2 |
| Pleura | 5 |
| Bladder | 2 |
| Testis | 1 |
| Thymus | 1 |
| **Diagnosis** | |
| Malignant | 273 |
| Benign | 235 |

Table S1.6. Baseline characteristics of patients in the retrospective GZH cohort.

| Characteristics | Number of Patients |
| --- | --- |
| **Patient Counts** | 225 |
| **Slide Counts** | 226 |
| **Data Collection Period** | 2023.11-2024.4 |
| **Age, median (range)** | 44.75 (14, 81) |
| **Sex** | |
| Male | 28 |
| Female | 197 |
| **Specimen Site** | |
| Lung | 9 |
| Breast | 68 |
| Thyroid | 45 |
| Head & Neck | 6 |
| Brain | 4 |
| Ovary | 26 |
| Lymph Node | 9 |
| Uterus | 35 |
| Cervix | 9 |
| Skin | 4 |
| Soft Tissue | 3 |
| Liver | 1 |
| Pancreas | 1 |
| Colorectum | 2 |
| Pleura | 1 |
| Thymus | 2 |
| **Diagnosis** | |
| Malignant | 47 |
| Benign | 178 |

Table S1.7. Baseline characteristics of patients in the retrospective HZH cohort.

| Characteristics | Number of Patients |
|---|---|
| **Patient Counts** | 114 |
| **Slide Counts** | 114 |
| **Data Collection Period** | 2023.10-2024.2 |
| **Age, median (range)** | 46.54 (26, 81) |
| **Sex** | |
| Male | 27 |
| Female | 87 |
| **Specimen Site** | |
| Breast | 26 |
| Thyroid | 78 |
| Ovary | 3 |
| Skin | 2 |
| Liver | 1 |
| Pancreas | 1 |
| Colorectum | 2 |
| Bone | 1 |
| **Diagnosis** | |
| Malignant | 114 |
| Benign | 0 |

Table S1.8. Baseline characteristics of patients in the retrospective DGH cohort.

| Characteristics | Number of Patients |
|---|---:|
| **Patient Counts** | 1330 |
| **Slide Counts** | 1570 |
| **Received Date** | 2024.11.11-2025.7.1 |
| **Age, median (range)** | 41.69 (18, 80) |
| **Sex** | |
| Male | 374 |
| Female | 956 |
| **Diagnosis** | |
| Malignant | 1088 |
| Benign | 242 |

Table S1.9. Baseline characteristics of patients in the prospective PCS-THYR cohort.

| Characteristics | Number of Patients |
|---|---|
| **Patient Counts** | 391 |
| **Slide Counts** | 1770 |
| **Received Date** | 2024.11.12-2025.5.1 |
| **Age, median (range)** | 48.85 (18, 87) |
| **Sex** | |
| Male | 1 |
| Female | 390 |
| **Diagnosis** | |
| Malignant | 119 |
| Benign | 272 |

Table S1.10. Baseline characteristics of patients in the prospective PCS-BREA cohort.

| Characteristics | Number of Patients |
| --- | --- |
| **Patient Counts** | 350 |
| **Slide Counts** | 803 |
| **Received Date** | 2024.12.4-2025.5.1 |
| **Age, median (range)** | 56.94 (24, 82) |
| **Sex** | |
| Male | 136 |
| Female | 214 |
| **Diagnosis** | |
| Malignant | 284 |
| Benign | 66 |

Table S1.11. Baseline characteristics of patients in the prospective PCS-LUNG cohort.

| Cohort | Data Split | Sites | | | | | | |
| --- | --- | --- | --- | --- | --- | --- | --- | --- |
| | | Lung | Breast | Thyroid | H&N | Brain | Ovary | Abdomen |
| SYCC | Internal-Train | 257 | 272 | 296 | 261 | 234 | 253 | 208 |
| | Internal-Validation | 110 | 97 | 89 | 87 | 108 | 112 | 95 |
| | Internal-Test | 147 | 133 | 163 | 131 | 165 | 134 | 139 |
| ZSH | External-Test | 176 | 119 | 186 | 110 | 37 | 171 | 70 |
| NFH | External-Test | 240 | 209 | 216 | 125 | 482 | 190 | 137 |
| GZH | External-Test | 0 | 187 | 0 | 13 | 12 | 0 | 0 |
| DGH | External-Test | 0 | 26 | 0 | 0 | 0 | 0 | 0 |
| HZH | External-Test | 0 | 63 | 0 | 2 | 0 | 0 | 0 |
| | Sum | 930 | 1106 | 950 | 729 | 1038 | 860 | 649 |

Table S2.1. Data composition across 21 benign-malignant classification tasks

| Model | AUROC | Spe (at 90% Sen) | Spe (at 95% Sen) | Accuracy | F1 |
| --- | --- | --- | --- | --- | --- |
| CHIEF | 0.897±0.091 | 0.612±0.285 | 0.431±0.287 | 0.863±0.088 | 0.822±0.125 |
| CONCH | 0.942±0.042 | 0.802±0.188 | 0.607±0.260 | 0.914±0.047 | 0.885±0.085 |
| UNI | 0.947±0.032 | 0.877±0.195 | 0.591±0.277 | 0.924±0.033 | 0.898±0.066 |
| Virchow | 0.944±0.032 | 0.777±0.205 | 0.565±0.274 | 0.892±0.055 | 0.866±0.077 |
| Virchow2 | 0.953±0.030 | 0.840±0.142 | 0.644±0.274 | 0.927±0.041 | 0.901±0.065 |
| CRISP | **0.970±0.022** | **0.893±0.100** | **0.718±0.264** | **0.945±0.033** | **0.924±0.053** |

Table S2.2. Average performance of various models across 21 benign-malignant differentiation tasks.

| Cohort | Model | AUROC | Spe (at 90% Sen) | Spe (at 95% Sen) | Accuracy | F1 |
|---|---|---|---|---|---|---|
| SYCC | CHIEF | 0.988 (0.971, 1.000) | **1.000 (1.000, 1.000)** | 0.902 (0.836, 0.963) | 0.980 (0.952, 1.000) | 0.976 (0.941, 1.000) |
| | CONCH | **0.999 (0.996, 1.000)** | **1.000 (1.000, 1.000)** | **1.000 (1.000, 1.000)** | 0.986 (0.966, 1.000) | 0.984 (0.961, 1.000) |
| | UNI | 0.996 (0.990, 1.000) | **1.000 (1.000, 1.000)** | **1.000 (1.000, 1.000)** | 0.986 (0.966, 1.000) | 0.984 (0.960, 1.000) |
| | Virchow | 0.997 (0.992, 1.000) | **1.000 (1.000, 1.000)** | 0.976 (0.938, 1.000) | 0.973 (0.946, 0.993) | 0.970 (0.936, 0.993) |
| | Virchow2 | 0.996 (0.988, 1.000) | **1.000 (1.000, 1.000)** | 1.000 (1.000, 1.000) | 0.986 (0.966, 1.000) | 0.984 (0.959, 1.000) |
| | CRISP | **0.999 (0.996, 1.000)** | **1.000 (1.000, 1.000)** | **1.000 (1.000, 1.000)** | **0.993 (0.980, 1.000)** | **0.992 (0.975, 1.000)** |
| ZSH | CHIEF | 0.903 (0.854, 0.953) | 0.630 (0.473, 0.765) | 0.457 (0.311, 0.605) | 0.824 (0.767, 0.875) | 0.869 (0.823, 0.912) |
| | CONCH | 0.935 (0.891, 0.978) | 0.870 (0.767, 0.960) | 0.761 (0.635, 0.882) | 0.920 (0.875, 0.960) | 0.946 (0.913, 0.973) |
| | UNI | 0.950 (0.907, 0.993) | **0.935 (0.857, 1.000)** | 0.826 (0.712, 0.927) | 0.915 (0.875, 0.955) | 0.940 (0.908, 0.967) |
| | Virchow | 0.893 (0.846, 0.941) | 0.543 (0.394, 0.696) | 0.370 (0.231, 0.508) | 0.847 (0.790, 0.898) | 0.890 (0.849, 0.927) |
| | Virchow2 | 0.962 (0.925, 0.998) | **0.935 (0.860, 1.000)** | 0.913 (0.826, 0.981) | 0.949 (0.915, 0.977) | 0.965 (0.941, 0.985) |
| | CRISP | **0.967 (0.932, 1.000)** | **0.935 (0.848, 1.000)** | **0.935 (0.863, 1.000)** | **0.960 (0.926, 0.983)** | **0.973 (0.949, 0.992)** |
| NFH | CHIEF | 0.946 (0.916, 0.975) | 0.824 (0.759, 0.886) | 0.588 (0.504, 0.669) | 0.896 (0.858, 0.933) | 0.876 (0.829, 0.922) |
| | CONCH | 0.967 (0.941, 0.993) | 0.934 (0.887, 0.971) | 0.853 (0.787, 0.906) | 0.946 (0.917, 0.971) | 0.935 (0.898, 0.967) |
| | UNI | 0.961 (0.932, 0.989) | 0.897 (0.844, 0.944) | 0.772 (0.701, 0.839) | 0.938 (0.904, 0.967) | 0.925 (0.882, 0.958) |
| | Virchow | 0.960 (0.933, 0.987) | 0.912 (0.863, 0.959) | 0.809 (0.743, 0.875) | 0.912 (0.875, 0.946) | 0.900 (0.851, 0.940) |
| | Virchow2 | 0.975 (0.954, 0.995) | 0.934 (0.889, 0.972) | 0.919 (0.871, 0.961) | 0.942 (0.908, 0.967) | 0.935 (0.899, 0.966) |
| | CRISP | **0.979 (0.958, 1.000)** | **0.963 (0.930, 0.993)** | **0.941 (0.897, 0.977)** | **0.958 (0.933, 0.983)** | **0.953 (0.922, 0.980)** |

Table S2.3. Performance of various models for the discrimination of benign and malignant lung lesions across different cohorts.

| Cohort | Model | AUROC | Spe (at 90% Sen) | Spe (at 95% Sen) | Accuracy | F1 |
|---|---|---|---|---|---|---|
| SYCC | CHIEF | 0.933 (0.892, 0.974) | 0.716 (0.611, 0.806) | 0.543 (0.442, 0.653) | 0.887 (0.835, 0.940) | 0.851 (0.769, 0.917) |
| | CONCH | 0.968 (0.937, 0.999) | 0.889 (0.820, 0.951) | 0.605 (0.494, 0.711) | 0.940 (0.895, 0.977) | 0.922 (0.860, 0.970) |
| | UNI | 0.972 (0.942, 1.000) | 0.877 (0.805, 0.948) | 0.642 (0.537, 0.747) | 0.962 (0.932, 0.992) | 0.949 (0.897, 0.989) |
| | Virchow | 0.968 (0.936, 1.000) | 0.963 (0.917, 1.000) | 0.642 (0.545, 0.747) | 0.947 (0.902, 0.978) | 0.932 (0.878, 0.979) |
| | Virchow2 | 0.987 (0.973, 1.000) | **1.000 (1.000, 1.000)** | 0.802 (0.718, 0.885) | **0.970 (0.940, 0.992)** | **0.960 (0.920, 0.991)** |
| | CRISP | **0.989 (0.975, 1.000)** | **1.000 (1.000, 1.000)** | 0.790 (0.699, 0.880) | **0.970 (0.940, 0.992)** | **0.960 (0.913, 0.991)** |
| ZHDG | CHIEF | 0.852 (0.792, 0.912) | 0.506 (0.426, 0.581) | 0.256 (0.195, 0.324) | 0.861 (0.816, 0.906) | 0.735 (0.635, 0.815) |
| | CONCH | 0.947 (0.912, 0.982) | 0.810 (0.747, 0.863) | 0.464 (0.389, 0.535) | 0.924 (0.888, 0.955) | 0.847 (0.768, 0.911) |
| | UNI | 0.918 (0.868, 0.968) | 0.762 (0.696, 0.829) | 0.280 (0.216, 0.349) | 0.901 (0.861, 0.937) | 0.804 (0.720, 0.874) |
| | Virchow | 0.884 (0.828, 0.939) | 0.804 (0.742, 0.861) | 0.143 (0.091, 0.200) | 0.857 (0.812, 0.901) | 0.758 (0.677, 0.837) |
| | Virchow2 | 0.928 (0.885, 0.971) | 0.833 (0.775, 0.888) | **0.744 (0.673, 0.810)** | 0.901 (0.861, 0.937) | 0.817 (0.738, 0.889) |
| | CRISP | **0.950 (0.910, 0.990)** | **0.929 (0.885, 0.966)** | 0.298 (0.233, 0.366) | **0.928 (0.892, 0.960)** | **0.867 (0.796, 0.923)** |
| NFH | CHIEF | 0.906 (0.866, 0.947) | 0.590 (0.495, 0.687) | 0.410 (0.321, 0.500) | 0.847 (0.794, 0.895) | 0.840 (0.784, 0.892) |
| | CONCH | 0.975 (0.954, 0.995) | 0.950 (0.902, 0.989) | 0.950 (0.901, 0.989) | 0.957 (0.928, 0.981) | 0.959 (0.931, 0.982) |
| | UNI | 0.979 (0.962, 0.997) | 0.980 (0.949, 1.000) | 0.900 (0.837, 0.954) | 0.952 (0.923, 0.981) | 0.953 (0.923, 0.978) |
| | Virchow | 0.942 (0.910, 0.974) | 0.860 (0.792, 0.923) | 0.700 (0.610, 0.788) | 0.890 (0.842, 0.928) | 0.893 (0.845, 0.933) |
| | Virchow2 | 0.987 (0.977, 0.998) | **0.980 (0.949, 1.000)** | 0.910 (0.851, 0.962) | 0.957 (0.928, 0.981) | 0.958 (0.928, 0.985) |
| | CRISP | **0.989 (0.978, 1.000)** | 0.970 (0.932, 1.000) | **0.970 (0.929, 1.000)** | **0.967 (0.943, 0.990)** | **0.968 (0.942, 0.991)** |

Table S2.4. Performance of various models for the discrimination of benign and malignant breast lesions across different cohorts.

| Cohort | Model | AUROC | Spe (at 90% Sen) | Spe (at 95% Sen) | Accuracy | F1 |
|---|---|---|---|---|---|---|
| SYCC | CHIEF | 0.911 (0.859, 0.962) | 0.609 (0.505, 0.709) | 0.126 (0.064, 0.202) | 0.877 (0.822, 0.926) | 0.853 (0.785, 0.912) |
| | CONCH | 0.921 (0.877, 0.965) | 0.552 (0.448, 0.652) | 0.356 (0.256, 0.467) | 0.896 (0.847, 0.939) | 0.878 (0.817, 0.932) |
| | UNI | 0.922 (0.875, 0.968) | 0.644 (0.541, 0.744) | 0.241 (0.154, 0.333) | **0.902 (0.853, 0.945)** | **0.884 (0.821, 0.936)** |
| | Virchow | 0.914 (0.866, 0.962) | 0.632 (0.527, 0.735) | 0.310 (0.215, 0.405) | 0.877 (0.822, 0.926) | 0.861 (0.795, 0.915) |
| | Virchow2 | 0.916 (0.874, 0.958) | 0.621 (0.520, 0.719) | 0.448 (0.344, 0.557) | 0.871 (0.822, 0.920) | 0.844 (0.773, 0.905) |
| | CRISP | **0.933 (0.896, 0.969)** | **0.690 (0.593, 0.775)** | **0.494 (0.391, 0.600)** | 0.890 (0.840, 0.933) | 0.870 (0.806, 0.924) |
| ZSH | CHIEF | 0.907 (0.862, 0.952) | 0.558 (0.465, 0.661) | 0.308 (0.223, 0.396) | 0.871 (0.822, 0.919) | 0.836 (0.766, 0.895) |
| | CONCH | 0.936 (0.897, 0.975) | 0.769 (0.680, 0.847) | 0.404 (0.312, 0.495) | 0.892 (0.849, 0.935) | 0.873 (0.820, 0.925) |
| | UNI | 0.930 (0.888, 0.971) | 0.702 (0.617, 0.791) | 0.413 (0.325, 0.509) | 0.909 (0.866, 0.946) | 0.889 (0.835, 0.942) |
| | Virchow | 0.917 (0.874, 0.959) | 0.702 (0.618, 0.786) | 0.375 (0.287, 0.465) | 0.860 (0.806, 0.909) | 0.845 (0.778, 0.900) |
| | Virchow2 | 0.925 (0.878, 0.971) | 0.740 (0.657, 0.815) | 0.183 (0.115, 0.266) | 0.914 (0.871, 0.952) | 0.892 (0.832, 0.941) |
| | CRISP | **0.952 (0.919, 0.985)** | **0.837 (0.755, 0.909)** | **0.519 (0.419, 0.611)** | **0.919 (0.882, 0.952)** | **0.902 (0.853, 0.945)** |
| NFH | CHIEF | 0.955 (0.917, 0.993) | 0.845 (0.786, 0.898) | 0.716 (0.646, 0.785) | 0.935 (0.903, 0.963) | 0.883 (0.819, 0.940) |
| | CONCH | 0.962 (0.922, 1.000) | **0.955 (0.920, 0.986)** | 0.477 (0.399, 0.561) | **0.949 (0.917, 0.977)** | **0.913 (0.857, 0.961)** |
| | UNI | 0.960 (0.924, 0.996) | 0.923 (0.878, 0.961) | 0.574 (0.500, 0.648) | 0.944 (0.912, 0.972) | 0.902 (0.842, 0.950) |
| | Virchow | 0.957 (0.924, 0.990) | 0.858 (0.800, 0.912) | 0.535 (0.459, 0.615) | 0.935 (0.903, 0.968) | 0.879 (0.807, 0.937) |
| | Virchow2 | 0.949 (0.911, 0.988) | 0.806 (0.748, 0.864) | 0.665 (0.589, 0.739) | 0.903 (0.866, 0.940) | 0.837 (0.760, 0.901) |
| | CRISP | **0.970 (0.943, 0.997)** | 0.929 (0.887, 0.966) | **0.716 (0.642, 0.785)** | 0.931 (0.898, 0.958) | 0.884 (0.817, 0.937) |

Table S2.5. Performance of various models for the discrimination of benign and malignant thyroid lesions across different cohorts.

| Cohort | Model | AUROC | Spe (at 90% Sen) | Spe (at 95% Sen) | Accuracy | F1 |
| --- | --- | --- | --- | --- | --- | --- |
| SYCC | CHIEF | 0.930 (0.882, 0.979) | 0.628 (0.524, 0.732) | 0.410 (0.291, 0.521) | 0.893 (0.840, 0.947) | 0.868 (0.791, 0.933) |
| | CONCH | 0.975 (0.948, 1.000) | 0.910 (0.846, 0.971) | 0.744 (0.654, 0.840) | 0.947 (0.908, 0.977) | 0.932 (0.876, 0.976) |
| | UNI | 0.971 (0.934, 1.000) | 0.949 (0.900, 0.988) | 0.795 (0.695, 0.875) | 0.947 (0.901, 0.977) | 0.932 (0.875, 0.975) |
| | Virchow | 0.992 (0.982, 1.000) | 0.962 (0.910, 1.000) | 0.885 (0.811, 0.949) | 0.954 (0.916, 0.985) | 0.941 (0.889, 0.982) |
| | Virchow2 | 0.991 (0.979, 1.000) | 0.987 (0.959, 1.000) | 0.885 (0.813, 0.950) | **0.962 (0.924, 0.992)** | 0.952 (0.911, 0.990) |
| | CRISP | **0.996 (0.991, 1.000)** | **1.000 (1.000, 1.000)** | **0.949 (0.899, 0.988)** | **0.962 (0.924, 0.992)** | **0.954 (0.906, 0.990)** |
| ZHG | CHIEF | 0.963 (0.932, 0.993) | 0.871 (0.802, 0.938) | 0.634 (0.531, 0.733) | 0.888 (0.832, 0.936) | 0.816 (0.716, 0.896) |
| | CONCH | 0.920 (0.872, 0.968) | 0.731 (0.634, 0.818) | 0.656 (0.558, 0.744) | 0.888 (0.832, 0.944) | 0.781 (0.650, 0.886) |
| | UNI | 0.914 (0.848, 0.979) | 0.559 (0.456, 0.659) | 0.312 (0.219, 0.409) | 0.888 (0.832, 0.936) | 0.800 (0.690, 0.896) |
| | Virchow | 0.960 (0.931, 0.990) | 0.903 (0.837, 0.957) | **0.882 (0.816, 0.943)** | 0.912 (0.856, 0.960) | 0.853 (0.756, 0.928) |
| | Virchow2 | 0.950 (0.905, 0.994) | 0.806 (0.724, 0.885) | 0.473 (0.376, 0.571) | **0.944 (0.904, 0.984)** | **0.881 (0.780, 0.958)** |
| | CRISP | **0.977 (0.955, 0.999)** | **0.914 (0.851, 0.967)** | 0.753 (0.660, 0.835) | 0.920 (0.872, 0.960) | 0.857 (0.758, 0.938) |
| NFH | CHIEF | 0.869 (0.784, 0.954) | 0.241 (0.146, 0.346) | 0.013 (0.000, 0.041) | 0.880 (0.816, 0.936) | 0.835 (0.744, 0.911) |
| | CONCH | 0.928 (0.875, 0.981) | 0.823 (0.735, 0.903) | 0.684 (0.575, 0.781) | 0.872 (0.808, 0.928) | 0.840 (0.755, 0.906) |
| | UNI | 0.954 (0.919, 0.990) | 0.785 (0.694, 0.868) | 0.646 (0.540, 0.743) | 0.912 (0.856, 0.952) | 0.876 (0.795, 0.941) |
| | Virchow | 0.942 (0.895, 0.989) | **0.911 (0.843, 0.965)** | **0.835 (0.756, 0.913)** | **0.920 (0.872, 0.960)** | **0.896 (0.823, 0.953)** |
| | Virchow2 | 0.936 (0.890, 0.982) | 0.785 (0.688, 0.870) | 0.481 (0.373, 0.586) | 0.888 (0.832, 0.944) | 0.848 (0.761, 0.914) |
| | CRISP | **0.965 (0.938, 0.992)** | 0.772 (0.677, 0.863) | 0.722 (0.613, 0.813) | 0.912 (0.856, 0.960) | 0.882 (0.810, 0.943) |

Table S2.6. Performance of various models for the discrimination of benign and malignant head and neck lesions across different cohorts.

| Cohort | Model | AUROC | Spe (at 90% Sen) | Spe (at 95% Sen) | Accuracy | F1 |
| --- | --- | --- | --- | --- | --- | --- |
| SYCC | CHIEF | 0.961 (0.930, 0.992) | 0.902 (0.809, 0.979) | 0.902 (0.805, 0.979) | 0.945 (0.909, 0.976) | 0.961 (0.933, 0.985) |
| | CONCH | 0.987 (0.973, 1.000) | 0.961 (0.896, 1.000) | 0.941 (0.870, 1.000) | 0.964 (0.933, 0.988) | 0.974 (0.950, 0.991) |
| | UNI | 0.996 (0.991, 1.000) | 0.980 (0.935, 1.000) | 0.961 (0.889, 1.000) | 0.964 (0.933, 0.988) | 0.973 (0.948, 0.991) |
| | Virchow | 0.992 (0.983, 1.000) | **1.000 (1.000, 1.000)** | 0.922 (0.839, 0.982) | 0.952 (0.915, 0.982) | 0.964 (0.936, 0.986) |
| | Virchow2 | 0.998 (0.995, 1.000) | **1.000 (1.000, 1.000)** | **1.000 (1.000, 1.000)** | 0.976 (0.952, 0.994) | 0.982 (0.962, 0.996) |
| | CRISP | **1.000 (0.999, 1.000)** | **1.000 (1.000, 1.000)** | **1.000 (1.000, 1.000)** | **0.994 (0.982, 1.000)** | **0.996 (0.986, 1.000)** |
| ZSGZ | CHIEF | 0.840 (0.722, 0.958) | 0.083 (0.000, 0.286) | 0.083 (0.000, 0.278) | 0.755 (0.633, 0.878) | 0.812 (0.700, 0.909) |
| | CONCH | 0.948 (0.880, 1.000) | 0.750 (0.500, 1.000) | 0.667 (0.400, 0.917) | 0.898 (0.816, 0.980) | 0.930 (0.853, 0.986) |
| | UNI | 0.921 (0.837, 1.000) | 0.750 (0.500, 1.000) | 0.500 (0.222, 0.786) | 0.898 (0.816, 0.980) | 0.933 (0.870, 0.986) |
| | Virchow | 0.872 (0.774, 0.969) | 0.417 (0.133, 0.727) | 0.250 (0.000, 0.501) | 0.755 (0.633, 0.878) | 0.806 (0.679, 0.899) |
| | Virchow2 | 0.908 (0.793, 1.000) | 0.667 (0.400, 0.923) | 0.500 (0.200, 0.800) | 0.857 (0.755, 0.939) | 0.899 (0.808, 0.971) |
| | CRISP | **0.984 (0.952, 1.000)** | **0.917 (0.714, 1.000)** | **0.917 (0.714, 1.000)** | **0.980 (0.939, 1.000)** | **0.987 (0.955, 1.000)** |
| NFH | CHIEF | 0.943 (0.897, 0.989) | 0.860 (0.750, 0.956) | 0.860 (0.738, 0.953) | 0.969 (0.952, 0.983) | 0.983 (0.974, 0.991) |
| | CONCH | 0.917 (0.865, 0.969) | 0.837 (0.714, 0.943) | 0.721 (0.583, 0.854) | 0.929 (0.905, 0.952) | 0.960 (0.946, 0.974) |
| | UNI | 0.945 (0.899, 0.992) | 0.907 (0.810, 0.979) | 0.860 (0.742, 0.955) | 0.932 (0.909, 0.954) | 0.961 (0.948, 0.974) |
| | Virchow | 0.941 (0.916, 0.965) | 0.791 (0.657, 0.912) | 0.605 (0.465, 0.745) | 0.817 (0.784, 0.849) | 0.889 (0.866, 0.910) |
| | Virchow2 | 0.964 (0.932, 0.997) | **0.907 (0.806, 0.982)** | 0.884 (0.780, 0.971) | 0.977 (0.963, 0.990) | 0.987 (0.980, 0.994) |
| | CRISP | **0.967 (0.934, 1.000)** | **0.907 (0.811, 0.977)** | **0.907 (0.814, 0.979)** | **0.983 (0.971, 0.994)** | **0.991 (0.984, 0.997)** |

Table S2.7. Performance of various models for the discrimination of benign and malignant brain lesions across different cohorts.

| Cohort | Model | AUROC | Spe (at 90% Sen) | Spe (at 95% Sen) | Accuracy | F1 |
|---|---|---|---|---|---|---|
| SYCC | CHIEF | 0.954 (0.909, 0.998) | 0.887 (0.815, 0.951) | 0.312 (0.217, 0.420) | 0.933 (0.888, 0.970) | 0.916 (0.855, 0.964) |
| | CONCH | 0.959 (0.916, 1.000) | **0.963 (0.918, 1.000)** | 0.350 (0.247, 0.451) | 0.948 (0.910, 0.985) | 0.936 (0.884, 0.977) |
| | UNI | 0.973 (0.936, 1.000) | **0.963 (0.919, 1.000)** | 0.650 (0.542, 0.750) | **0.963 (0.925, 0.993)** | **0.951 (0.905, 0.990)** |
| | Virchow | 0.959 (0.919, 1.000) | 0.950 (0.899, 0.989) | 0.525 (0.412, 0.634) | 0.940 (0.896, 0.978) | 0.926 (0.870, 0.972) |
| | Virchow2 | 0.969 (0.933, 1.000) | 0.950 (0.897, 0.989) | 0.713 (0.609, 0.800) | 0.940 (0.896, 0.978) | 0.927 (0.871, 0.971) |
| | CRISP | **0.979 (0.960, 0.999)** | 0.912 (0.841, 0.966) | **0.775 (0.679, 0.865)** | 0.948 (0.910, 0.978) | 0.931 (0.874, 0.978) |
| ZSH | CHIEF | 0.892 (0.835, 0.948) | 0.628 (0.548, 0.715) | 0.535 (0.445, 0.620) | 0.795 (0.731, 0.854) | 0.673 (0.555, 0.768) |
| | CONCH | 0.949 (0.915, 0.982) | 0.822 (0.750, 0.884) | 0.744 (0.667, 0.817) | 0.848 (0.789, 0.901) | 0.750 (0.652, 0.835) |
| | UNI | 0.936 (0.901, 0.971) | 0.822 (0.759, 0.883) | 0.798 (0.730, 0.866) | 0.842 (0.789, 0.895) | 0.752 (0.659, 0.831) |
| | Virchow | 0.915 (0.874, 0.957) | 0.760 (0.685, 0.832) | 0.698 (0.620, 0.775) | 0.830 (0.766, 0.883) | 0.724 (0.612, 0.814) |
| | Virchow2 | 0.943 (0.903, 0.984) | 0.837 (0.768, 0.898) | 0.760 (0.686, 0.829) | 0.848 (0.789, 0.901) | 0.755 (0.653, 0.839) |
| | CRISP | **0.974 (0.952, 0.995)** | **0.891 (0.837, 0.940)** | **0.837 (0.769, 0.899)** | **0.942 (0.906, 0.971)** | **0.881 (0.800, 0.944)** |
| NFH | CHIEF | 0.963 (0.933, 0.992) | 0.873 (0.816, 0.923) | 0.675 (0.599, 0.745) | 0.911 (0.868, 0.947) | 0.779 (0.667, 0.876) |
| | CONCH | 0.983 (0.968, 0.999) | 0.943 (0.903, 0.975) | **0.815 (0.755, 0.873)** | 0.942 (0.900, 0.974) | 0.849 (0.750, 0.930) |
| | UNI | 0.981 (0.961, 1.000) | 0.949 (0.911, 0.981) | 0.720 (0.652, 0.789) | 0.947 (0.911, 0.979) | 0.861 (0.767, 0.939) |
| | Virchow | 0.976 (0.955, 0.998) | 0.924 (0.878, 0.963) | 0.732 (0.662, 0.797) | 0.926 (0.889, 0.963) | 0.816 (0.708, 0.901) |
| | Virchow2 | 0.972 (0.940, 1.000) | 0.930 (0.889, 0.968) | 0.497 (0.422, 0.578) | 0.963 (0.937, 0.984) | 0.896 (0.806, 0.964) |
| | CRISP | **0.984 (0.967, 1.000)** | **0.975 (0.948, 0.994)** | 0.790 (0.725, 0.851) | **0.968 (0.942, 0.989)** | **0.912 (0.830, 0.974)** |

Table S2.8. Performance of various models for the discrimination of benign and malignant ovary lesions across different cohorts.

| Cohort | Model | AUROC | Spe (at 90% Sen) | Spe (at 95% Sen) | Accuracy | F1 |
| --- | --- | --- | --- | --- | --- | --- |
| SYCC | CHIEF | 0.877 (0.818, 0.937) | 0.397 (0.273, 0.519) | 0.172 (0.078, 0.271) | 0.835 (0.770, 0.892) | 0.837 (0.765, 0.899) |
|  | CONCH | 0.910 (0.855, 0.966) | 0.466 (0.340, 0.600) | 0.103 (0.032, 0.189) | 0.899 (0.849, 0.942) | 0.905 (0.848, 0.950) |
|  | UNI | 0.923 (0.874, 0.973) | 0.517 (0.392, 0.648) | 0.259 (0.149, 0.380) | 0.906 (0.856, 0.950) | 0.913 (0.865, 0.957) |
|  | Virchow | 0.931 (0.885, 0.976) | 0.690 (0.569, 0.811) | 0.328 (0.208, 0.443) | 0.906 (0.856, 0.957) | 0.913 (0.859, 0.955) |
|  | Virchow2 | **0.938 (0.895, 0.980)** | 0.638 (0.508, 0.763) | **0.414 (0.288, 0.545)** | **0.921 (0.871, 0.964)** | **0.927 (0.881, 0.967)** |
|  | CRISP | **0.938 (0.898, 0.979)** | **0.741 (0.623, 0.850)** | 0.310 (0.194, 0.426) | 0.899 (0.842, 0.942) | 0.907 (0.853, 0.951) |
| ZSH | CHIEF | 0.772 (0.657, 0.887) | 0.214 (0.095, 0.343) | 0.143 (0.045, 0.250) | 0.786 (0.700, 0.886) | 0.651 (0.471, 0.800) |
|  | CONCH | 0.918 (0.853, 0.984) | 0.643 (0.500, 0.780) | **0.357 (0.205, 0.500)** | 0.871 (0.786, 0.943) | 0.836 (0.714, 0.933) |
|  | UNI | 0.923 (0.847, 0.999) | 0.476 (0.333, 0.625) | 0.119 (0.025, 0.225) | 0.914 (0.843, 0.971) | 0.885 (0.789, 0.964) |
|  | Virchow | 0.874 (0.782, 0.966) | 0.381 (0.225, 0.524) | 0.238 (0.106, 0.372) | 0.900 (0.829, 0.957) | 0.863 (0.756, 0.945) |
|  | Virchow2 | 0.918 (0.844, 0.993) | 0.524 (0.375, 0.675) | 0.143 (0.047, 0.243) | 0.900 (0.829, 0.971) | 0.863 (0.739, 0.952) |
|  | CRISP | **0.957 (0.895, 1.000)** | **0.714 (0.564, 0.837)** | 0.167 (0.059, 0.286) | **0.943 (0.886, 0.986)** | **0.929 (0.847, 0.984)** |
| NFH | CHIEF | 0.567 (0.455, 0.680) | 0.000 (0.000, 0.000) | 0.000 (0.000, 0.000) | 0.577 (0.496, 0.657) | 0.420 (0.289, 0.537) |
|  | CONCH | 0.798 (0.709, 0.888) | 0.257 (0.177, 0.343) | 0.099 (0.046, 0.162) | 0.781 (0.715, 0.847) | 0.634 (0.492, 0.756) |
|  | UNI | 0.870 (0.788, 0.951) | 0.307 (0.221, 0.404) | 0.149 (0.082, 0.222) | **0.898 (0.847, 0.942)** | 0.781 (0.655, 0.875) |
|  | Virchow | 0.820 (0.735, 0.906) | 0.356 (0.274, 0.452) | 0.099 (0.048, 0.160) | 0.832 (0.766, 0.891) | 0.667 (0.526, 0.789) |
|  | Virchow2 | 0.905 (0.841, 0.968) | 0.752 (0.663, 0.833) | 0.188 (0.116, 0.264) | **0.898 (0.847, 0.949)** | **0.821 (0.712, 0.907)** |
|  | CRISP | **0.919 (0.867, 0.972)** | **0.762 (0.676, 0.846)** | **0.287 (0.196, 0.377)** | 0.883 (0.825, 0.934) | 0.805 (0.700, 0.894) |

Table S2.9. Performance of various models for the discrimination of benign and malignant abdomen lesions across different cohorts.

| Tasks | Data Split | | | | |
|---|---|---|---|---|---|
| | Internal-Train | Internal-Validation | Internal-Test | External-Test1 (NFH) | External-Test2 (ZSH) |
| PCNSLD | 434 | 179 | 328 | 461 | |
| CNSLC | 434 | 179 | 328 | 461 | |
| OCCGS | 253 | 112 | 134 | 166 (NFH) + 169 (ZSH) | |
| OCFGS | 253 | 112 | 134 | 166 (NFH) + 169 (ZSH) | |
| TNDD | 290 | 88 | 158 | | 192 |
| PGD | 302 | 96 | 172 | 495 | |
| SLMD | 262 | 87 | 150 | 233 | 196 |
| CLMD | 197 | 75 | 93 | 228 | |
| BSMA | 216 | 118 | 150 | 208 | 195 |

Supplementary Table S3.1. Dataset composition across key intraoperative decision-making tasks.

| Model | AUROC | Spe (at 90% Sen) | Spe (at 95% Sen) | Accuracy | F1 |
|---|---|---|---|---|---|
| CHIEF | 0.890±0.105 | 0.661±0.247 | 0.501±0.317 | 0.846±0.144 | 0.697±0.195 |
| CONCH | 0.935±0.065 | 0.757±0.214 | 0.580±0.214 | 0.903±0.118 | 0.799±0.150 |
| UNI | 0.932±0.069 | 0.722±0.266 | 0.603±0.274 | 0.890±0.134 | 0.782±0.140 |
| Virchow | 0.949±0.037 | 0.791±0.173 | 0.607±0.302 | 0.924±0.059 | 0.791±0.118 |
| Virchow2 | 0.960±0.033 | 0.795±0.214 | 0.634±0.249 | 0.936±0.047 | 0.829±0.097 |
| CRISP | **0.979±0.028** | **0.906±0.121** | **0.823±0.156** | **0.951±0.044** | **0.863±0.126** |

Table S3.2. Average performance of various models across 12 clinical binary classification tasks.

| Model | Macro AUC | Balanced Acc | Weighted F1 |
|---|---|---|---|
| CHIEF | 0.908±0.032 | 0.562±0.135 | 0.831±0.061 |
| CONCH | 0.937±0.022 | 0.724±0.093 | 0.855±0.055 |
| UNI | 0.929±0.043 | 0.730±0.159 | 0.869±0.070 |
| Virchow | 0.936±0.030 | 0.667±0.145 | 0.844±0.077 |
| Virchow2 | 0.945±0.032 | 0.740±0.155 | 0.856±0.092 |
| CRISP | **0.963±0.017** | **0.781±0.138** | **0.899±0.048** |

Table S3.3. Average performance of various models across 8 clinical multi-class classification tasks.

| Cohort | Model | AUROC | Spe (at 90% Sen) | Spe (at 95% Sen) | Accuracy | F1 |
|---|---|---|---|---|---|---|
| SYCC | CHIEF | 0.980 (0.957, 1.000) | 0.933 (0.905, 0.961) | 0.891 (0.854, 0.925) | 0.933 (0.905, 0.957) | 0.792 (0.701, 0.869) |
| | CONCH | 0.995 (0.988, 1.000) | **1.000 (1.000, 1.000)** | 0.926 (0.892, 0.956) | **0.991 (0.979, 1.000)** | **0.965 (0.918, 1.000)** |
| | UNI | 0.991 (0.983, 0.999) | 0.947 (0.921, 0.971) | 0.944 (0.915, 0.969) | 0.948 (0.924, 0.970) | 0.835 (0.745, 0.904) |
| | Virchow | 0.991 (0.980, 1.000) | 0.993 (0.982, 1.000) | 0.898 (0.863, 0.931) | 0.985 (0.970, 0.997) | 0.943 (0.877, 0.987) |
| | Virchow2 | 0.992 (0.981, 1.000) | 0.989 (0.976, 1.000) | 0.933 (0.903, 0.961) | 0.973 (0.954, 0.988) | 0.903 (0.833, 0.960) |
| | CRISP | **0.998 (0.995, 1.000)** | 0.982 (0.966, 0.996) | **0.982 (0.965, 0.996)** | 0.982 (0.966, 0.994) | 0.935 (0.874, 0.980) |
| NFH | CHIEF | 0.970 (0.947, 0.994) | 0.852 (0.821, 0.883) | 0.813 (0.779, 0.848) | 0.822 (0.787, 0.857) | 0.349 (0.243, 0.454) |
| | CONCH | 0.982 (0.954, 1.000) | **0.938 (0.914, 0.959)** | 0.692 (0.649, 0.734) | **0.983 (0.970, 0.993)** | **0.833 (0.698, 0.933)** |
| | UNI | 0.951 (0.906, 0.996) | 0.661 (0.617, 0.705) | 0.658 (0.616, 0.702) | 0.959 (0.939, 0.976) | 0.667 (0.500, 0.800) |
| | Virchow | 0.984 (0.969, 0.998) | 0.916 (0.888, 0.939) | 0.870 (0.837, 0.898) | 0.967 (0.950, 0.983) | 0.727 (0.578, 0.846) |
| | Virchow2 | 0.981 (0.961, 1.000) | 0.854 (0.819, 0.886) | 0.829 (0.793, 0.865) | 0.944 (0.922, 0.963) | 0.606 (0.449, 0.736) |
| | CRISP | **0.987 (0.975, 0.999)** | **0.938 (0.916, 0.959)** | 0.895 (0.867, 0.922) | 0.900 (0.874, 0.926) | 0.489 (0.346, 0.612) |

Table S3.4. Performance of various models for the PCNSLD task across different cohorts.

| Cohort | Model | Macro AUC | Balanced ACC | Weighted F1 | Class-specific AUCs ||||||
|---|---|---|---|---|---|---|---|---|---|---|
| | | | | | Glioma | MB | Inflammation | CN | MC | Lymphoma |
| SYCC | CHIEF | 0.947 | 0.608 | 0.888 | 0.976 | 0.977 | 0.887 | 0.854 | 0.993 | 0.993 |
| | CONCH | 0.966 | 0.699 | 0.911 | **0.991** | 0.98 | 0.972 | 0.862 | 0.994 | 0.996 |
| | UNI | 0.970 | 0.666 | 0.905 | 0.987 | 0.986 | 0.929 | **0.926** | 0.993 | **0.999** |
| | Virchow | 0.964 | 0.622 | 0.927 | **0.991** | 0.985 | 0.976 | 0.842 | **0.999** | 0.993 |
| | Virchow2 | 0.972 | **0.746** | 0.924 | 0.989 | 0.994 | 0.975 | 0.883 | 0.998 | 0.994 |
| | CRISP | **0.981** | 0.733 | **0.935** | 0.991 | **0.997** | **0.978** | 0.923 | 0.996 | 0.998 |
| NFH | CHIEF | 0.857 | 0.471 | 0.813 | 0.918 | **0.963** | 0.768 | 0.572 | 0.967 | 0.957 |
| | CONCH | 0.924 | **0.555** | 0.761 | 0.934 | 0.926 | 0.923 | 0.775 | **0.997** | **0.986** |
| | UNI | 0.834 | 0.385 | 0.744 | **0.94** | 0.843 | 0.795 | 0.561 | 0.942 | 0.924 |
| | Virchow | 0.895 | 0.394 | 0.775 | 0.911 | 0.937 | 0.908 | 0.651 | 0.989 | 0.972 |
| | Virchow2 | 0.904 | 0.417 | 0.733 | 0.919 | 0.94 | 0.931 | 0.658 | 0.984 | 0.991 |
| | CRISP | **0.942** | 0.470 | **0.816** | 0.933 | 0.959 | **0.97** | **0.815** | 0.994 | 0.984 |

Table S3.5. Performance of various models for the CNSLC task across different cohorts.

| Cohort | Model | Macro AUC | Balanced ACC | Weighted F1 | Class-specific AUCs | | |
|---|---|---|---|---|---|---|---|
| | | | | | Benign | Primary | Metastatic |
| SYCC | CHIEF | 0.941 | 0.768 | 0.882 | 0.964 | 0.958 | 0.9 |
| | CONCH | 0.961 | 0.836 | 0.914 | 0.971 | 0.954 | 0.959 |
| | UNI | 0.961 | 0.877 | 0.924 | 0.962 | 0.967 | 0.953 |
| | Virchow | 0.975 | 0.766 | 0.886 | **0.979** | 0.971 | **0.975** |
| | Virchow2 | 0.972 | **0.921** | **0.947** | 0.977 | **0.987** | 0.953 |
| | CRISP | **0.976** | 0.908 | 0.925 | 0.978 | 0.982 | 0.968 |
| NFH | CHIEF | 0.904 | 0.504 | 0.893 | 0.934 | 0.932 | 0.847 |
| | CONCH | 0.941 | 0.786 | 0.877 | 0.955 | 0.941 | 0.929 |
| | UNI | 0.949 | 0.850 | 0.926 | 0.933 | 0.921 | **0.994** |
| | Virchow | 0.953 | 0.667 | 0.901 | 0.953 | 0.947 | 0.96 |
| | Virchow2 | 0.961 | 0.688 | 0.892 | 0.953 | 0.948 | 0.982 |
| | CRISP | **0.968** | **0.858** | **0.930** | **0.974** | **0.963** | 0.968 |

Table S3.6. Performance of various models for the OCCGS task across different cohorts.

| Cohort | Model | Macro AUC | Balanced ACC | Weighted F1 | Class-specific AUCs | | | | |
|--------|-------|-----------|--------------|-------------|--------|------|------|------|------------|
| | | | | | Benign | SC | CCC | EA | Metastatic |
| SYCC | CHIEF | 0.934 | 0.592 | 0.820 | 0.964 | 0.953 | 0.936 | 0.919 | 0.901 |
| | CONCH | 0.943 | 0.776 | 0.906 | 0.98 | 0.97 | 0.86 | 0.934 | 0.973 |
| | UNI | 0.941 | 0.814 | 0.918 | 0.959 | 0.956 | 0.886 | 0.958 | 0.945 |
| | Virchow | 0.951 | 0.750 | 0.905 | 0.966 | 0.958 | 0.893 | 0.956 | 0.983 |
| | Virchow2 | 0.975 | **0.883** | **0.941** | **0.978** | **0.986** | 0.957 | **0.998** | 0.957 |
| | CRISP | **0.985** | 0.850 | 0.933 | **0.978** | 0.975 | **0.99** | 0.993 | **0.987** |
| NFZS | CHIEF | 0.892 | 0.332 | 0.854 | 0.927 | 0.955 | 0.81 | 0.918 | 0.85 |
| | CONCH | 0.933 | 0.622 | 0.832 | 0.941 | 0.977 | 0.914 | 0.895 | 0.935 |
| | UNI | 0.906 | 0.676 | 0.899 | 0.91 | 0.907 | 0.848 | 0.89 | 0.973 |
| | Virchow | 0.895 | 0.533 | 0.866 | 0.902 | 0.896 | 0.87 | 0.845 | 0.96 |
| | Virchow2 | 0.947 | 0.708 | 0.888 | 0.964 | 0.949 | 0.903 | 0.925 | **0.992** |
| | CRISP | **0.960** | **0.790** | **0.925** | **0.968** | **0.981** | **0.925** | **0.946** | 0.979 |

Table S3.7. Performance of various models for the OCCFS task across different cohorts.

| Cohort | Model | Macro AUC | Balanced ACC | Weighted F1 | Class-specific AUCs | | |
|---|---|---|---|---|---|---|---|
| | | | | | OB | FL | PTC |
| SYCC | CHIEF | 0.914 | 0.688 | 0.777 | 0.894 | 0.897 | 0.952 |
| | CONCH | 0.934 | 0.770 | 0.821 | 0.925 | 0.891 | 0.986 |
| | UNI | 0.943 | 0.790 | 0.785 | 0.925 | **0.927** | 0.978 |
| | Virchow | 0.925 | 0.792 | 0.719 | 0.899 | 0.924 | 0.953 |
| | Virchow2 | 0.935 | **0.820** | 0.806 | 0.916 | 0.919 | 0.97 |
| | CRISP | **0.951** | 0.773 | **0.830** | **0.938** | **0.927** | **0.989** |
| NFH | CHIEF | 0.873 | 0.529 | 0.719 | 0.953 | 0.767 | 0.901 |
| | CONCH | 0.895 | 0.744 | 0.820 | 0.938 | 0.847 | 0.902 |
| | UNI | 0.925 | 0.786 | 0.849 | 0.926 | 0.936 | 0.913 |
| | Virchow | 0.931 | 0.812 | 0.772 | 0.927 | **0.957** | 0.909 |
| | Virchow2 | 0.893 | 0.739 | 0.719 | 0.881 | 0.865 | 0.933 |
| | CRISP | **0.941** | **0.862** | **0.897** | **0.96** | 0.922 | **0.94** |

Table S3.8. Performance of various models for the TNDD task across different cohorts.

| Cohort | Model | AUROC | Spe (at 90% Sen) | Spe (at 95% Sen) | Accuracy | F1 |
|---|---|---|---|---|---|---|
| SYCC | CHIEF | 0.793 (0.726, 0.860) | 0.612 (0.505, 0.707) | 0.388 (0.291, 0.484) | 0.744 (0.680, 0.814) | 0.756 (0.675, 0.822) |
| | CONCH | 0.836 (0.778, 0.894) | 0.602 (0.495, 0.701) | 0.449 (0.347, 0.547) | 0.762 (0.698, 0.826) | 0.745 (0.658, 0.814) |
| | UNI | 0.804 (0.741, 0.867) | 0.551 (0.455, 0.649) | 0.500 (0.404, 0.604) | 0.721 (0.657, 0.785) | 0.747 (0.677, 0.816) |
| | Virchow | 0.904 (0.862, 0.946) | **0.673 (0.585, 0.761)** | **0.582 (0.490, 0.670)** | 0.802 (0.738, 0.860) | 0.795 (0.717, 0.854) |
| | Virchow2 | 0.885 (0.838, 0.932) | 0.612 (0.510, 0.702) | 0.480 (0.376, 0.571) | 0.820 (0.762, 0.878) | 0.780 (0.698, 0.849) |
| | CRISP | **0.905 (0.863, 0.947)** | **0.673 (0.577, 0.763)** | 0.551 (0.453, 0.652) | **0.831 (0.779, 0.884)** | **0.808 (0.732, 0.868)** |
| NFH | CHIEF | 0.628 (0.576, 0.679) | 0.413 (0.365, 0.459) | 0.364 (0.312, 0.412) | 0.489 (0.444, 0.531) | 0.407 (0.348, 0.467) |
| | CONCH | 0.775 (0.728, 0.822) | 0.483 (0.433, 0.531) | 0.431 (0.385, 0.477) | 0.592 (0.547, 0.636) | 0.445 (0.381, 0.508) |
| | UNI | 0.784 (0.735, 0.832) | 0.468 (0.419, 0.517) | 0.433 (0.384, 0.483) | 0.527 (0.485, 0.572) | 0.435 (0.378, 0.495) |
| | Virchow | 0.871 (0.824, 0.917) | 0.592 (0.544, 0.641) | 0.240 (0.200, 0.283) | 0.826 (0.788, 0.857) | 0.629 (0.547, 0.700) |
| | Virchow2 | 0.913 (0.879, 0.947) | 0.653 (0.606, 0.697) | 0.480 (0.435, 0.529) | 0.901 (0.875, 0.927) | 0.738 (0.663, 0.806) |
| | CRISP | **0.944 (0.918, 0.971)** | **0.720 (0.676, 0.764)** | **0.559 (0.507, 0.612)** | **0.956 (0.937, 0.974)** | **0.872 (0.814, 0.919)** |

Table S3.9. Performance of various models for the PGD task across different cohorts.

| Cohort | Model | AUROC | Spe (at 90% Sen) | Spe (at 95% Sen) | Accuracy | F1 |
|---|---|---|---|---|---|---|
| SYCC | CHIEF | 0.989 (0.975, 1.000) | 0.989 (0.962, 1.000) | 0.888 (0.819, 0.947) | 0.967 (0.933, 0.993) | 0.959 (0.920, 0.992) |
| | CONCH | 0.991 (0.978, 1.000) | 0.989 (0.963, 1.000) | 0.978 (0.944, 1.000) | 0.973 (0.947, 0.993) | 0.967 (0.932, 0.993) |
| | UNI | 0.994 (0.986, 1.000) | 0.989 (0.962, 1.000) | 0.989 (0.966, 1.000) | 0.980 (0.953, 1.000) | **0.976 (0.943, 1.000)** |
| | Virchow | 0.990 (0.978, 1.000) | 0.978 (0.940, 1.000) | 0.944 (0.897, 0.988) | 0.960 (0.927, 0.987) | 0.950 (0.909, 0.984) |
| | Virchow2 | 0.997 (0.991, 1.000) | **1.000 (1.000, 1.000)** | 0.989 (0.963, 1.000) | **0.980 (0.953, 1.000)** | 0.975 (0.945, 1.000) |
| | CRISP | **0.998 (0.994, 1.000)** | 0.989 (0.963, 1.000) | 0.978 (0.940, 1.000) | **0.980 (0.953, 1.000)** | 0.976 (0.945, 1.000) |
| NFH | CHIEF | 0.866 (0.781, 0.951) | 0.308 (0.247, 0.371) | 0.111 (0.068, 0.158) | 0.931 (0.897, 0.961) | 0.758 (0.633, 0.865) |
| | CONCH | 0.945 (0.897, 0.993) | 0.591 (0.522, 0.660) | 0.434 (0.364, 0.503) | 0.966 (0.940, 0.987) | 0.879 (0.780, 0.955) |
| | UNI | 0.909 (0.836, 0.983) | 0.333 (0.268, 0.396) | 0.146 (0.096, 0.197) | 0.936 (0.901, 0.966) | 0.795 (0.686, 0.889) |
| | Virchow | 0.916 (0.853, 0.979) | 0.455 (0.394, 0.526) | 0.288 (0.225, 0.354) | 0.940 (0.906, 0.970) | 0.800 (0.690, 0.895) |
| | Virchow2 | 0.945 (0.885, 1.000) | 0.692 (0.631, 0.760) | 0.182 (0.134, 0.241) | **0.970 (0.944, 0.991)** | **0.901 (0.821, 0.962)** |
| | CRISP | **0.969 (0.941, 0.998)** | **0.732 (0.671, 0.796)** | **0.687 (0.621, 0.748)** | 0.966 (0.940, 0.987) | 0.886 (0.793, 0.957) |
| ZSH | CHIEF | 0.971 (0.941, 1.000) | 0.870 (0.812, 0.924) | 0.675 (0.597, 0.743) | 0.969 (0.944, 0.990) | 0.923 (0.849, 0.977) |
| | CONCH | 0.957 (0.906, 1.000) | 0.974 (0.947, 0.994) | 0.558 (0.480, 0.641) | 0.964 (0.934, 0.990) | 0.918 (0.846, 0.970) |
| | UNI | 0.981 (0.958, 1.000) | **0.994 (0.980, 1.000)** | 0.675 (0.595, 0.747) | **0.980 (0.959, 0.995)** | 0.951 (0.892, 0.989) |
| | Virchow | 0.968 (0.943, 0.993) | 0.935 (0.893, 0.973) | 0.721 (0.653, 0.790) | 0.954 (0.923, 0.980) | 0.894 (0.815, 0.958) |
| | Virchow2 | 0.970 (0.940, 1.000) | 0.955 (0.919, 0.986) | 0.812 (0.746, 0.873) | 0.949 (0.918, 0.980) | 0.886 (0.812, 0.949) |
| | CRISP | **0.986 (0.963, 1.000)** | **0.994 (0.980, 1.000)** | **0.890 (0.840, 0.932)** | **0.980 (0.959, 0.995)** | **0.952 (0.895, 0.989)** |

Table S3.10. Performance of various models for the SLMD task across different cohorts.

| Cohort | Model | AUROC | Spe (at 90% Sen) | Accuracy | F1 |
|---|---|---|---|---|---|
| SYCC | CHIEF | 0.965 (0.929, 1.000) | 0.900 (0.826, 0.958) | 0.914 (0.849, 0.968) | 0.846 (0.732, 0.939) |
| | CONCH | 0.965 (0.913, 1.000) | **0.957 (0.908, 1.000)** | **0.957 (0.914, 0.989)** | **0.917 (0.818, 0.982)** |
| | UNI | 0.948 (0.881, 1.000) | 0.929 (0.863, 0.986) | 0.935 (0.882, 0.978) | 0.880 (0.773, 0.962) |
| | Virchow | 0.943 (0.861, 1.000) | 0.943 (0.881, 0.987) | 0.946 (0.903, 0.989) | 0.898 (0.791, 0.980) |
| | Virchow2 | 0.968 (0.918, 1.000) | 0.914 (0.842, 0.973) | 0.925 (0.871, 0.978) | 0.863 (0.750, 0.947) |
| | CRISP | **0.986 (0.966, 1.000)** | 0.943 (0.886, 0.987) | 0.946 (0.892, 0.989) | 0.898 (0.792, 0.978) |
| NFH | CHIEF | 0.914 (0.807, 1.000) | 0.288 (0.229, 0.344) | 0.969 (0.943, 0.991) | 0.741 (0.518, 0.903) |
| | CONCH | 0.954 (0.882, 1.000) | 0.516 (0.450, 0.584) | 0.961 (0.934, 0.982) | 0.727 (0.533, 0.875) |
| | UNI | 0.928 (0.813, 1.000) | 0.237 (0.185, 0.298) | 0.974 (0.952, 0.991) | 0.786 (0.592, 0.923) |
| | Virchow | 0.971 (0.922, 1.000) | 0.674 (0.615, 0.739) | 0.978 (0.956, 0.996) | 0.828 (0.647, 0.957) |
| | Virchow2 | 0.940 (0.827, 1.000) | 0.251 (0.192, 0.309) | 0.974 (0.952, 0.991) | 0.800 (0.600, 0.933) |
| | CRISP | **0.998 (0.995, 1.000)** | **0.981 (0.959, 0.995)** | **0.982 (0.965, 0.996)** | **0.867 (0.714, 0.973)** |

Table S3.11. Performance of various models for the CLMD task across different cohorts.

| Cohort | Model | AUROC | Spe (at 90% Sen) | Spe (at 95% Sen) | Accuracy | F1 |
|---|---|---|---|---|---|---|
| SYCC | CHIEF | 0.885 (0.821, 0.949) | 0.587 (0.504, 0.672) | 0.149 (0.090, 0.215) | 0.814 (0.750, 0.872) | 0.688 (0.568, 0.784) |
|  | CONCH | 0.960 (0.926, 0.994) | 0.860 (0.795, 0.917) | 0.504 (0.417, 0.597) | 0.865 (0.808, 0.917) | 0.764 (0.657, 0.854) |
|  | UNI | 0.962 (0.933, 0.992) | 0.860 (0.795, 0.923) | 0.669 (0.583, 0.750) | 0.878 (0.827, 0.923) | 0.776 (0.674, 0.865) |
|  | Virchow | 0.946 (0.907, 0.985) | 0.719 (0.637, 0.803) | 0.545 (0.458, 0.635) | 0.923 (0.878, 0.962) | 0.833 (0.719, 0.919) |
|  | Virchow2 | 0.978 (0.957, 0.998) | 0.876 (0.810, 0.932) | 0.736 (0.658, 0.811) | 0.891 (0.840, 0.936) | 0.800 (0.685, 0.889) |
|  | CRISP | **0.995 (0.989, 1.000)** | **0.983 (0.957, 1.000)** | **0.934 (0.885, 0.975)** | **0.949 (0.910, 0.981)** | **0.897 (0.822, 0.962)** |
| NFH | CHIEF | 0.813 (0.751, 0.876) | 0.656 (0.582, 0.727) | 0.489 (0.415, 0.556) | 0.697 (0.635, 0.760) | 0.462 (0.348, 0.569) |
|  | CONCH | 0.922 (0.880, 0.964) | 0.694 (0.627, 0.758) | 0.667 (0.604, 0.737) | 0.885 (0.837, 0.928) | 0.657 (0.517, 0.773) |
|  | UNI | 0.978 (0.961, 0.995) | 0.894 (0.844, 0.938) | 0.889 (0.840, 0.935) | 0.904 (0.861, 0.942) | 0.737 (0.610, 0.836) |
|  | Virchow | 0.956 (0.930, 0.983) | 0.872 (0.819, 0.918) | 0.861 (0.806, 0.914) | 0.880 (0.837, 0.923) | 0.691 (0.557, 0.796) |
|  | Virchow2 | 0.974 (0.936, 1.000) | 0.894 (0.849, 0.938) | 0.461 (0.392, 0.536) | **0.971 (0.947, 0.990)** | 0.897 (0.792, 0.966) |
|  | CRISP | **0.991 (0.979, 1.000)** | **0.972 (0.947, 0.994)** | **0.900 (0.855, 0.942)** | **0.971 (0.947, 0.990)** | **0.900 (0.807, 0.970)** |
| ZSH | CHIEF | 0.911 (0.830, 0.992) | 0.521 (0.446, 0.595) | 0.030 (0.006, 0.058) | 0.903 (0.862, 0.938) | 0.689 (0.545, 0.810) |
|  | CONCH | 0.935 (0.871, 0.999) | 0.485 (0.413, 0.553) | 0.325 (0.257, 0.395) | 0.933 (0.897, 0.964) | 0.772 (0.638, 0.884) |
|  | UNI | 0.954 (0.912, 0.997) | 0.799 (0.744, 0.855) | 0.491 (0.420, 0.569) | 0.944 (0.908, 0.974) | 0.800 (0.667, 0.906) |
|  | Virchow | 0.948 (0.911, 0.985) | 0.746 (0.677, 0.812) | 0.686 (0.613, 0.756) | 0.928 (0.892, 0.964) | 0.759 (0.640, 0.870) |
|  | Virchow2 | 0.974 (0.944, 1.000) | 0.846 (0.793, 0.898) | 0.645 (0.571, 0.714) | 0.938 (0.897, 0.969) | 0.800 (0.667, 0.900) |
|  | CRISP | **0.991 (0.980, 1.000)** | **0.964 (0.935, 0.994)** | **0.888 (0.834, 0.930)** | **0.964 (0.938, 0.990)** | **0.877 (0.772, 0.955)** |

Table S3.12. Performance of various models for the BSMA task across different cohorts.

| Organ | Annual incidence in the U.S. | Classification |
|---|---|---|
| Breast | 319,750 | Common |
| Lung | 226,650 | |
| Colorectum | 154,270 | |
| H&N | 113,840 | |
| Skin | 112,690 | |
| Uterus | 69,120 | |
| Pancreas | 67,440 | |
| Thyroid | 44,020 | |
| Liver | 42,240 | |
| Brain | 24,820 | Rare |
| Ovary | 20,890 | |
| Soft Tissue | 13,520 | |
| Cervix | 13,360 | |
| Bile Duct | 12,610 | |
| Pleura | 6,030 | |

Table S4.1. Definition of rare cancers and estimated annual incidence.

| Subset | Organ | Slides | PositiveCase |
|---|---|---|---|
| Train | Brain | 234 | 143 |
| | Breast | 272 | 71 |
| | H&N | 261 | 106 |
| | Lung | 257 | 96 |
| | Ovary | 253 | 111 |
| | Thyroid | 296 | 125 |
| | **Sum** | **1573** | **652** |
| Val | Brain | 108 | 63 |
| | Breast | 97 | 35 |
| | H&N | 87 | 30 |
| | Lung | 110 | 43 |
| | Ovary | 112 | 43 |
| | Thyroid | 89 | 42 |
| | **Sum** | **603** | **256** |
| Internal-Test | Brain | 165 | 114 |
| | Breast | 133 | 50 |
| | H&N | 131 | 54 |
| | Lung | 151 | 67 |
| | Ovary | 140 | 56 |
| | Thyroid | 175 | 80 |
| | **Sum** | **895** | **421** |

Table S4.2. Data distribution of the SYCC cohort in pan-cancer classification.

| Organ | Cases | Slides | PositiveCase |
|---|---|---|---|
| Brain | 30 | 38 | 28 |
| Breast | 186 | 585 | 23 |
| H&N | 207 | 626 | 56 |
| Lung | 201 | 253 | 167 |
| Ovary | 38 | 74 | 6 |
| Thyroid | 182 | 284 | 78 |
| Bile Duct | 35 | 73 | 5 |
| Cervix | 43 | 163 | 5 |
| Colorectum | 7 | 20 | 2 |
| Liver | 13 | 30 | 8 |
| Lymph Node | 486 | 2152 | 152 |
| Pancreas | 13 | 29 | 3 |
| Pleura | 7 | 13 | 2 |
| Skin | 42 | 265 | 25 |
| Soft Tissue | 29 | 55 | 13 |
| Uterus | 145 | 307 | 38 |
| Else | 12 | 22 | 3 |
| **Sum** | **1676** | **4989** | **614** |

Table S4.3. Data distribution of the ZSH cohort in pan-cancer classification.

| Organ | Cases | Slides | PositiveCase |
|---|---|---|---|
| Brain | 669 | 700 | 587 |
| Breast | 682 | 898 | 169 |
| H&N | 293 | 514 | 73 |
| Lung | 561 | 760 | 413 |
| Ovary | 119 | 209 | 27 |
| Thyroid | 206 | 294 | 78 |
| Bile Duct | 35 | 48 | 8 |
| Cervix | 77 | 117 | 20 |
| Colorectum | 146 | 167 | 69 |
| Liver | 25 | 29 | 11 |
| Lymph Node | 241 | 495 | 48 |
| Pancreas | 8 | 9 | 3 |
| Pleura | 11 | 13 | 4 |
| Skin | 56 | 147 | 15 |
| Soft Tissue | 50 | 77 | 14 |
| Uterus | 35 | 59 | 2 |
| Else | 130 | 161 | 24 |
| **Sum** | **3344** | **4697** | **1565** |

Table S4.4. Data distribution of the NFH cohort in pan-cancer classification.

| Organ | Cases | Slides | PositiveCase |
|---|---|---|---|
| Brain | 20 | 21 | 13 |
| Breast | 314 | 356 | 92 |
| H&N | 25 | 42 | 5 |
| Lung | 151 | 207 | 136 |
| Ovary | 45 | 73 | 20 |
| Thyroid | 182 | 244 | 145 |
| Bile Duct | 3 | 4 | 1 |
| Cervix | 10 | 11 | 1 |
| Colorectum | 4 | 4 | 2 |
| Liver | 2 | 2 | 1 |
| Lymph Node | 24 | 41 | 5 |
| Pancreas | 2 | 2 | 1 |
| Pleura | 6 | 6 | 3 |
| Skin | 8 | 9 | 2 |
| Soft Tissue | 5 | 6 | 1 |
| Uterus | 39 | 56 | 1 |
| Else | 7 | 10 | 3 |
| **Sum** | **847** | **1094** | **432** |

Table S4.5. Data distribution of the HDGZ cohort in pan-cancer classification.

| Cohort | Model | AUROC | Spe (at 95% Sen) | Accuracy | F1 |
|---|---|---|---|---|---|
| SYCC | CHIEF | 0.954 (0.941,0.967) | 0.811 (0.779, 0.846) | 0.903 (0.885, 0.919) | 0.897 (0.875, 0.916) |
| | CONCH | 0.976 (0.969,0.984) | 0.850 (0.821, 0.880) | 0.932 (0.914, 0.946) | 0.928 (0.911, 0.945) |
| | UNI | 0.978 (0.970,0.986) | 0.880 (0.852, 0.908) | 0.939 (0.924, 0.954) | 0.937 (0.920, 0.951) |
| | Virchow | 0.976 (0.967,0.985) | 0.916 (0.893, 0.938) | 0.945 (0.932, 0.959) | 0.942 (0.925, 0.957) |
| | Virchow2 | 0.988 (0.982,0.993) | 0.923 (0.901, 0.947) | **0.958 (0.945, 0.970)** | **0.955 (0.941, 0.968)** |
| | CRISP | **0.990 (0.985,0.994)** | **0.940 (0.920, 0.960)** | 0.957 (0.945, 0.968) | 0.954 (0.941, 0.969) |
| ZSH | CHIEF | 0.755 (0.731,0.779) | 0.240 (0.214, 0.266) | 0.705 (0.685, 0.727) | 0.618 (0.587, 0.650) |
| | CONCH | 0.886 (0.868,0.903) | 0.429 (0.400, 0.461) | 0.838 (0.821, 0.855) | 0.763 (0.735, 0.791) |
| | UNI | 0.876 (0.858,0.893) | 0.439 (0.411, 0.470) | 0.800 (0.780, 0.819) | 0.737 (0.709, 0.763) |
| | Virchow | 0.752 (0.729,0.776) | 0.339 (0.312, 0.367) | 0.693 (0.671, 0.715) | 0.660 (0.633, 0.687) |
| | Virchow2 | 0.879 (0.863,0.896) | 0.521 (0.492, 0.551) | 0.791 (0.770, 0.811) | 0.747 (0.717, 0.769) |
| | CRISP | **0.953 (0.943,0.963)** | **0.723 (0.699, 0.750)** | **0.893 (0.878, 0.907)** | **0.856 (0.836, 0.878)** |
| NFH | CHIEF | 0.840 (0.827,0.853) | 0.365 (0.343, 0.388) | 0.771 (0.757, 0.786) | 0.757 (0.741, 0.772) |
| | CONCH | 0.801 (0.786,0.816) | 0.165 (0.148, 0.182) | 0.748 (0.733, 0.762) | 0.718 (0.699, 0.734) |
| | UNI | 0.900 (0.889,0.910) | 0.485 (0.463, 0.508) | 0.833 (0.820, 0.845) | 0.819 (0.804, 0.834) |
| | Virchow | 0.878 (0.866,0.890) | 0.367 (0.344, 0.388) | 0.809 (0.795, 0.822) | 0.785 (0.768, 0.800) |
| | Virchow2 | 0.952 (0.945,0.959) | 0.720 (0.700, 0.743) | 0.894 (0.885, 0.904) | 0.886 (0.873, 0.897) |
| | CRISP | **0.973 (0.968,0.977)** | **0.879 (0.864, 0.894)** | **0.919 (0.909, 0.927)** | **0.915 (0.904, 0.925)** |
| HDGZ | CHIEF | 0.757 (0.724,0.789) | 0.161 (0.128, 0.197) | 0.711 (0.678, 0.741) | 0.740 (0.708, 0.771) |
| | CONCH | 0.796 (0.765,0.826) | 0.152 (0.117, 0.190) | 0.744 (0.713, 0.772) | 0.721 (0.686, 0.755) |
| | UNI | 0.907 (0.886,0.927) | 0.523 (0.475, 0.569) | 0.848 (0.823, 0.871) | 0.850 (0.824, 0.876) |
| | Virchow | 0.847 (0.820,0.873) | 0.299 (0.257, 0.342) | 0.787 (0.761, 0.816) | 0.789 (0.758, 0.819) |
| | Virchow2 | 0.869 (0.845,0.892) | 0.388 (0.345, 0.435) | 0.803 (0.774, 0.830) | 0.796 (0.763, 0.824) |
| | CRISP | **0.960 (0.948,0.973)** | **0.778 (0.739, 0.819)** | **0.913 (0.896, 0.932)** | **0.912 (0.891, 0.932)** |

Table S4.6. Comparative overall performance of different models for pan-cancer classification across four cohorts.

| Cohort | Model | AUROC | Spe (at 95% Sen) | Accuracy | F1 |
|---|---|---|---|---|---|
| SYCC | CHIEF | 0.992 | 0.976 | 0.967 | 0.963 |
| | CONCH | 0.995 | **1.000** | 0.980 | 0.977 |
| | UNI | 0.998 | **1.000** | 0.980 | 0.977 |
| | Virchow | 0.997 | **1.000** | 0.987 | 0.985 |
| | Virchow2 | 0.999 | **1.000** | 0.987 | 0.985 |
| | CRISP | **1.000** | **1.000** | **1.000** | **1.000** |
| ZSH | CHIEF | 0.856 | 0.441 | 0.861 | 0.915 |
| | CONCH | 0.898 | 0.471 | 0.756 | 0.829 |
| | UNI | 0.949 | 0.735 | 0.905 | 0.941 |
| | Virchow | 0.940 | 0.706 | 0.905 | 0.942 |
| | Virchow2 | 0.961 | 0.853 | 0.900 | 0.937 |
| | CRISP | **0.983** | **0.882** | **0.925** | **0.953** |
| NFH | CHIEF | 0.897 | 0.561 | 0.852 | 0.896 |
| | CONCH | 0.906 | 0.466 | 0.786 | 0.837 |
| | UNI | 0.908 | 0.426 | 0.813 | 0.860 |
| | Virchow | 0.906 | 0.351 | 0.845 | 0.888 |
| | Virchow2 | 0.944 | 0.770 | 0.893 | 0.925 |
| | CRISP | **0.965** | **0.858** | **0.906** | **0.933** |
| HDGZ | CHIEF | 0.830 | 0.067 | 0.781 | 0.864 |
| | CONCH | 0.775 | 0.133 | 0.715 | 0.815 |
| | UNI | 0.959 | 1.000 | **0.960** | **0.977** |
| | Virchow | 0.885 | 0.267 | 0.881 | 0.930 |
| | Virchow2 | 0.887 | 0.333 | 0.868 | 0.922 |
| | CRISP | **0.984** | **0.933** | 0.934 | 0.962 |
| Overall | CHIEF | 0.901 (0.881,0.921) | 0.523 (0.466, 0.580) | 0.842 (0.820, 0.863) | 0.888 (0.870, 0.903) |
| | CONCH | 0.922 (0.905,0.938) | 0.580 (0.526, 0.635) | 0.820 (0.795, 0.842) | 0.866 (0.846, 0.884) |
| | UNI | 0.942 (0.929,0.955) | 0.673 (0.616, 0.723) | 0.861 (0.839, 0.880) | 0.900 (0.883, 0.916) |
| | Virchow | 0.931 (0.916,0.946) | 0.594 (0.536, 0.654) | 0.852 (0.832, 0.872) | 0.893 (0.874, 0.908) |
| | Virchow2 | 0.959 (0.947,0.971) | 0.811 (0.766, 0.857) | 0.904 (0.884, 0.921) | 0.933 (0.920, 0.946) |
| | CRISP | **0.977 (0.968,0.987)** | **0.911 (0.878, 0.944)** | **0.933 (0.919, 0.948)** | **0.954 (0.943, 0.965)** |

Table S4.7. Comparative performance of models in lung subgroup of pan-cancer classification task.

| Cohort | Model | AUROC | Spe (at 95% Sen) | Accuracy | F1 |
|---|---|---|---|---|---|
| SYCC | CHIEF | 0.971 | 0.855 | 0.887 | 0.870 |
| | CONCH | 0.978 | 0.867 | 0.910 | 0.891 |
| | UNI | 0.997 | **1.000** | 0.985 | 0.980 |
| | Virchow | 0.975 | 0.747 | 0.947 | 0.931 |
| | Virchow2 | **0.998** | **1.000** | **0.992** | **0.990** |
| | CRISP | 0.997 | **1.000** | **0.992** | **0.990** |
| ZSH | CHIEF | 0.811 | 0.466 | 0.828 | 0.484 |
| | CONCH | 0.881 | 0.577 | 0.812 | 0.533 |
| | UNI | 0.886 | 0.362 | 0.801 | 0.519 |
| | Virchow | 0.813 | 0.466 | 0.753 | 0.439 |
| | Virchow2 | 0.894 | 0.687 | 0.828 | 0.543 |
| | CRISP | **0.961** | **0.939** | **0.941** | **0.800** |
| NFH | CHIEF | 0.789 | 0.172 | 0.770 | 0.590 |
| | CONCH | 0.912 | 0.520 | 0.868 | 0.755 |
| | UNI | 0.932 | 0.651 | 0.871 | 0.771 |
| | Virchow | 0.919 | 0.493 | 0.908 | 0.807 |
| | Virchow2 | 0.970 | 0.797 | 0.933 | 0.873 |
| | CRISP | **0.983** | **0.922** | **0.956** | **0.913** |
| HDGZ | CHIEF | 0.808 | 0.419 | 0.736 | 0.641 |
| | CONCH | 0.817 | 0.180 | 0.764 | 0.667 |
| | UNI | 0.939 | **0.505** | 0.908 | 0.840 |
| | Virchow | 0.880 | 0.455 | 0.841 | 0.740 |
| | Virchow2 | 0.902 | 0.432 | 0.876 | 0.785 |
| | CRISP | **0.942** | **0.505** | **0.936** | **0.884** |
| Overall | CHIEF | 0.820 (0.792,0.847) | 0.242 (0.216, 0.269) | 0.794 (0.772, 0.815) | 0.626 (0.584, 0.668) |
| | CONCH | 0.870 (0.845,0.894) | 0.328 (0.298, 0.356) | 0.824 (0.805, 0.844) | 0.680 (0.639, 0.720) |
| | UNI | 0.927 (0.911,0.942) | 0.665 (0.635, 0.694) | 0.867 (0.847, 0.884) | 0.763 (0.727, 0.795) |
| | Virchow | 0.892 (0.870,0.914) | 0.398 (0.366, 0.427) | 0.858 (0.840, 0.876) | 0.726 (0.688, 0.760) |
| | Virchow2 | 0.952 (0.939,0.966) | 0.663 (0.633, 0.691) | 0.910 (0.893, 0.925) | 0.833 (0.804, 0.861) |
| | CRISP | **0.970 (0.959,0.981)** | **0.863 (0.841, 0.883)** | **0.937 (0.924, 0.950)** | **0.880 (0.855, 0.905)** |

Table S4.8. Comparative performance of models in breast subgroup of pan-cancer classification task.

| Cohort | Model | AUROC | Spe (at 95% Sen) | Accuracy | F1 |
|---|---|---|---|---|---|
| SYCC | CHIEF | 0.873 | 0.411 | 0.834 | 0.810 |
| | CONCH | 0.908 | 0.505 | 0.863 | 0.844 |
| | UNI | 0.938 | 0.568 | 0.874 | 0.867 |
| | Virchow | 0.932 | 0.653 | 0.874 | 0.867 |
| | Virchow2 | 0.946 | **0.716** | **0.897** | **0.873** |
| | CRISP | **0.953** | 0.653 | 0.891 | 0.871 |
| ZSH | CHIEF | 0.772 | 0.317 | 0.703 | 0.703 |
| | CONCH | 0.898 | 0.356 | 0.852 | 0.821 |
| | UNI | 0.879 | 0.481 | 0.824 | 0.778 |
| | Virchow | 0.876 | 0.442 | 0.808 | 0.793 |
| | Virchow2 | 0.946 | 0.692 | 0.868 | 0.846 |
| | CRISP | **0.965** | **0.788** | **0.912** | **0.899** |
| NFH | CHIEF | 0.880 | 0.484 | 0.816 | 0.753 |
| | CONCH | 0.903 | 0.250 | 0.888 | 0.835 |
| | UNI | 0.877 | 0.156 | 0.864 | 0.797 |
| | Virchow | 0.860 | 0.250 | 0.806 | 0.765 |
| | Virchow2 | 0.954 | 0.609 | 0.917 | 0.881 |
| | CRISP | **0.974** | **0.859** | **0.942** | **0.923** |
| HDGZ | CHIEF | 0.883 | 0.351 | 0.813 | 0.871 |
| | CONCH | 0.742 | 0.189 | 0.665 | 0.749 |
| | UNI | 0.937 | 0.649 | 0.857 | 0.902 |
| | Virchow | 0.902 | 0.459 | 0.863 | 0.908 |
| | Virchow2 | 0.920 | 0.541 | 0.835 | 0.886 |
| | CRISP | **0.959** | **0.865** | **0.912** | **0.942** |
| Overall | CHIEF | 0.865 (0.839,0.891) | 0.382 (0.332, 0.433) | 0.805 (0.776, 0.832) | 0.800 (0.767, 0.830) |
| | CONCH | 0.792 (0.760,0.824) | 0.245 (0.200, 0.291) | 0.734 (0.703, 0.765) | 0.724 (0.687, 0.758) |
| | UNI | 0.913 (0.892,0.934) | 0.503 (0.450, 0.553) | 0.860 (0.835, 0.885) | 0.854 (0.825, 0.880) |
| | Virchow | 0.848 (0.820,0.877) | 0.245 (0.203, 0.290) | 0.791 (0.761, 0.820) | 0.792 (0.759, 0.822) |
| | Virchow2 | 0.926 (0.908,0.945) | 0.607 (0.557, 0.660) | 0.860 (0.835, 0.885) | 0.861 (0.833, 0.886) |
| | CRISP | **0.965 (0.954,0.977)** | **0.805 (0.762, 0.845)** | **0.917 (0.897, 0.937)** | **0.918 (0.897, 0.938)** |

Table S4.9. Comparative performance of models in thyroid subgroup of pan-cancer classification task.

| Cohort | Model | AUROC | Spe (at 95% Sen) | Accuracy | F1 |
|---|---|---|---|---|---|
| SYCC | CHIEF | 0.947 | 0.584 | 0.916 | 0.897 |
| | CONCH | 0.985 | 0.792 | 0.962 | 0.952 |
| | UNI | 0.987 | 0.883 | 0.962 | 0.952 |
| | Virchow | 0.989 | 0.935 | 0.954 | 0.947 |
| | Virchow2 | **1.000** | **1.000** | **0.992** | **0.991** |
| | CRISP | 0.996 | 0.935 | 0.977 | 0.971 |
| ZSH | CHIEF | 0.771 | 0.126 | 0.725 | 0.590 |
| | CONCH | 0.861 | 0.338 | 0.787 | 0.676 |
| | UNI | 0.853 | 0.464 | 0.816 | 0.703 |
| | Virchow | 0.848 | 0.576 | 0.787 | 0.681 |
| | Virchow2 | 0.885 | 0.543 | 0.744 | 0.667 |
| | CRISP | **0.957** | **0.874** | **0.899** | **0.837** |
| NFH | CHIEF | 0.713 | 0.127 | 0.700 | 0.532 |
| | CONCH | 0.794 | 0.200 | 0.768 | 0.580 |
| | UNI | 0.850 | 0.273 | 0.826 | 0.687 |
| | Virchow | 0.791 | 0.282 | 0.706 | 0.578 |
| | Virchow2 | 0.910 | 0.518 | 0.874 | 0.764 |
| | CRISP | **0.921** | **0.541** | **0.891** | **0.787** |
| HDGZ | CHIEF | 0.650 | 0.500 | 0.600 | 0.500 |
| | CONCH | 0.520 | 0.250 | 0.400 | 0.400 |
| | UNI | **0.820** | **0.550** | 0.800 | 0.615 |
| | Virchow | 0.480 | 0.150 | 0.400 | 0.211 |
| | Virchow2 | 0.530 | 0.200 | 0.400 | 0.118 |
| | CRISP | **0.820** | **0.550** | **0.920** | **0.750** |
| Overall | CHIEF | 0.799 (0.760,0.838) | 0.237 (0.200, 0.274) | 0.720 (0.688, 0.755) | 0.613 (0.562, 0.663) |
| | CONCH | 0.860 (0.827,0.892) | 0.361 (0.317, 0.404) | 0.802 (0.771, 0.832) | 0.680 (0.626, 0.732) |
| | UNI | 0.885 (0.857,0.913) | 0.511 (0.464, 0.558) | 0.828 (0.799, 0.857) | 0.718 (0.668, 0.765) |
| | Virchow | 0.853 (0.820,0.886) | 0.436 (0.391, 0.482) | 0.747 (0.715, 0.777) | 0.667 (0.621, 0.713) |
| | Virchow2 | 0.918 (0.893,0.942) | 0.530 (0.488, 0.578) | 0.855 (0.826, 0.881) | 0.774 (0.730, 0.816) |
| | CRISP | **0.951 (0.933,0.970)** | **0.737 (0.693, 0.778)** | **0.907 (0.884, 0.928)** | **0.843 (0.803, 0.881)** |

Table S4.10. Comparative performance of models in H&N subgroup of pan-cancer classification task.

| Cohort | Model | AUROC | Spe (at 95% Sen) | Accuracy | F1 |
|---|---|---|---|---|---|
| SYCC | CHIEF | 0.963 | 0.902 | 0.945 | 0.960 |
| | CONCH | 0.997 | 0.980 | 0.964 | 0.973 |
| | UNI | 0.995 | 0.980 | 0.970 | 0.978 |
| | Virchow | 0.994 | 0.980 | 0.970 | 0.978 |
| | Virchow2 | **0.999** | **1.000** | **0.994** | **0.996** |
| | CRISP | **0.999** | 1.000 | 0.988 | 0.991 |
| ZSH | CHIEF | 0.929 | 0.000 | 0.933 | 0.963 |
| | CONCH | 0.714 | 0.000 | 0.600 | 0.727 |
| | UNI | 0.946 | 0.500 | 0.933 | 0.963 |
| | Virchow | 0.786 | 0.000 | 0.800 | 0.880 |
| | Virchow2 | 0.946 | 0.500 | 0.933 | 0.963 |
| | CRISP | **1.000** | **1.000** | **1.000** | **1.000** |
| NFH | CHIEF | 0.923 | 0.573 | 0.753 | 0.837 |
| | CONCH | 0.752 | 0.098 | 0.638 | 0.747 |
| | UNI | 0.910 | 0.561 | 0.825 | 0.892 |
| | Virchow | 0.927 | 0.512 | 0.833 | 0.896 |
| | Virchow2 | 0.959 | **0.756** | 0.861 | 0.915 |
| | CRISP | **0.961** | 0.744 | **0.886** | **0.932** |
| HDGZ | CHIEF | 0.725 | 0.143 | 0.800 | 0.857 |
| | CONCH | 0.659 | 0.143 | 0.700 | 0.700 |
| | UNI | 0.868 | 0.571 | 0.850 | 0.897 |
| | Virchow | 0.626 | 0.000 | 0.650 | 0.667 |
| | Virchow2 | 0.945 | 0.286 | 0.950 | 0.960 |
| | CRISP | **1.000** | **1.000** | **1.000** | **1.000** |
| Overall | CHIEF | 0.925 (0.903,0.947) | 0.641 (0.568, 0.720) | 0.836 (0.811, 0.860) | 0.894 (0.875, 0.910) |
| | CONCH | 0.845 (0.814,0.876) | 0.373 (0.297, 0.449) | 0.699 (0.669, 0.730) | 0.788 (0.763, 0.813) |
| | UNI | 0.939 (0.921,0.957) | 0.697 (0.622, 0.771) | 0.855 (0.834, 0.877) | 0.908 (0.891, 0.923) |
| | Virchow | 0.944 (0.929,0.958) | 0.606 (0.522, 0.694) | 0.861 (0.838, 0.882) | 0.911 (0.895, 0.926) |
| | Virchow2 | 0.973 (0.963,0.983) | 0.824 (0.756, 0.888) | 0.911 (0.891, 0.930) | 0.945 (0.932, 0.957) |
| | CRISP | **0.977 (0.967,0.987)** | **0.852 (0.789, 0.907)** | **0.917 (0.899, 0.934)** | **0.949 (0.937, 0.960)** |

Table S4.11. Comparative performance of models in brain subgroup of pan-cancer classification task.

| Cohort | Model | AUROC | Spe (at 95% Sen) | Accuracy | F1 |
|---|---|---|---|---|---|
| SYCC | CHIEF | 0.984 | 0.905 | 0.936 | 0.924 |
| | CONCH | 0.998 | 0.988 | 0.986 | 0.982 |
| | UNI | 0.989 | 0.976 | 0.971 | 0.964 |
| | Virchow | 0.992 | 0.976 | 0.979 | 0.972 |
| | Virchow2 | **0.999** | **1.000** | **0.993** | **0.991** |
| | CRISP | **0.999** | **1.000** | **0.993** | **0.991** |
| ZSH | CHIEF | 0.677 | 0.250 | 0.711 | 0.421 |
| | CONCH | 0.974 | 0.938 | 0.947 | 0.857 |
| | UNI | 0.891 | 0.531 | 0.842 | 0.625 |
| | Virchow | 0.953 | 0.875 | 0.895 | 0.750 |
| | Virchow2 | **0.995** | **0.969** | **0.974** | **0.923** |
| | CRISP | 0.984 | **0.969** | **0.974** | **0.923** |
| NFH | CHIEF | 0.925 | 0.837 | 0.866 | 0.765 |
| | CONCH | 0.987 | 0.913 | 0.966 | 0.926 |
| | UNI | 0.952 | 0.728 | 0.891 | 0.787 |
| | Virchow | 0.980 | 0.924 | 0.924 | 0.857 |
| | Virchow2 | 0.979 | 0.924 | 0.966 | 0.926 |
| | CRISP | **0.994** | **0.978** | **0.983** | **0.964** |
| HDGZ | CHIEF | 0.812 | 0.360 | 0.800 | 0.769 |
| | CONCH | 0.898 | 0.440 | 0.867 | 0.833 |
| | UNI | 0.922 | 0.600 | 0.867 | 0.842 |
| | Virchow | 0.912 | 0.560 | 0.844 | 0.829 |
| | Virchow2 | 0.924 | 0.560 | **0.889** | 0.872 |
| | CRISP | **0.956** | **0.840** | **0.889** | 0.884 |
| Overall | CHIEF | 0.928 (0.899,0.958) | 0.738 (0.680, 0.794) | 0.842 (0.804, 0.877) | 0.789 (0.731, 0.842) |
| | CONCH | 0.959 (0.940,0.978) | 0.807 (0.758, 0.855) | 0.906 (0.871, 0.936) | 0.857 (0.806, 0.902) |
| | UNI | 0.966 (0.947,0.984) | 0.803 (0.752, 0.857) | 0.912 (0.880, 0.942) | 0.862 (0.814, 0.908) |
| | Virchow | 0.959 (0.938,0.981) | 0.760 (0.703, 0.815) | 0.898 (0.863, 0.927) | 0.854 (0.804, 0.898) |
| | Virchow2 | 0.981 (0.969,0.993) | 0.893 (0.853, 0.931) | 0.950 (0.927, 0.971) | 0.922 (0.883, 0.955) |
| | CRISP | **0.990 (0.982,0.998)** | **0.966 (0.941, 0.987)** | **0.962 (0.939, 0.980)** | **0.942 (0.908, 0.971)** |

Table S4.12. Comparative performance of models in ovary subgroup of pan-cancer classification task.

| Cohort | Model | AUROC | Spe (at 95% Sen) | Accuracy | F1 |
|---|---|---|---|---|---|
| ZSH | CHIEF | 0.745 | 0.269 | 0.667 | 0.593 |
| | CONCH | 0.855 | 0.317 | 0.856 | 0.731 |
| | UNI | 0.813 | 0.296 | 0.811 | 0.638 |
| | Virchow | 0.617 | 0.132 | 0.628 | 0.478 |
| | Virchow2 | 0.792 | 0.278 | 0.796 | 0.602 |
| | CRISP | **0.936** | **0.614** | **0.901** | **0.834** |
| NFH | CHIEF | 0.805 | 0.301 | 0.776 | 0.578 |
| | CONCH | 0.834 | 0.223 | 0.876 | 0.688 |
| | UNI | 0.877 | 0.534 | 0.780 | 0.619 |
| | Virchow | 0.872 | 0.477 | 0.838 | 0.649 |
| | Virchow2 | 0.939 | 0.699 | 0.909 | 0.780 |
| | CRISP | **0.956** | **0.715** | **0.929** | **0.835** |
| HDGZ | CHIEF | 0.621 | 0.211 | 0.792 | 0.444 |
| | CONCH | 0.800 | 0.316 | 0.875 | 0.667 |
| | UNI | 0.895 | 0.684 | 0.750 | 0.625 |
| | Virchow | 0.937 | 0.842 | 0.875 | 0.769 |
| | Virchow2 | **0.979** | **0.895** | **0.917** | **0.833** |
| | CRISP | 0.968 | 0.842 | 0.875 | 0.769 |
| Overall | CHIEF | 0.765 (0.728,0.802) | 0.273 (0.235, 0.309) | 0.699 (0.668, 0.732) | 0.569 (0.511, 0.617) |
| | CONCH | 0.855 (0.820,0.890) | 0.249 (0.213, 0.287) | 0.874 (0.850, 0.896) | 0.726 (0.673, 0.778) |
| | UNI | 0.829 (0.796,0.862) | 0.425 (0.383, 0.464) | 0.688 (0.658, 0.722) | 0.602 (0.554, 0.646) |
| | Virchow | 0.727 (0.687,0.766) | 0.273 (0.237, 0.312) | 0.601 (0.565, 0.634) | 0.534 (0.489, 0.579) |
| | Virchow2 | 0.840 (0.809,0.872) | 0.390 (0.349, 0.431) | 0.738 (0.707, 0.770) | 0.620 (0.571, 0.670) |
| | CRISP | **0.944 (0.927,0.962)** | **0.707 (0.665, 0.743)** | **0.911 (0.889, 0.928)** | **0.829 (0.786, 0.869)** |

Table S4.13. Comparative performance of models in LN subgroup of pan-cancer classification task.

| Cohort | Model | AUROC | Spe (at 95% Sen) | Accuracy | F1 |
|---|---|---|---|---|---|
| ZSH | CHIEF | 0.878 | 0.346 | 0.883 | 0.767 |
| | CONCH | 0.902 | 0.187 | 0.862 | 0.767 |
| | UNI | 0.902 | 0.449 | 0.855 | 0.747 |
| | Virchow | 0.874 | 0.393 | 0.828 | 0.725 |
| | Virchow2 | 0.945 | 0.720 | 0.876 | 0.800 |
| | CRISP | **0.966** | **0.860** | **0.890** | **0.826** |
| NFH | CHIEF | 0.939 | 0.939 | 0.943 | 0.667 |
| | CONCH | 0.955 | 0.909 | 0.914 | 0.571 |
| | UNI | 0.939 | 0.879 | 0.886 | 0.500 |
| | Virchow | 0.955 | 0.909 | 0.914 | 0.571 |
| | Virchow2 | 0.727 | 0.636 | 0.657 | 0.250 |
| | CRISP | **1.000** | **1.000** | **1.000** | **1.000** |
| HDGZ | CHIEF | 0.974 | 0.974 | 0.974 | 0.667 |
| | CONCH | **1.000** | **1.000** | **1.000** | **1.000** |
| | UNI | 0.842 | 0.842 | 0.846 | 0.250 |
| | Virchow | **1.000** | **1.000** | **1.000** | **1.000** |
| | Virchow2 | **1.000** | **1.000** | **1.000** | **1.000** |
| | CRISP | **1.000** | **1.000** | **1.000** | **1.000** |
| Overall | CHIEF | 0.843 (0.781,0.906) | 0.562 (0.486, 0.637) | 0.840 (0.790, 0.886) | 0.624 (0.494, 0.738) |
| | CONCH | 0.911 (0.851,0.972) | 0.517 (0.443, 0.584) | 0.854 (0.808, 0.895) | 0.692 (0.589, 0.790) |
| | UNI | 0.872 (0.814,0.931) | 0.646 (0.574, 0.718) | 0.763 (0.703, 0.817) | 0.587 (0.479, 0.679) |
| | Virchow | 0.877 (0.813,0.940) | 0.506 (0.433, 0.576) | 0.813 (0.758, 0.863) | 0.643 (0.529, 0.737) |
| | Virchow2 | 0.933 (0.887,0.979) | 0.697 (0.627, 0.762) | **0.872 (0.831, 0.918)** | 0.725 (0.621, 0.817) |
| | CRISP | **0.967 (0.945,0.989)** | **0.848 (0.798, 0.898)** | 0.868 (0.817, 0.909) | **0.734 (0.635, 0.816)** |

Table S4.14. Comparative performance of models in uterus subgroup of pan-cancer classification task.

| Cohort | Model | AUROC | Spe (at 95% Sen) | Accuracy | F1 |
|---|---|---|---|---|---|
| ZSH | CHIEF | 0.732 | 0.526 | 0.581 | 0.357 |
| | CONCH | 0.916 | 0.763 | 0.791 | 0.526 |
| | UNI | 0.947 | 0.816 | 0.837 | 0.588 |
| | Virchow | 0.868 | 0.763 | 0.791 | 0.526 |
| | Virchow2 | 0.858 | 0.447 | 0.884 | 0.615 |
| | CRISP | **0.995** | **0.974** | **0.977** | **0.909** |
| NFH | CHIEF | 0.803 | 0.404 | 0.675 | 0.590 |
| | CONCH | 0.841 | 0.684 | 0.753 | 0.667 |
| | UNI | 0.804 | 0.351 | 0.831 | 0.629 |
| | Virchow | 0.797 | 0.579 | 0.675 | 0.603 |
| | Virchow2 | 0.896 | 0.474 | 0.909 | 0.811 |
| | CRISP | **0.974** | **0.965** | **0.961** | **0.927** |
| HDGZ | CHIEF | 0.889 | 0.889 | 0.900 | 0.667 |
| | CONCH | **1.000** | **1.000** | **1.000** | **1.000** |
| | UNI | **1.000** | **1.000** | **1.000** | **1.000** |
| | Virchow | **1.000** | **1.000** | **1.000** | **1.000** |
| | Virchow2 | **1.000** | **1.000** | **1.000** | **1.000** |
| | CRISP | **1.000** | **1.000** | **1.000** | **1.000** |
| Overall | CHIEF | 0.733 (0.633,0.833) | 0.298 (0.215, 0.385) | 0.562 (0.469, 0.646) | 0.457 (0.337, 0.566) |
| | CONCH | 0.881 (0.820,0.941) | 0.635 (0.540, 0.726) | 0.823 (0.754, 0.885) | 0.657 (0.500, 0.779) |
| | UNI | 0.858 (0.776,0.941) | 0.356 (0.265, 0.449) | 0.715 (0.638, 0.792) | 0.554 (0.416, 0.667) |
| | Virchow | 0.848 (0.776,0.919) | 0.673 (0.586, 0.762) | 0.731 (0.654, 0.808) | 0.588 (0.447, 0.706) |
| | Virchow2 | 0.898 (0.826,0.969) | 0.490 (0.389, 0.590) | 0.915 (0.869, 0.962) | 0.776 (0.621, 0.893) |
| | CRISP | **0.980 (0.956,1.000)** | **0.865 (0.800, 0.926)** | **0.954 (0.915, 0.985)** | **0.889 (0.780, 0.964)** |

Table S4.15. Comparative performance of models in cervix subgroup of pan-cancer classification task.

| Cohort | Model | AUROC | Spe (at 95% Sen) | Accuracy | F1 |
|---|---|---|---|---|---|
| ZSH | CHIEF | 0.835 | 0.294 | 0.833 | 0.868 |
| | CONCH | 0.911 | 0.529 | **0.881** | **0.898** |
| | UNI | 0.864 | 0.294 | 0.786 | 0.791 |
| | Virchow | 0.852 | 0.294 | 0.786 | 0.809 |
| | Virchow2 | 0.934 | 0.412 | 0.905 | 0.913 |
| | CRISP | **0.941** | **0.588** | 0.881 | 0.898 |
| NFH | CHIEF | 0.777 | 0.195 | 0.768 | 0.649 |
| | CONCH | 0.792 | 0.268 | 0.839 | 0.667 |
| | UNI | 0.911 | 0.317 | 0.875 | 0.800 |
| | Virchow | 0.837 | 0.049 | 0.875 | 0.774 |
| | Virchow2 | 0.902 | 0.195 | **0.929** | **0.857** |
| | CRISP | **0.925** | **0.415** | 0.893 | 0.813 |
| HDGZ | CHIEF | 0.833 | 0.833 | 0.875 | 0.800 |
| | CONCH | **1.000** | **1.000** | **1.000** | **1.000** |
| | UNI | **1.000** | **1.000** | **1.000** | **1.000** |
| | Virchow | 0.833 | 0.667 | 0.750 | 0.667 |
| | Virchow2 | **1.000** | **1.000** | **1.000** | **1.000** |
| | CRISP | **1.000** | **1.000** | **1.000** | **1.000** |
| Overall | CHIEF | 0.819 (0.736,0.903) | 0.391 (0.277, 0.508) | 0.774 (0.689, 0.849) | 0.755 (0.652, 0.835) |
| | CONCH | 0.870 (0.796,0.943) | 0.375 (0.262, 0.493) | 0.840 (0.764, 0.906) | 0.805 (0.699, 0.889) |
| | UNI | 0.900 (0.838,0.961) | 0.531 (0.407, 0.652) | 0.821 (0.745, 0.887) | 0.796 (0.700, 0.872) |
| | Virchow | 0.858 (0.776,0.940) | 0.219 (0.121, 0.324) | 0.830 (0.755, 0.887) | 0.791 (0.676, 0.874) |
| | Virchow2 | 0.929 (0.872,0.986) | 0.438 (0.312, 0.561) | **0.906 (0.849, 0.962)** | **0.875 (0.783, 0.944)** |
| | CRISP | **0.943 (0.897,0.988)** | **0.672 (0.548, 0.778)** | 0.887 (0.821, 0.943) | 0.860 (0.775, 0.927) |

Table S4.16. Comparative performance of models in skin subgroup of pan-cancer classification task.

| Cohort | Model | AUROC | Spe (at 95% Sen) | Accuracy | F1 |
|---|---|---|---|---|---|
| ZSH | CHIEF | 0.767 | 0.100 | 0.857 | 0.615 |
| | CONCH | 0.447 | 0.167 | 0.229 | 0.182 |
| | UNI | 0.747 | 0.333 | 0.829 | 0.500 |
| | Virchow | 0.607 | 0.133 | 0.686 | 0.353 |
| | Virchow2 | 0.693 | 0.167 | 0.914 | 0.667 |
| | CRISP | **0.940** | **0.700** | **0.971** | **0.889** |
| NFH | CHIEF | 0.870 | 0.444 | 0.829 | 0.700 |
| | CONCH | 0.954 | 0.852 | 0.886 | 0.800 |
| | UNI | 0.968 | **0.926** | **0.943** | **0.889** |
| | Virchow | 0.856 | 0.667 | 0.743 | 0.640 |
| | Virchow2 | 0.926 | 0.519 | 0.886 | 0.778 |
| | CRISP | **0.981** | 0.889 | 0.914 | 0.842 |
| HDGZ | CHIEF | **1.000** | **1.000** | **1.000** | **1.000** |
| | CONCH | **1.000** | **1.000** | **1.000** | **1.000** |
| | UNI | **1.000** | **1.000** | **1.000** | **1.000** |
| | Virchow | **1.000** | **1.000** | **1.000** | **1.000** |
| | Virchow2 | **1.000** | **1.000** | **1.000** | **1.000** |
| | CRISP | **1.000** | **1.000** | **1.000** | **1.000** |
| Overall | CHIEF | 0.783 (0.633,0.934) | 0.153 (0.068, 0.254) | 0.849 (0.780, 0.932) | 0.621 (0.370, 0.800) |
| | CONCH | 0.835 (0.687,0.984) | 0.254 (0.145, 0.370) | 0.904 (0.836, 0.973) | 0.741 (0.522, 0.889) |
| | UNI | 0.887 (0.791,0.984) | 0.407 (0.268, 0.535) | 0.890 (0.822, 0.959) | 0.750 (0.533, 0.903) |
| | Virchow | 0.774 (0.632,0.915) | 0.119 (0.048, 0.197) | 0.740 (0.644, 0.836) | 0.537 (0.324, 0.696) |
| | Virchow2 | 0.854 (0.730,0.977) | 0.322 (0.213, 0.456) | 0.904 (0.836, 0.959) | 0.741 (0.500, 0.909) |
| | CRISP | **0.973 (0.935,1.000)** | **0.746 (0.631, 0.853)** | **0.932 (0.863, 0.986)** | **0.839 (0.667, 0.960)** |

Table S4.17. Comparative performance of models in bile duct subgroup of pan-cancer classification task.

| Cohort | Model | AUROC | Spe (at 95% Sen) | Accuracy | F1 |
|---|---|---|---|---|---|
| ZSH | CHIEF | 0.918 | 0.375 | 0.897 | 0.889 |
| | CONCH | 0.913 | 0.500 | 0.897 | 0.870 |
| | UNI | 0.918 | 0.750 | 0.862 | 0.867 |
| | Virchow | 0.933 | 0.688 | 0.862 | 0.833 |
| | Virchow2 | 0.942 | **0.813** | **0.897** | **0.897** |
| | CRISP | **0.957** | **0.813** | **0.897** | **0.897** |
| NFH | CHIEF | 0.823 | 0.222 | 0.800 | 0.667 |
| | CONCH | 0.956 | 0.694 | 0.920 | 0.857 |
| | UNI | 0.853 | 0.417 | 0.840 | 0.733 |
| | Virchow | 0.903 | 0.472 | 0.860 | 0.774 |
| | Virchow2 | **0.976** | **0.917** | **0.940** | **0.903** |
| | CRISP | 0.974 | 0.861 | **0.940** | 0.897 |
| HDGZ | CHIEF | **1.000** | **1.000** | **1.000** | **1.000** |
| | CONCH | **1.000** | **1.000** | **1.000** | **1.000** |
| | UNI | **1.000** | **1.000** | **1.000** | **1.000** |
| | Virchow | **1.000** | **1.000** | **1.000** | **1.000** |
| | Virchow2 | **1.000** | **1.000** | **1.000** | **1.000** |
| | CRISP | **1.000** | **1.000** | **1.000** | **1.000** |
| Overall | CHIEF | 0.862 (0.776,0.949) | 0.304 (0.196, 0.431) | 0.821 (0.738, 0.905) | 0.746 (0.612, 0.857) |
| | CONCH | 0.945 (0.901,0.990) | 0.732 (0.614, 0.850) | 0.869 (0.798, 0.929) | 0.814 (0.700, 0.912) |
| | UNI | 0.885 (0.807,0.964) | 0.339 (0.218, 0.464) | 0.833 (0.762, 0.917) | 0.781 (0.655, 0.875) |
| | Virchow | 0.933 (0.880,0.986) | 0.625 (0.500, 0.750) | 0.845 (0.762, 0.917) | 0.800 (0.667, 0.892) |
| | Virchow2 | 0.958 (0.920,0.995) | 0.821 (0.711, 0.921) | 0.881 (0.810, 0.940) | 0.848 (0.750, 0.931) |
| | CRISP | **0.974 (0.947,1.000)** | **0.893 (0.804, 0.977)** | **0.905 (0.833, 0.964)** | **0.875 (0.772, 0.952)** |

Table S4.18. Comparative performance of models in soft tissue subgroup of pan-cancer classification task.

| Cohort | Model | AUROC | Spe (at 95% Sen) | Accuracy | F1 |
|---|---|---|---|---|---|
| ZSH | CHIEF | 0.675 | 0.000 | 0.385 | 0.556 |
|  | CONCH | 0.775 | 0.400 | 0.769 | 0.824 |
|  | UNI | 0.800 | 0.200 | 0.846 | 0.875 |
|  | Virchow | **1.000** | **1.000** | **1.000** | **1.000** |
|  | Virchow2 | 0.700 | 0.400 | 0.769 | 0.842 |
|  | CRISP | 0.975 | 0.800 | 0.923 | 0.933 |
| NFH | CHIEF | 0.747 | 0.071 | 0.720 | 0.741 |
|  | CONCH | 0.818 | 0.214 | 0.880 | 0.842 |
|  | UNI | 0.877 | 0.286 | 0.840 | 0.833 |
|  | Virchow | 0.903 | 0.500 | 0.880 | 0.857 |
|  | Virchow2 | 0.935 | **0.571** | 0.880 | 0.870 |
|  | CRISP | **0.955** | **0.571** | **0.920** | **0.909** |
| HDGZ | CHIEF | **1.000** | **1.000** | **1.000** | **1.000** |
|  | CONCH | **1.000** | **1.000** | **1.000** | **1.000** |
|  | UNI | **1.000** | 0.000 | 0.000 | 0.000 |
|  | Virchow | **1.000** | **1.000** | **1.000** | **1.000** |
|  | Virchow2 | **1.000** | **1.000** | **1.000** | **1.000** |
|  | CRISP | **1.000** | **1.000** | **1.000** | **1.000** |
| Overall | CHIEF | 0.667 (0.491,0.844) | 0.300 (0.111, 0.500) | 0.700 (0.550, 0.825) | 0.700 (0.500, 0.844) |
|  | CONCH | 0.792 (0.645,0.940) | 0.300 (0.100, 0.500) | 0.800 (0.675, 0.925) | 0.789 (0.625, 0.914) |
|  | UNI | 0.830 (0.695,0.965) | **0.600 (0.389, 0.810)** | 0.825 (0.700, 0.925) | 0.821 (0.667, 0.933) |
|  | Virchow | 0.917 (0.832,1.000) | 0.450 (0.227, 0.667) | 0.850 (0.725, 0.950) | 0.857 (0.718, 0.952) |
|  | Virchow2 | 0.872 (0.762,0.983) | **0.600 (0.389, 0.800)** | 0.850 (0.725, 0.950) | 0.850 (0.710, 0.952) |
|  | CRISP | **0.935 (0.860,1.000)** | **0.600 (0.391, 0.818)** | **0.900 (0.800, 0.975)** | **0.889 (0.759, 0.977)** |

Table S4.19. Comparative performance of models in liver subgroup of pan-cancer classification task.

| Cohort | Model | AUROC | Spe (at 95% Sen) | Accuracy | F1 |
|---|---|---|---|---|---|
| ZSH | CHIEF | 0.433 | 0.300 | 0.462 | 0.222 |
|  | CONCH | 0.900 | 0.700 | 0.769 | 0.667 |
|  | UNI | 0.933 | 0.800 | 0.846 | 0.750 |
|  | Virchow | 0.700 | 0.200 | 0.846 | 0.667 |
|  | Virchow2 | 0.833 | 0.700 | 0.769 | 0.667 |
|  | CRISP | **0.967** | **0.900** | **0.923** | **0.857** |
| NFH | CHIEF | 0.467 | 0.200 | 0.375 | 0.286 |
|  | CONCH | 0.667 | 0.400 | 0.625 | 0.667 |
|  | UNI | 0.600 | 0.000 | 0.250 | 0.250 |
|  | Virchow | 0.733 | 0.600 | 0.750 | 0.750 |
|  | Virchow2 | **1.000** | **1.000** | **1.000** | **1.000** |
|  | CRISP | **1.000** | **1.000** | **1.000** | **1.000** |
| HDGZ | CHIEF | **1.000** | **1.000** | **1.000** | **1.000** |
|  | CONCH | **1.000** | **1.000** | **1.000** | **1.000** |
|  | UNI | **1.000** | **1.000** | **1.000** | **1.000** |
|  | Virchow | **1.000** | **1.000** | **1.000** | **1.000** |
|  | Virchow2 | **1.000** | **1.000** | **1.000** | **1.000** |
|  | CRISP | **1.000** | **1.000** | **1.000** | **1.000** |
| Overall | CHIEF | 0.500 (0.222, 0.778) | 0.062 (0.000, 0.200) | 0.565 (0.348, 0.783) | 0.500 (0.211, 0.741) |
|  | CONCH | 0.795 (0.587, 1.000) | 0.438 (0.176, 0.706) | 0.783 (0.609, 0.957) | 0.667 (0.286, 0.889) |
|  | UNI | 0.679 (0.342, 1.000) | 0.000 (0.000, 0.000) | 0.783 (0.609, 0.913) | 0.667 (0.286, 0.900) |
|  | Virchow | 0.696 (0.411, 0.982) | 0.250 (0.062, 0.467) | 0.826 (0.652, 0.957) | 0.667 (0.250, 0.933) |
|  | Virchow2 | 0.866 (0.719, 1.000) | 0.688 (0.461, 0.909) | 0.783 (0.609, 0.914) | 0.737 (0.444, 0.933) |
|  | CRISP | **0.946 (0.863, 1.000)** | **0.812 (0.615, 1.000)** | **0.870 (0.696, 1.000)** | **0.824 (0.571, 1.000)** |

Table S4.20. Comparative performance of models in pancreas subgroup of pan-cancer classification task.

| Cohort | Model | AUROC | Spe (at 95% Sen) | Accuracy | F1 |
|---|---|---|---|---|---|
| ZSH | CHIEF | **1.000** | **1.000** | **1.000** | **1.000** |
|  | CONCH | **1.000** | **1.000** | **1.000** | **1.000** |
|  | UNI | 0.900 | 0.800 | 0.857 | 0.800 |
|  | Virchow | **1.000** | **1.000** | **1.000** | **1.000** |
|  | Virchow2 | **1.000** | **1.000** | **1.000** | **1.000** |
|  | CRISP | **1.000** | **1.000** | **1.000** | **1.000** |
| NFH | CHIEF | 0.894 | 0.584 | 0.856 | 0.851 |
|  | CONCH | 0.885 | 0.623 | 0.829 | 0.830 |
|  | UNI | 0.919 | 0.558 | 0.856 | 0.857 |
|  | Virchow | 0.909 | 0.636 | 0.856 | 0.853 |
|  | Virchow2 | 0.948 | 0.766 | 0.897 | 0.895 |
|  | CRISP | **0.980** | **0.857** | **0.932** | **0.929** |
| HDGZ | CHIEF | 1.000 | 1.000 | 1.000 | 1.000 |
|  | CONCH | 1.000 | 1.000 | 1.000 | 1.000 |
|  | UNI | 1.000 | 1.000 | 1.000 | 1.000 |
|  | Virchow | 1.000 | 1.000 | 1.000 | 1.000 |
|  | Virchow2 | 1.000 | 1.000 | 1.000 | 1.000 |
|  | CRISP | 1.000 | 1.000 | 1.000 | 1.000 |
| Overall | CHIEF | 0.892 (0.840,0.945) | 0.595 (0.494, 0.703) | 0.854 (0.796, 0.904) | 0.848 (0.778, 0.908) |
|  | CONCH | 0.892 (0.840,0.943) | 0.655 (0.556, 0.750) | 0.834 (0.771, 0.892) | 0.831 (0.765, 0.889) |
|  | UNI | 0.913 (0.870,0.957) | 0.560 (0.460, 0.655) | 0.854 (0.790, 0.905) | 0.854 (0.795, 0.908) |
|  | Virchow | 0.918 (0.873,0.963) | 0.667 (0.567, 0.770) | 0.866 (0.809, 0.917) | 0.861 (0.800, 0.915) |
|  | Virchow2 | 0.953 (0.925,0.982) | 0.774 (0.689, 0.857) | 0.904 (0.860, 0.943) | 0.901 (0.848, 0.944) |
|  | CRISP | **0.981 (0.965,0.996)** | **0.857 (0.776, 0.924)** | **0.936 (0.898, 0.968)** | **0.932 (0.887, 0.971)** |

Table S4.21. Comparative performance of models in colorectum subgroup of pan-cancer classification task.

| Cohort | Model | AUROC | Spe (at 95% Sen) | Accuracy | F1 |
|---|---|---|---|---|---|
| ZSH | CHIEF | **1.000** | **1.000** | **1.000** | **1.000** |
| | CONCH | **1.000** | **1.000** | **1.000** | **1.000** |
| | UNI | **1.000** | **1.000** | **1.000** | **1.000** |
| | Virchow | **1.000** | **1.000** | **1.000** | **1.000** |
| | Virchow2 | **1.000** | **1.000** | **1.000** | **1.000** |
| | CRISP | **1.000** | **1.000** | **1.000** | **1.000** |
| NFH | CHIEF | 0.821 | 0.571 | 0.727 | 0.727 |
| | CONCH | 0.893 | 0.714 | 0.818 | 0.800 |
| | UNI | **1.000** | **1.000** | **1.000** | **1.000** |
| | Virchow | 0.786 | 0.143 | 0.909 | 0.857 |
| | Virchow2 | **1.000** | **1.000** | **1.000** | **1.000** |
| | CRISP | **1.000** | **1.000** | **1.000** | **1.000** |
| HDGZ | CHIEF | 0.778 | 0.667 | 0.833 | 0.857 |
| | CONCH | **1.000** | **1.000** | **1.000** | **1.000** |
| | UNI | 0.778 | 0.667 | 0.833 | 0.857 |
| | Virchow | 0.889 | 0.667 | 0.833 | 0.857 |
| | Virchow2 | **1.000** | **1.000** | **1.000** | **1.000** |
| | CRISP | **1.000** | **1.000** | **1.000** | **1.000** |
| Overall | CHIEF | 0.837 (0.653,1.000) | 0.333 (0.118, 0.571) | 0.833 (0.667, 0.958) | 0.778 (0.500, 0.947) |
| | CONCH | 0.889 (0.734,1.000) | 0.333 (0.091, 0.588) | 0.875 (0.708, 1.000) | 0.824 (0.545, 1.000) |
| | UNI | 0.941 (0.836,1.000) | 0.867 (0.667, 1.000) | 0.917 (0.792, 1.000) | 0.900 (0.727, 1.000) |
| | Virchow | 0.844 (0.643,1.000) | 0.133 (0.000, 0.333) | 0.875 (0.750, 1.000) | 0.800 (0.545, 1.000) |
| | Virchow2 | 0.978 (0.928,1.000) | 0.800 (0.583, 1.000) | 0.958 (0.875, 1.000) | 0.941 (0.778, 1.000) |
| | CRISP | **1.000 (1.000,1.000)** | **1.000 (1.000, 1.000)** | **1.000 (1.000, 1.000)** | **1.000 (1.000, 1.000)** |

Table S4.22. Comparative performance of models in pleura subgroup of pan-cancer classification task.

| Cohort | Model | AUROC | Spe (at 95% Sen) | Accuracy | F1 |
|---|---|---|---|---|---|
| ZSH | CHIEF | 0.778 | 0.667 | 0.750 | 0.667 |
|  | CONCH | 0.815 | 0.667 | 0.750 | 0.667 |
|  | UNI | 0.667 | 0.556 | 0.667 | 0.600 |
|  | Virchow | 0.778 | 0.333 | **0.917** | **0.800** |
|  | Virchow2 | 0.778 | 0.667 | 0.750 | 0.667 |
|  | CRISP | **0.889** | **0.667** | 0.750 | 0.667 |
| NFH | CHIEF | 0.702 | 0.198 | 0.746 | 0.441 |
|  | CONCH | 0.933 | **0.755** | 0.869 | 0.721 |
|  | UNI | 0.835 | 0.302 | 0.831 | 0.633 |
|  | Virchow | 0.873 | 0.358 | 0.869 | 0.691 |
|  | Virchow2 | 0.925 | 0.330 | 0.923 | 0.815 |
|  | CRISP | **0.934** | 0.566 | **0.938** | **0.846** |
| HDGZ | CHIEF | 0.667 | 0.000 | 0.286 | 0.286 |
|  | CONCH | 0.583 | 0.000 | 0.286 | 0.286 |
|  | UNI | 0.667 | 0.000 | **0.857** | 0.800 |
|  | Virchow | 0.833 | 0.500 | **0.857** | 0.800 |
|  | Virchow2 | 0.667 | 0.000 | **0.857** | 0.800 |
|  | CRISP | **0.917** | **0.750** | **0.857** | **0.857** |
| Overall | CHIEF | 0.673 (0.560,0.786) | 0.160 (0.097, 0.230) | 0.745 (0.678, 0.812) | 0.441 (0.290, 0.583) |
|  | CONCH | 0.873 (0.783,0.962) | 0.042 (0.009, 0.082) | 0.852 (0.799, 0.906) | 0.703 (0.565, 0.810) |
|  | UNI | 0.803 (0.705,0.900) | 0.286 (0.207, 0.374) | 0.805 (0.745, 0.866) | 0.613 (0.473, 0.738) |
|  | Virchow | 0.836 (0.745,0.928) | 0.235 (0.162, 0.314) | 0.852 (0.792, 0.899) | 0.667 (0.522, 0.782) |
|  | Virchow2 | 0.890 (0.804,0.977) | 0.294 (0.218, 0.371) | 0.893 (0.839, 0.940) | 0.771 (0.648, 0.875) |
|  | CRISP | **0.914 (0.841,0.987)** | **0.571 (0.478, 0.664)** | **0.919 (0.872, 0.960)** | **0.800 (0.679, 0.900)** |

Table S4.23. Comparative performance of models in else subgroup of pan-cancer classification task.

| Cohort | Model | AUROC | Spe (at 95% Sen) | Accuracy | F1 |
|---|---|---|---|---|---|
| SYCC | CHIEF | 0.961 (0.944,0.978) | 0.821 (0.766, 0.870) | 0.917 (0.893, 0.942) | 0.927 (0.899, 0.948) |
| | CONCH | 0.986 (0.977,0.995) | 0.954 (0.921, 0.980) | 0.953 (0.933, 0.971) | 0.957 (0.936, 0.974) |
| | UNI | 0.982 (0.971,0.993) | 0.928 (0.891, 0.962) | 0.958 (0.938, 0.975) | 0.962 (0.944, 0.977) |
| | Virchow | 0.982 (0.972,0.993) | 0.944 (0.909, 0.974) | 0.962 (0.942, 0.980) | 0.966 (0.948, 0.982) |
| | Virchow2 | 0.992 (0.985,0.998) | **0.995 (0.984, 1.000)** | **0.971 (0.955, 0.984)** | **0.974 (0.958, 0.987)** |
| | CRISP | **0.993 (0.989,0.998)** | 0.985 (0.964, 1.000) | 0.969 (0.951, 0.984) | 0.972 (0.956, 0.986) |
| ZSH | CHIEF | 0.775 (0.703,0.848) | 0.220 (0.153, 0.297) | 0.680 (0.613, 0.747) | 0.627 (0.532, 0.705) |
| | CONCH | 0.857 (0.794,0.920) | 0.303 (0.229, 0.389) | 0.825 (0.768, 0.876) | 0.730 (0.639, 0.809) |
| | UNI | 0.899 (0.847,0.952) | 0.417 (0.333, 0.504) | 0.856 (0.804, 0.897) | 0.791 (0.714, 0.859) |
| | Virchow | 0.817 (0.746,0.888) | 0.136 (0.083, 0.199) | 0.799 (0.742, 0.851) | 0.698 (0.600, 0.781) |
| | Virchow2 | 0.916 (0.873,0.959) | 0.530 (0.444, 0.609) | 0.887 (0.840, 0.933) | 0.804 (0.716, 0.879) |
| | CRISP | **0.974 (0.956,0.993)** | **0.856 (0.792, 0.910)** | **0.912 (0.871, 0.948)** | **0.872 (0.807, 0.930)** |
| NFH | CHIEF | 0.906 (0.889,0.923) | 0.548 (0.502, 0.599) | 0.808 (0.786, 0.832) | 0.834 (0.811, 0.856) |
| | CONCH | 0.731 (0.701,0.761) | 0.098 (0.070, 0.127) | 0.676 (0.650, 0.705) | 0.702 (0.672, 0.731) |
| | UNI | 0.923 (0.908,0.938) | 0.622 (0.576, 0.672) | 0.829 (0.806, 0.851) | 0.853 (0.833, 0.872) |
| | Virchow | 0.906 (0.889,0.924) | 0.405 (0.358, 0.451) | 0.835 (0.813, 0.857) | 0.862 (0.841, 0.882) |
| | Virchow2 | 0.975 (0.967,0.982) | 0.845 (0.807, 0.880) | 0.913 (0.895, 0.929) | 0.928 (0.913, 0.942) |
| | CRISP | **0.983 (0.977,0.989)** | **0.916 (0.891, 0.941)** | **0.938 (0.923, 0.951)** | **0.950 (0.938, 0.962)** |
| HDGZ | CHIEF | 0.781 (0.687,0.874) | 0.315 (0.192, 0.462) | 0.750 (0.656, 0.833) | 0.707 (0.582, 0.814) |
| | CONCH | 0.831 (0.737,0.925) | 0.111 (0.036, 0.204) | 0.833 (0.760, 0.906) | 0.789 (0.679, 0.880) |
| | UNI | 0.913 (0.851,0.975) | 0.611 (0.471, 0.741) | 0.875 (0.812, 0.938) | 0.854 (0.758, 0.925) |
| | Virchow | 0.848 (0.766,0.931) | 0.370 (0.245, 0.508) | 0.823 (0.740, 0.896) | 0.779 (0.667, 0.867) |
| | Virchow2 | 0.925 (0.865,0.984) | 0.500 (0.370, 0.640) | 0.896 (0.833, 0.948) | 0.875 (0.789, 0.946) |
| | CRISP | **0.963 (0.928,0.997)** | **0.741 (0.622, 0.852)** | **0.917 (0.854, 0.969)** | **0.900 (0.827, 0.960)** |
| Overall | CHIEF | 0.896 (0.882,0.910) | 0.532 (0.496, 0.569) | 0.812 (0.794, 0.829) | 0.829 (0.812, 0.846) |
| | CONCH | 0.830 (0.812,0.848) | 0.296 (0.265, 0.329) | 0.756 (0.736, 0.775) | 0.763 (0.742, 0.784) |
| | UNI | 0.939 (0.928,0.949) | 0.657 (0.626, 0.689) | 0.864 (0.847, 0.879) | 0.875 (0.861, 0.890) |
| | Virchow | 0.919 (0.906,0.931) | 0.506 (0.468, 0.542) | 0.852 (0.836, 0.868) | 0.866 (0.851, 0.882) |
| | Virchow2 | 0.969 (0.963,0.976) | 0.816 (0.789, 0.841) | 0.908 (0.896, 0.921) | 0.916 (0.903, 0.928) |
| | CRISP | **0.983 (0.978,0.987)** | **0.907 (0.887, 0.926)** | **0.934 (0.923, 0.946)** | **0.941 (0.931, 0.951)** |

Table S4.24. Comparative performance of models in rare cancers of pan-cancer classification task.

| Cohort | Model | AUROC | Spe (at 95% Sen) | Accuracy | F1 |
|---|---|---|---|---|---|
| SYCC | CHIEF | 0.949 (0.930,0.968) | 0.805 (0.760, 0.845) | 0.900 (0.875, 0.924) | 0.880 (0.848, 0.908) |
| | CONCH | 0.968 (0.955,0.981) | 0.817 (0.776, 0.857) | 0.917 (0.895, 0.939) | 0.901 (0.874, 0.929) |
| | UNI | 0.976 (0.966,0.986) | 0.850 (0.809, 0.886) | 0.927 (0.905, 0.949) | 0.913 (0.886, 0.937) |
| | Virchow | 0.972 (0.958,0.985) | 0.900 (0.868, 0.929) | 0.932 (0.912, 0.951) | 0.922 (0.898, 0.947) |
| | Virchow2 | 0.984 (0.975,0.992) | 0.882 (0.845, 0.915) | **0.954 (0.936, 0.969)** | **0.944 (0.922, 0.964)** |
| | CRISP | **0.987 (0.979,0.994)** | **0.909 (0.878, 0.938)** | 0.951 (0.932, 0.968) | 0.941 (0.918, 0.960) |
| ZSH | CHIEF | 0.753 (0.727,0.778) | 0.247 (0.221, 0.276) | 0.702 (0.676, 0.726) | 0.619 (0.589, 0.651) |
| | CONCH | 0.889 (0.871,0.907) | 0.442 (0.413, 0.472) | 0.838 (0.820, 0.857) | 0.769 (0.739, 0.799) |
| | UNI | 0.872 (0.853,0.891) | 0.458 (0.426, 0.489) | 0.793 (0.771, 0.814) | 0.733 (0.703, 0.761) |
| | Virchow | 0.741 (0.716,0.766) | 0.356 (0.327, 0.389) | 0.679 (0.657, 0.705) | 0.657 (0.628, 0.683) |
| | Virchow2 | 0.874 (0.856,0.892) | 0.501 (0.467, 0.531) | 0.785 (0.764, 0.806) | 0.744 (0.717, 0.773) |
| | CRISP | **0.951 (0.940,0.962)** | **0.716 (0.684, 0.745)** | **0.890 (0.873, 0.906)** | **0.855 (0.832, 0.877)** |
| NFH | CHIEF | 0.796 (0.777,0.815) | 0.281 (0.258, 0.305) | 0.753 (0.734, 0.771) | 0.688 (0.663, 0.711) |
| | CONCH | 0.881 (0.866,0.896) | 0.382 (0.356, 0.407) | 0.831 (0.815, 0.846) | 0.773 (0.752, 0.794) |
| | UNI | 0.879 (0.864,0.894) | 0.426 (0.399, 0.454) | 0.825 (0.810, 0.840) | 0.777 (0.754, 0.798) |
| | Virchow | 0.867 (0.851,0.883) | 0.350 (0.324, 0.374) | 0.810 (0.794, 0.826) | 0.753 (0.732, 0.775) |
| | Virchow2 | 0.935 (0.924,0.945) | 0.647 (0.621, 0.671) | 0.883 (0.871, 0.897) | 0.849 (0.830, 0.866) |
| | CRISP | **0.966 (0.959,0.973)** | **0.848 (0.829, 0.866)** | **0.911 (0.900, 0.923)** | **0.890 (0.875, 0.905)** |
| HDGZ | CHIEF | 0.752 (0.717,0.787) | 0.144 (0.106, 0.184) | 0.712 (0.679, 0.744) | 0.746 (0.712, 0.779) |
| | CONCH | 0.791 (0.758,0.823) | 0.158 (0.121, 0.198) | 0.735 (0.703, 0.766) | 0.727 (0.690, 0.760) |
| | UNI | 0.907 (0.884,0.929) | 0.560 (0.507, 0.608) | 0.850 (0.823, 0.876) | 0.856 (0.828, 0.881) |
| | Virchow | 0.845 (0.817,0.873) | 0.302 (0.256, 0.352) | 0.784 (0.755, 0.812) | 0.786 (0.754, 0.819) |
| | Virchow2 | 0.861 (0.835,0.887) | 0.368 (0.319, 0.417) | 0.792 (0.762, 0.820) | 0.787 (0.755, 0.820) |
| | CRISP | **0.960 (0.947,0.974)** | **0.784 (0.738, 0.824)** | **0.913 (0.891, 0.932)** | **0.914 (0.893, 0.934)** |
| Overall | CHIEF | 0.795 (0.783,0.807) | 0.278 (0.263, 0.293) | 0.728 (0.717, 0.741) | 0.696 (0.681, 0.711) |
| | CONCH | 0.871 (0.861,0.881) | 0.332 (0.316, 0.347) | 0.807 (0.796, 0.818) | 0.764 (0.749, 0.779) |
| | UNI | 0.889 (0.879,0.898) | 0.458 (0.442, 0.476) | 0.821 (0.811, 0.833) | 0.783 (0.770, 0.797) |
| | Virchow | 0.830 (0.819,0.842) | 0.336 (0.320, 0.352) | 0.766 (0.755, 0.777) | 0.735 (0.720, 0.750) |
| | Virchow2 | 0.905 (0.897,0.914) | 0.539 (0.520, 0.558) | 0.832 (0.822, 0.842) | 0.801 (0.787, 0.813) |
| | CRISP | **0.962 (0.958,0.967)** | **0.798 (0.785, 0.813)** | **0.909 (0.901, 0.917)** | **0.889 (0.879, 0.899)** |

Table S4.25. Comparative performance of models in common cancers of pan-cancer classification task.

| Cohort | Model | AUROC | Spe (at 95% Sen) | Accuracy | F1 |
|---|---|---|---|---|---|
| ZSH | CHIEF | 0.832 (0.741,0.924) | 0.143 (0.078, 0.219) | 0.770 (0.690, 0.841) | 0.613 (0.484, 0.730) |
| | CONCH | 0.862 (0.778,0.945) | 0.367 (0.271, 0.465) | 0.849 (0.786, 0.913) | 0.678 (0.528, 0.806) |
| | UNI | 0.876 (0.805,0.946) | 0.490 (0.389, 0.590) | 0.778 (0.698, 0.849) | 0.641 (0.487, 0.759) |
| | Virchow | 0.819 (0.730,0.908) | 0.378 (0.286, 0.475) | 0.786 (0.714, 0.857) | 0.597 (0.426, 0.732) |
| | Virchow2 | 0.842 (0.763,0.921) | 0.429 (0.333, 0.524) | 0.786 (0.706, 0.857) | 0.609 (0.453, 0.732) |
| | CRISP | **0.950 (0.906,0.995)** | **0.786 (0.703, 0.859)** | **0.881 (0.825, 0.937)** | **0.769 (0.636, 0.871)** |
| NFH | CHIEF | 0.772 (0.711,0.833) | 0.275 (0.217, 0.328) | 0.762 (0.710, 0.809) | 0.526 (0.427, 0.610) |
| | CONCH | 0.893 (0.856,0.930) | 0.635 (0.576, 0.697) | 0.802 (0.756, 0.842) | 0.677 (0.600, 0.749) |
| | UNI | 0.852 (0.801,0.902) | 0.391 (0.333, 0.452) | 0.789 (0.742, 0.835) | 0.648 (0.565, 0.724) |
| | Virchow | 0.857 (0.803,0.910) | 0.339 (0.279, 0.403) | 0.785 (0.739, 0.828) | 0.652 (0.564, 0.726) |
| | Virchow2 | 0.932 (0.898,0.967) | 0.764 (0.706, 0.819) | 0.809 (0.762, 0.851) | 0.698 (0.620, 0.763) |
| | CRISP | **0.963 (0.931,0.995)** | **0.820 (0.770, 0.864)** | **0.934 (0.908, 0.960)** | **0.867 (0.803, 0.919)** |
| HDGZ | CHIEF | 0.707 (0.478,0.936) | 0.045 (0.000, 0.167) | 0.677 (0.516, 0.839) | 0.583 (0.300, 0.788) |
| | CONCH | 0.808 (0.557,1.000) | 0.091 (0.000, 0.211) | 0.935 (0.839, 1.000) | 0.875 (0.615, 1.000) |
| | UNI | 0.874 (0.666,1.000) | 0.045 (0.000, 0.150) | 0.935 (0.839, 1.000) | 0.889 (0.667, 1.000) |
| | Virchow | 0.949 (0.864,1.000) | 0.636 (0.416, 0.833) | 0.903 (0.774, 1.000) | 0.842 (0.615, 1.000) |
| | Virchow2 | 0.909 (0.730,1.000) | 0.182 (0.043, 0.353) | 0.968 (0.903, 1.000) | 0.941 (0.769, 1.000) |
| | CRISP | **0.955 (0.883,1.000)** | **0.727 (0.545, 0.905)** | **0.935 (0.839, 1.000)** | **0.875 (0.615, 1.000)** |
| Overall | CHIEF | 0.760 (0.707,0.812) | 0.232 (0.189, 0.278) | 0.735 (0.693, 0.776) | 0.534 (0.460, 0.600) |
| | CONCH | 0.878 (0.840,0.916) | 0.510 (0.458, 0.560) | 0.796 (0.759, 0.830) | 0.662 (0.594, 0.724) |
| | UNI | 0.854 (0.812,0.897) | 0.351 (0.297, 0.400) | 0.780 (0.743, 0.817) | 0.643 (0.577, 0.706) |
| | Virchow | 0.846 (0.803,0.889) | 0.357 (0.307, 0.406) | 0.774 (0.735, 0.807) | 0.631 (0.564, 0.690) |
| | Virchow2 | 0.900 (0.866,0.934) | 0.666 (0.616, 0.715) | 0.820 (0.780, 0.852) | 0.680 (0.609, 0.742) |
| | CRISP | **0.955 (0.930,0.980)** | **0.759 (0.715, 0.805)** | **0.937 (0.913, 0.959)** | **0.860 (0.808, 0.907)** |

Table S4.26. Comparative performance of models in unseen rare cancers of pan-cancer classification task.

| Cohort | Model | AUROC | Spe (at 95% Sen) | Accuracy | F1 |
|---|---|---|---|---|---|
| ZSH | CHIEF | 0.778 (0.744,0.811) | 0.278 (0.241, 0.314) | 0.709 (0.675, 0.738) | 0.618 (0.572, 0.660) |
| | CONCH | 0.846 (0.815,0.877) | 0.314 (0.278, 0.349) | 0.835 (0.809, 0.858) | 0.704 (0.657, 0.753) |
| | UNI | 0.845 (0.815,0.875) | 0.368 (0.329, 0.406) | 0.744 (0.714, 0.773) | 0.660 (0.617, 0.704) |
| | Virchow | 0.671 (0.634,0.709) | 0.193 (0.161, 0.226) | 0.599 (0.567, 0.632) | 0.539 (0.491, 0.582) |
| | Virchow2 | 0.841 (0.812,0.870) | 0.382 (0.339, 0.418) | 0.725 (0.695, 0.755) | 0.653 (0.607, 0.693) |
| | CRISP | **0.940 (0.924,0.957)** | **0.646 (0.606, 0.686)** | **0.898 (0.875, 0.918)** | **0.827 (0.790, 0.862)** |
| NFH | CHIEF | 0.817 (0.785,0.849) | 0.362 (0.323, 0.400) | 0.780 (0.751, 0.807) | 0.635 (0.580, 0.682) |
| | CONCH | 0.869 (0.839,0.899) | 0.310 (0.275, 0.351) | 0.824 (0.799, 0.853) | 0.706 (0.658, 0.748) |
| | UNI | 0.876 (0.849,0.904) | 0.456 (0.417, 0.494) | 0.806 (0.780, 0.832) | 0.702 (0.658, 0.745) |
| | Virchow | 0.851 (0.819,0.883) | 0.262 (0.229, 0.299) | 0.753 (0.725, 0.783) | 0.643 (0.601, 0.690) |
| | Virchow2 | 0.923 (0.901,0.945) | 0.520 (0.477, 0.560) | 0.854 (0.828, 0.878) | 0.762 (0.717, 0.806) |
| | CRISP | **0.953 (0.935,0.971)** | **0.758 (0.722, 0.792)** | **0.894 (0.873, 0.915)** | **0.820 (0.783, 0.859)** |
| HDGZ | CHIEF | 0.701 (0.564,0.838) | 0.169 (0.095, 0.245) | 0.682 (0.591, 0.764) | 0.462 (0.302, 0.603) |
| | CONCH | 0.763 (0.631,0.895) | 0.101 (0.045, 0.172) | 0.764 (0.691, 0.836) | 0.536 (0.367, 0.679) |
| | UNI | 0.845 (0.741,0.948) | 0.685 (0.576, 0.778) | 0.736 (0.655, 0.818) | 0.580 (0.441, 0.706) |
| | Virchow | 0.849 (0.762,0.935) | 0.640 (0.542, 0.733) | 0.700 (0.609, 0.782) | 0.548 (0.393, 0.676) |
| | Virchow2 | 0.892 (0.802,0.983) | **0.775 (0.689, 0.859)** | 0.809 (0.727, 0.882) | 0.656 (0.500, 0.776) |
| | CRISP | **0.929 (0.868,0.990)** | 0.685 (0.593, 0.781) | **0.936 (0.891, 0.973)** | **0.821 (0.667, 0.936)** |
| Overall | CHIEF | 0.774 (0.751,0.797) | 0.281 (0.255, 0.306) | 0.679 (0.656, 0.701) | 0.583 (0.551, 0.614) |
| | CONCH | 0.854 (0.833,0.876) | 0.301 (0.278, 0.326) | 0.804 (0.785, 0.823) | 0.686 (0.653, 0.717) |
| | UNI | 0.852 (0.832,0.873) | 0.366 (0.339, 0.393) | 0.741 (0.720, 0.760) | 0.652 (0.620, 0.678) |
| | Virchow | 0.774 (0.751,0.798) | 0.272 (0.245, 0.295) | 0.665 (0.644, 0.687) | 0.585 (0.553, 0.613) |
| | Virchow2 | 0.880 (0.863,0.898) | 0.454 (0.428, 0.483) | 0.797 (0.778, 0.815) | 0.687 (0.655, 0.716) |
| | CRISP | **0.946 (0.934,0.958)** | **0.702 (0.676, 0.726)** | **0.880 (0.864, 0.894)** | **0.802 (0.777, 0.827)** |

Table S4.27. Comparative performance of models in unseen sites of pan-cancer classification task.

| Cohort | Model | AUROC | Spe (at 95% Sen) | Accuracy | F1 |
|---|---|---|---|---|---|
| SYCC | CHIEF | 0.954 (0.941,0.967) | 0.811 (0.778, 0.844) | 0.903 (0.882, 0.921) | 0.897 (0.876, 0.917) |
| | CONCH | 0.976 (0.969,0.984) | 0.850 (0.820, 0.880) | 0.932 (0.917, 0.947) | 0.928 (0.911, 0.943) |
| | UNI | 0.978 (0.970,0.986) | 0.880 (0.850, 0.907) | 0.939 (0.924, 0.955) | 0.937 (0.919, 0.951) |
| | Virchow | 0.976 (0.967,0.985) | 0.916 (0.892, 0.939) | 0.945 (0.932, 0.959) | 0.942 (0.928, 0.956) |
| | Virchow2 | 0.988 (0.982,0.993) | 0.923 (0.899, 0.945) | **0.958 (0.945, 0.970)** | **0.955 (0.941, 0.967)** |
| | CRISP | **0.990 (0.985,0.994)** | **0.940 (0.919, 0.959)** | 0.957 (0.943, 0.969) | 0.954 (0.941, 0.967) |
| ZSH | CHIEF | 0.783 (0.752,0.814) | 0.255 (0.217, 0.292) | 0.726 (0.697, 0.758) | 0.690 (0.650, 0.725) |
| | CONCH | 0.916 (0.896,0.935) | 0.553 (0.510, 0.601) | 0.831 (0.806, 0.855) | 0.813 (0.781, 0.844) |
| | UNI | 0.901 (0.880,0.923) | 0.514 (0.472, 0.556) | 0.825 (0.801, 0.850) | 0.802 (0.771, 0.832) |
| | Virchow | 0.884 (0.861,0.908) | 0.432 (0.388, 0.473) | 0.822 (0.797, 0.846) | 0.789 (0.756, 0.821) |
| | Virchow2 | 0.923 (0.905,0.941) | 0.658 (0.616, 0.703) | 0.855 (0.831, 0.878) | 0.835 (0.804, 0.862) |
| | CRISP | **0.965 (0.953,0.977)** | **0.819 (0.786, 0.855)** | **0.921 (0.902, 0.940)** | **0.906 (0.883, 0.927)** |
| NFH | CHIEF | 0.844 (0.829,0.859) | 0.358 (0.330, 0.385) | 0.771 (0.755, 0.788) | 0.776 (0.758, 0.794) |
| | CONCH | 0.784 (0.767,0.802) | 0.143 (0.123, 0.162) | 0.715 (0.697, 0.732) | 0.675 (0.652, 0.698) |
| | UNI | 0.904 (0.892,0.916) | 0.498 (0.470, 0.528) | 0.841 (0.826, 0.854) | 0.844 (0.830, 0.859) |
| | Virchow | 0.891 (0.879,0.904) | 0.410 (0.382, 0.439) | 0.816 (0.801, 0.832) | 0.812 (0.794, 0.828) |
| | Virchow2 | 0.957 (0.949,0.964) | 0.746 (0.719, 0.772) | 0.902 (0.890, 0.913) | 0.906 (0.895, 0.918) |
| | CRISP | **0.975 (0.970,0.981)** | **0.893 (0.875, 0.910)** | **0.927 (0.917, 0.936)** | **0.932 (0.922, 0.941)** |
| HDGZ | CHIEF | 0.765 (0.730,0.799) | 0.150 (0.114, 0.189) | 0.727 (0.696, 0.759) | 0.768 (0.735, 0.796) |
| | CONCH | 0.800 (0.768,0.832) | 0.150 (0.113, 0.193) | 0.741 (0.710, 0.771) | 0.737 (0.700, 0.772) |
| | UNI | 0.913 (0.891,0.934) | 0.528 (0.473, 0.585) | 0.856 (0.829, 0.879) | 0.868 (0.845, 0.892) |
| | Virchow | 0.856 (0.829,0.883) | 0.301 (0.252, 0.348) | 0.794 (0.764, 0.822) | 0.808 (0.776, 0.836) |
| | Virchow2 | 0.871 (0.846,0.896) | 0.393 (0.341, 0.446) | 0.796 (0.767, 0.824) | 0.804 (0.772, 0.836) |
| | CRISP | **0.965 (0.952,0.977)** | **0.801 (0.755, 0.842)** | **0.917 (0.897, 0.936)** | **0.923 (0.903, 0.941)** |
| Overall | CHIEF | 0.846 (0.835,0.856) | 0.370 (0.352, 0.389) | 0.775 (0.763, 0.787) | 0.775 (0.763, 0.789) |
| | CONCH | 0.856 (0.846,0.866) | 0.336 (0.319, 0.355) | 0.778 (0.767, 0.789) | 0.766 (0.753, 0.779) |
| | UNI | 0.917 (0.909,0.924) | 0.556 (0.536, 0.576) | 0.848 (0.838, 0.857) | 0.846 (0.835, 0.855) |
| | Virchow | 0.900 (0.892,0.909) | 0.441 (0.423, 0.462) | 0.830 (0.821, 0.841) | 0.826 (0.815, 0.838) |
| | Virchow2 | 0.946 (0.940,0.952) | 0.672 (0.654, 0.691) | 0.884 (0.875, 0.891) | 0.883 (0.873, 0.892) |
| | CRISP | **0.974 (0.971,0.978)** | **0.884 (0.872, 0.896)** | **0.928 (0.921, 0.935)** | **0.929 (0.921, 0.935)** |

Table S4.28. Comparative performance of models in seen sites of pan-cancer classification task.

| Model | AUROC | Spe (at 90% Sen) | Spe (at 95% Sen) | Accuracy | F1 |
| --- | --- | --- | --- | --- | --- |
| CHIEF | 0.776 (0.752, 0.799) | 0.367 (0.333, 0.408) | 0.250 (0.214, 0.284) | 0.723 (0.704, 0.741) | 0.784 (0.766, 0.801) |
| CONCH | 0.833 (0.815, 0.850) | 0.488 (0.448, 0.527) | 0.326 (0.288, 0.361) | 0.733 (0.713, 0.754) | 0.791 (0.773, 0.807) |
| UNI | 0.867 (0.848, 0.886) | 0.590 (0.549, 0.627) | 0.390 (0.352, 0.430) | 0.822 (0.806, 0.838) | 0.868 (0.855, 0.881) |
| Virchow | 0.798 (0.779, 0.816) | 0.329 (0.291, 0.367) | 0.153 (0.126, 0.184) | 0.676 (0.656, 0.696) | 0.713 (0.693, 0.734) |
| Virchow2 | 0.916 (0.904, 0.927) | 0.602 (0.564, 0.642) | 0.495 (0.452, 0.538) | 0.820 (0.804, 0.837) | 0.861 (0.846, 0.874) |
| CRISP | **0.975 (0.969, 0.981)** | **0.972 (0.957, 0.985)** | **0.809 (0.777, 0.839)** | **0.926 (0.915, 0.938)** | **0.947 (0.938, 0.955)** |

Table S5.1. Comparison of overall model performance in the prospective cohort.

| Task | Model | AUROC | Spe (at 90% Sen) | Spe (at 95% Sen) | Accuracy | F1 |
|---|---|---|---|---|---|---|
| PNBM | CHIEF | 0.884 (0.845, 0.923) | 0.576 (0.458, 0.692) | 0.394 (0.286, 0.513) | 0.777 (0.734, 0.820) | 0.845 (0.809, 0.879) |
| | CONCH | 0.884 (0.846, 0.923) | 0.667 (0.551, 0.778) | 0.333 (0.222, 0.453) | 0.806 (0.769, 0.849) | 0.870 (0.837, 0.899) |
| | UNI | 0.960 (0.942, 0.979) | 0.939 (0.877, 0.986) | 0.667 (0.547, 0.783) | 0.917 (0.889, 0.943) | 0.947 (0.928, 0.965) |
| | Virchow | 0.888 (0.850, 0.926) | 0.636 (0.516, 0.759) | 0.303 (0.197, 0.417) | 0.806 (0.763, 0.843) | 0.868 (0.834, 0.898) |
| | Virchow2 | 0.960 (0.940, 0.980) | 0.864 (0.776, 0.938) | 0.697 (0.580, 0.803) | 0.891 (0.857, 0.920) | 0.929 (0.906, 0.951) |
| | CRISP | **0.991 (0.981, 1.000)** | **0.985 (0.951, 1.000)** | **0.955 (0.902, 1.000)** | **0.951 (0.929, 0.971)** | **0.969 (0.954, 0.984)** |
| TNBM | CHIEF | 0.943 (0.928, 0.958) | 0.857 (0.797, 0.912) | 0.586 (0.504, 0.667) | 0.878 (0.858, 0.897) | 0.925 (0.913, 0.937) |
| | CONCH | 0.943 (0.930, 0.957) | 0.864 (0.810, 0.918) | 0.586 (0.500, 0.672) | 0.867 (0.846, 0.887) | 0.918 (0.904, 0.930) |
| | UNI | 0.954 (0.942, 0.965) | 0.921 (0.873, 0.964) | 0.521 (0.439, 0.602) | 0.884 (0.864, 0.904) | 0.929 (0.916, 0.942) |
| | Virchow | 0.947 (0.934, 0.960) | 0.821 (0.759, 0.884) | 0.593 (0.509, 0.669) | 0.879 (0.859, 0.898) | 0.925 (0.913, 0.939) |
| | Virchow2 | 0.941 (0.927, 0.954) | 0.771 (0.691, 0.841) | 0.479 (0.401, 0.562) | 0.852 (0.830, 0.874) | 0.907 (0.892, 0.921) |
| | CRISP | **0.972 (0.962, 0.983)** | **0.986 (0.964, 1.000)** | **0.779 (0.706, 0.846)** | **0.924 (0.907, 0.939)** | **0.954 (0.944, 0.963)** |
| BNBM | CHIEF | 0.890 (0.827, 0.953) | 0.635 (0.532, 0.747) | 0.527 (0.414, 0.638) | 0.827 (0.750, 0.894) | 0.719 (0.596, 0.831) |
| | CONCH | 0.912 (0.853, 0.970) | 0.649 (0.536, 0.753) | 0.459 (0.342, 0.563) | 0.837 (0.760, 0.904) | 0.746 (0.618, 0.847) |
| | UNI | 0.962 (0.925, 0.999) | 0.892 (0.819, 0.956) | 0.527 (0.409, 0.645) | 0.904 (0.846, 0.962) | 0.848 (0.739, 0.933) |
| | Virchow | 0.940 (0.895, 0.985) | 0.824 (0.727, 0.915) | 0.595 (0.480, 0.704) | 0.856 (0.788, 0.923) | 0.789 (0.676, 0.881) |
| | Virchow2 | 0.935 (0.888, 0.981) | 0.770 (0.667, 0.872) | 0.581 (0.468, 0.691) | 0.865 (0.798, 0.933) | 0.794 (0.667, 0.886) |
| | CRISP | **0.981 (0.961, 1.000)** | **0.946 (0.885, 0.987)** | **0.919 (0.855, 0.974)** | **0.942 (0.885, 0.981)** | **0.909 (0.824, 0.973)** |
| CLMD | CHIEF | 0.939 (0.912, 0.967) | 0.755 (0.667, 0.836) | 0.461 (0.371, 0.558) | 0.887 (0.850, 0.921) | 0.903 (0.863, 0.936) |
| | CONCH | 0.953 (0.925, 0.980) | 0.951 (0.904, 0.989) | 0.422 (0.321, 0.515) | 0.932 (0.898, 0.962) | 0.942 (0.912, 0.966) |
| | UNI | 0.978 (0.961, 0.996) | **1.000 (1.000, 1.000)** | 0.902 (0.843, 0.958) | 0.955 (0.929, 0.977) | 0.962 (0.939, 0.981) |
| | Virchow | 0.979 (0.962, 0.995) | 0.971 (0.933, 1.000) | 0.931 (0.878, 0.980) | 0.944 (0.914, 0.966) | 0.953 (0.925, 0.974) |
| | Virchow2 | 0.981 (0.965, 0.996) | **1.000 (1.000, 1.000)** | **1.000 (1.000, 1.000)** | **0.974 (0.955, 0.989)** | **0.978 (0.960, 0.992)** |
| | CRISP | **0.994 (0.988, 0.999)** | **1.000 (1.000, 1.000)** | 0.951 (0.900, 0.989) | 0.962 (0.936, 0.981) | 0.969 (0.949, 0.986) |
| SLMD | CHIEF | 0.962 (0.932, 0.992) | 0.755 (0.678, 0.835) | 0.609 (0.519, 0.700) | 0.925 (0.887, 0.962) | 0.878 (0.800, 0.935) |
| | CONCH | 0.880 (0.815, 0.945) | 0.418 (0.325, 0.510) | 0.336 (0.250, 0.430) | 0.893 (0.843, 0.931) | 0.800 (0.699, 0.886) |
| | UNI | 0.926 (0.880, 0.972) | 0.664 (0.573, 0.752) | 0.436 (0.345, 0.530) | 0.862 (0.811, 0.918) | 0.788 (0.696, 0.871) |
| | Virchow | 0.946 (0.911, 0.981) | 0.818 (0.746, 0.891) | 0.582 (0.496, 0.669) | 0.893 (0.843, 0.937) | 0.838 (0.759, 0.908) |
| | Virchow2 | 0.930 (0.886, 0.973) | 0.755 (0.670, 0.833) | 0.645 (0.556, 0.735) | 0.887 (0.836, 0.931) | 0.816 (0.724, 0.889) |
| | CRISP | **0.985 (0.972, 0.998)** | **0.918 (0.862, 0.963)** | **0.855 (0.786, 0.916)** | **0.943 (0.906, 0.975)** | **0.909 (0.845, 0.964)** |
| BSMA | CHIEF | 0.782 (0.703, 0.861) | 0.443 (0.341, 0.552) | 0.398 (0.298, 0.507) | 0.719 (0.641, 0.797) | 0.654 (0.544, 0.763) |
| | CONCH | 0.746 (0.634, 0.859) | 0.114 (0.052, 0.183) | 0.091 (0.034, 0.156) | 0.844 (0.773, 0.906) | 0.706 (0.576, 0.821) |
| | UNI | 0.836 (0.762, 0.910) | 0.455 (0.349, 0.560) | 0.375 (0.279, 0.478) | 0.742 (0.664, 0.820) | 0.667 (0.555, 0.757) |
| | Virchow | 0.891 (0.828, 0.954) | 0.568 (0.473, 0.667) | 0.386 (0.287, 0.494) | 0.875 (0.812, 0.930) | 0.800 (0.703, 0.881) |
| | Virchow2 | 0.940 (0.895, 0.985) | 0.784 (0.690, 0.872) | 0.489 (0.382, 0.598) | 0.906 (0.859, 0.953) | 0.846 (0.746, 0.923) |
| | CRISP | **0.973 (0.944, 1.000)** | **0.898 (0.828, 0.953)** | **0.739 (0.641, 0.826)** | **0.938 (0.891, 0.977)** | **0.900 (0.823, 0.961)** |

Table S5.2. Comparison of model performance across tasks in the prospective cohort.

| Histotechnologist Group | Cases | AUROC |
|---|---|---|
| H1 | 8 | 1.000 |
| H2 | 19 | 1.000 |
| H3 | 30 | 1.000 |
| H4 | 52 | 1.000 |
| H5 | 16 | 1.000 |
| H6 | 2 | 1.000 |
| H7 | 5 | 1.000 |
| H8 | 113 | 0.996 |
| H9 | 191 | 0.995 |
| H10 | 100 | 0.994 |
| H11 | 80 | 0.993 |
| H12 | 110 | 0.992 |
| H13 | 89 | 0.991 |
| H14 | 78 | 0.991 |
| H15 | 82 | 0.990 |
| H16 | 27 | 0.986 |
| H17 | 87 | 0.985 |
| H18 | 101 | 0.984 |
| H19 | 115 | 0.983 |
| H20 | 141 | 0.979 |
| H21 | 99 | 0.976 |
| H22 | 85 | 0.970 |
| H23 | 80 | 0.964 |
| H24 | 70 | 0.963 |
| H25 | 73 | 0.955 |
| H26 | 100 | 0.954 |
| Else | 118 | 0.948 |
| | 2071 (Sum) | 0.975 (Overall) |

Table S5.3. Case distribution and CRISP performance across histotechnologist subgroups.

| Task | Cases | AUROC | Accuracy | F1 |
| --- | --- | --- | --- | --- |
| PNBM | 61 | 0.992 (0.977, 1.000) | 0.967 (0.918, 1.000) | 0.979 (0.944, 1.000) |
| TNBM | 184 | 0.942 (0.909, 0.974) | 0.853 (0.799, 0.897) | 0.852 (0.795, 0.905) |
| BNBM | 29 | 0.974 (0.920, 1.000) | 0.966 (0.897, 1.000) | 0.933 (0.750, 1.000) |
| CLMD | 37 | 1.000 (1.000, 1.000) | 1.000 (1.000, 1.000) | 1.000 (1.000, 1.000) |
| SLMD | 23 | 0.944 (0.860, 1.000) | 0.870 (0.739, 1.000) | 0.880 (0.720, 1.000) |
| BSMA | 37 | 0.926 (0.794, 1.000) | 0.919 (0.811, 1.000) | 0.857 (0.667, 1.000) |
| Overall | 371 | 0.936 (0.912, 0.959) | 0.846 (0.809, 0.884) | 0.846 (0.807, 0.885) |

Table S5.4. CRISP performance on challenging cases across tasks.

| Task | Cases | AUROC | Accuracy | F1 |
| --- | --- | --- | --- | --- |
| PNBM | 155 | 0.975 (0.945, 1.000) | 0.948 (0.910, 0.981) | 0.969 (0.944, 0.987) |
| TNBM | 218 | 0.943 (0.907, 0.980) | 0.858 (0.807, 0.899) | 0.897 (0.857, 0.932) |
| BNBM | 40 | 1.000 (1.000, 1.000) | 1.000 (1.000, 1.000) | 1.000 (1.000, 1.000) |
| CLMD | 44 | 1.000 (1.000, 1.000) | 1.000 (1.000, 1.000) | 1.000 (1.000, 1.000) |
| SLMD | 21 | 0.954 (0.875, 1.000) | 0.905 (0.762, 1.000) | 0.909 (0.750, 1.000) |
| BSMA | 32 | 0.996 (0.984, 1.000) | 0.969 (0.906, 1.000) | 0.974 (0.914, 1.000) |
| Overall | 510 | 0.963 (0.946, 0.979) | 0.898 (0.871, 0.924) | 0.925 (0.902, 0.945) |

Table S5.5. CRISP performance in IHC-requiring cases across tasks.